%% file: main.tex
\documentclass[10pt,twocolumn,letterpaper]{article}

\usepackage{cvpr}              % To produce the CAMERA-READY version

\definecolor{cvprblue}{rgb}{0.21,0.49,0.74}
\usepackage[pagebackref,breaklinks,colorlinks,allcolors=cvprblue]{hyperref}

\def\paperID{16} % *** Enter the Paper ID here
\def\confName{CVPR}
\def\confYear{2025}

\usepackage{Includes/Commands}
\usepackage{Includes/Packages}
\newcommand\blfootnote[1]{%
  \begingroup
  \renewcommand\thefootnote{}\footnote{#1}%
  \addtocounter{footnote}{-1}%
  \endgroup
}

\begin{document}
%%%%%%%%% TITLE - PLEASE UPDATE
\title{Efficient Image Generation with Variadic Attention Heads}

%%%%%%%%% AUTHORS - PLEASE UPDATE
\author{%
    Steven Walton\textsuperscript{1,2}\footnote{swalton2@uoregon.edu}\quad
    Ali Hassani\textsuperscript{2}\quad
    Xingqian Xu\textsuperscript{2}\quad
    Zhangyang Wang\textsuperscript{3}\quad
    Humphrey Shi\textsuperscript{2}
    \\
    {\textsuperscript{1}\small University of Oregon}\quad
    %{\textsuperscript{2}\small SHI Lab @ Georgia Institute of Technology}\\
    {\textsuperscript{2}\small SHI Lab @ GaTech \& UIUC}\quad
    %{\textsuperscript{3}\small SHI Lab @ University of Illinois Urbana-Champaign}\quad
    {\textsuperscript{3}\small University of Texas at Austin}
}

%\begin{document}
\twocolumn[{
\renewcommand\twocolumn[1][]{#1}
\maketitle
\begin{center}
    \centering
    \captionsetup{type=figure}
    \includegraphics[width=1.00\textwidth]{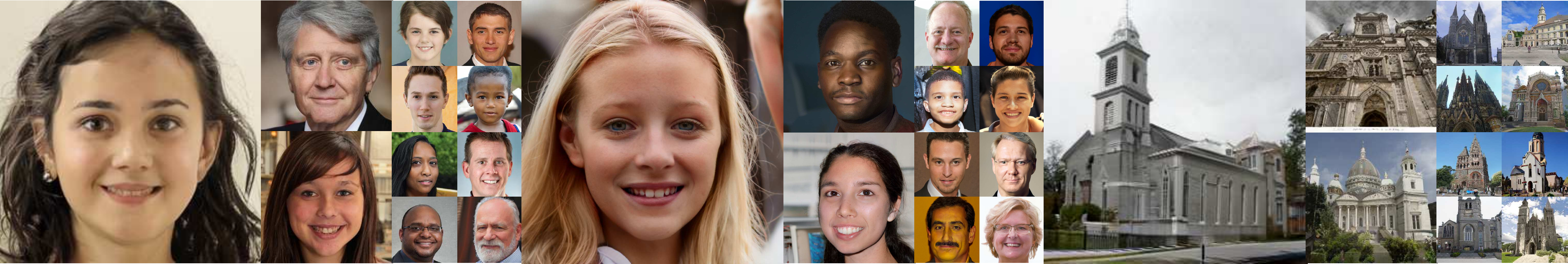}
    \captionof{figure}{Samples from FFHQ-256 (left) with FID:2.05, FFHQ-1024
    (center) with FID:4.17, and Church (right) with FID:3.40}
    \label{fig:header}
\end{center}
}]

\blfootnote{\textsuperscript{*}Corresponding author: swalton2@uoregon.edu}
\input{sec/0_abstract}
\input{sec/1_intro}

\input{sec/2_relatedworks}

\input{sec/3_methods}
\input{sec/4_experiments}

\input{sec/5_conclusion}

{
    \small
    \bibliographystyle{ieeenat_fullname}
    \bibliography{main}
}
\clearpage
\input{sec/6_appendix}

\end{document}

%% file: sec/0_abstract.tex
\begin{abstract}

While the integration of transformers in vision models have yielded significant
improvements on vision tasks they still require significant amounts of
computation for both training and inference.
Restricted attention mechanisms significantly reduce these computational burdens
but come at the cost of losing either global or local coherence.
We propose a simple, yet powerful method to reduce these trade-offs: allow
the attention heads of a single transformer to attend to multiple receptive
fields.

We demonstrate our method utilizing Neighborhood Attention (NA) and integrate it
into a StyleGAN based architecture for image generation. 
With this work, dubbed StyleNAT, we are able to achieve a FID of 2.05 on FFHQ, a 6\% improvement
over StyleGAN-XL, while utilizing 28\% fewer parameters and with 4$\times$ the
throughput capacity. 
StyleNAT achieves the Pareto Frontier on FFHQ-256 and demonstrates powerful and
efficient image generation on other datasets.
Our code and model checkpoints are publicly available at:
%\textbf{redacted for review}
\url{https://github.com/SHI-Labs/StyleNAT}

\end{abstract}

%% file: sec/1_intro.tex
\section{Introduction}\label{sec:intro}
Over the last decade the quality of image generation has significantly changed.
From Goodfellow~\etal's introduction of the Generative Adversarial Network
(GAN)~\cite{NIPS2014_f033ed80} barely able to create low resolution images of
faces to today's modern networks capable of generating high resolution realistic
imagery~\cite{esser2024scalingrectifiedflowtransformers,sauer2024fast,imagenteamgoogle2024imagen3,betker2023improving,podell2024sdxl}.
Yet, an ever-existing challenge remains: how to generate images that are 
high-resolution, photo-realistic, and to do this as quickly and cheaply as
possible.

\input{Includes/Figures/pareto}
Convolution based
GANs~\cite{karras2018progressive,Karras_2019_CVPR,Karras_2020_CVPR,NEURIPS2020_8d30aa96,NEURIPS2021_076ccd93,sauer2022stylegan},
and progressively growing
GANs~\cite{karras2018progressive,Karras_2019_CVPR,Karras_2020_CVPR},
have shown themselves to be fast, being able to generate images in `real-time'
($<$30 fps) and have even been viable on mobile based
hardware~\cite{belousov2021mobilestylegan}.
These inference speeds are necessary for applications such as Video Games 
graphics~\cite{hofmann3451256,muller3459812,watson10478358}, 
video conferencing and streaming~\cite{Kim_Oh_Kim_2020}, 
augmented reality~\cite{xiao3392376}, 
real-time face editing/filtering.
On the other hand, diffusion
models~\cite{NEURIPS2020_4c5bcfec,song2021denoising,podell2024sdxl,crowson2024scalable,imagenteamgoogle2024imagen3,esser2024scalingrectifiedflowtransformers}
have shown the capability of generating high quality and diverse imagery, but
come at the cost of requiring a large number of parameters and requiring large
data.
Being more computationally intensive they are ill-suited for real-time
applications but superior for tasks such as image-editing.
The computational bottleneck of these models is their attention
mechanism~\cite{crowson2024scalable}, but this is also what provides high
performance.
Conversely, many attempts have been made to integrate transformers into
GANs~\cite{hudson2021ganformer,hudson2021ganformer2,bond2022unleashing,NEURIPS2021_98dce83d},
including those with restricted receptive fields~\cite{Zhang_2022_CVPR}, but
these struggle to outperform their convolution based
counterparts~\cite{sauer2022stylegan,NEURIPS2021_9219adc5}.

Restricted attention
mechanisms~\cite{Hassani_2023_CVPR}
\cite{hassani2022dilated,hassani2024fasterneighborhoodattentionreducing,Liu_2022_CVPR,Liu_2021_ICCV,zhang2021aggregating}
give hope for solving this computation issue while achieving performances
similar to transformers, but when applied to GANs these
models~\cite{NEURIPS2021_98dce83d,Zhang_2022_CVPR} still fall behind convolution
counterparts~\cite{sauer2022stylegan}.
We hypothesize that this is due to the transformer's need for heavy
augmentations or large amounts of data in order to achieve high
performance~\cite{steiner2022how,hassani2022escapingbigdataparadigm,dosovitskiy2021imageworth16x16words},
thus struggling to optimize on smaller datasets, especially architectures that
are sensitive to augmentation~\cite{NEURIPS2020_8d30aa96}.

We propose a simple, yet effective solution: \emph{allow attention heads to be
independent} and attend to differing receptive fields.
This reduces restrictions from attention mechanisms with restricted receptive
fields, allowing for a
single restricted attention mechanism to incorporate both local information and
global. 
We demonstrate the effectiveness of this method by training a StyleGAN based
network and achieving real-time and high quality image generation, significantly
outperforming models with many more parameters.
This shows that our method leads to effective learning and increases the utility
of each neuron.

\noindent\textbf{Our main contributions are}:
\begin{itemize}
    \item Hydra-NA: a variadic attention head implementation of Neighborhood
        Attention (NA)~\cite{Hassani_2023_CVPR}.
    \item Demonstrating restricted attention mechanisms can incorporate both
        local and global information.
    \item StyleNAT: a style-based image generator achieving the Pareto-Frontier
        on FFHQ-256 and high performance on other datasets.
    \item A method for visualizing localized attention windows, including
        Swin~\cite{Liu_2021_ICCV} and NA.
\end{itemize}

%% file: Includes/Figures/pareto.tex
\begin{figure*}[htbp]
    \centering
    \includegraphics[trim={0.4cm 0.4cm 0.4cm 0.4cm},clip,width=0.93\textwidth]{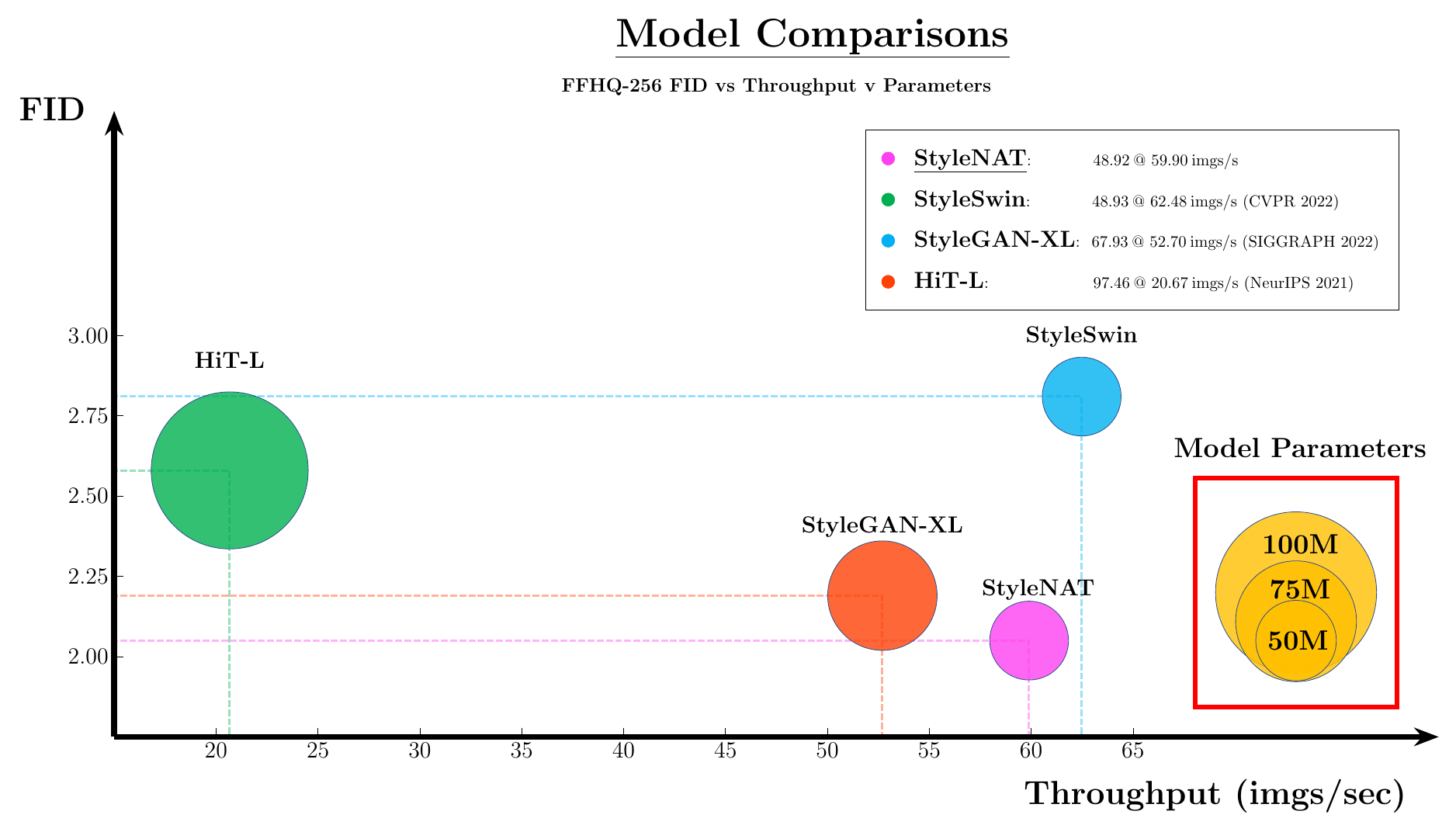}
    \caption{StyleNAT represents the Pareto Frontier for FID, Parameters, and
        Throughput.  FID (y-axis), throughput (x-axis), and the model size 
        (bubble size). 
        StyleNAT performs the best having the lowest FID (2.05),
        the fewest number of parameters (48.92M), and achieves real-time image
        geneartion (59.90 imgs/s).
        StyleNAT and StyleSwin have a similar number of
        parameters but StyleNAT has significantly better FID, and fidelity. 
        StyleGAN-XL has the closest FID but is substantially slower and larger. 
        HiT-L is equally fast, but larger and far less accurate.
        StyleSwin is only slightly faster, but equal size and significantly
        lower quality.
        }\label{fig:exp-pareto}
\end{figure*}

%% file: sec/2_relatedworks.tex
\section{Related Work}\label{sec:relatedwork}

Incorporating attention into synthetic image generators has proved to be
challenging due to the computational and memory complexities of the attention
mechanism.
Researchers have addressed this issue in various manners, including using
attention on latent spaces -- as is common in latent diffusion 
models~\cite{Rombach_2022_CVPR,podell2024sdxl,sauer2024fast} --, as well as
utilizing localized 
attention~\cite{NEURIPS2021_076ccd93,sauer2022stylegan,NEURIPS2021_98dce83d}.
We introduce some of these concepts here, along with StyleGAN, which will serve
as the base model for our experimentation.

\input{Includes/Figures/progressive.tex}
\input{sec/2_rw/StyleGAN}

\input{sec/2_rw/Attention}

%% file: Includes/Figures/progressive.tex
\begin{figure*}
    \centering
    \begin{subfigure}[b]{\linewidth}
        \centering
        % Layer 2 16
        %\hspace{-0.4em}
        \begin{subfigure}[b]{0.14\linewidth}
            \adjustbox{trim=0 {0.75\height} {0.75\width} 0, clip}
                {\includegraphics[width=\textwidth]{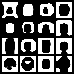}}
                \hspace{-0.49em}
            \adjustbox{trim=0 {0.75\height} {0.75\width} 0, clip}
                {\includegraphics[width=\textwidth]{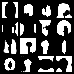}}
            \\
            \adjustbox{trim={0.75\width} 0 0 {0.75\height}, clip}
                {\includegraphics[width=\textwidth]{Images/NA/NATLayer_20kerneldensity_channel.png}}
                \hspace{-0.49em}
            \adjustbox{trim={0.75\width} 0 0 {0.75\height}, clip}
                {\includegraphics[width=\textwidth]{Images/NA/NATLayer_21kerneldensity_channel.png}}
        \end{subfigure}
        \hspace{-3.90em}
        %% Layer 3 32
        \begin{subfigure}[b]{0.17\linewidth}
            \adjustbox{trim=0 {0.75\height} {0.75\width} 0, clip}
                {\includegraphics[width=\textwidth]{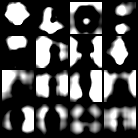}}
                \hspace{-0.50em}
            \adjustbox{trim=0 {0.75\height} {0.75\width} 0, clip}
                {\includegraphics[width=\textwidth]{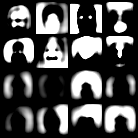}}
            \\
            \adjustbox{trim={0.75\width} 0 0 {0.75\height}, clip}
                {\includegraphics[width=\textwidth]{Images/NA/NATLayer_30kerneldensity_channel.png}}
                \hspace{-0.50em}
            \adjustbox{trim={0.75\width} 0 0 {0.75\height}, clip}
                {\includegraphics[width=\textwidth]{Images/NA/NATLayer_31kerneldensity_channel.png}}
        \end{subfigure}
        \hspace{-4.62em}
        % Layer 4 64
        \begin{subfigure}[b]{0.20\linewidth}
            \adjustbox{trim=0 {0.75\height} {0.75\width} 0, clip}
                {\includegraphics[width=\textwidth]{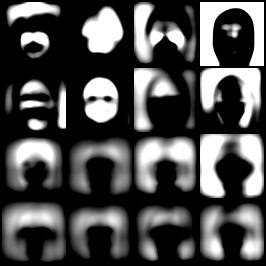}}
                \hspace{-0.50em}
            \adjustbox{trim=0 {0.75\height} {0.75\width} 0, clip}
                {\includegraphics[width=\textwidth]{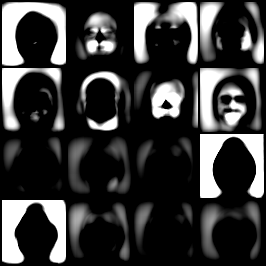}}
            \\
            \adjustbox{trim={0.75\width} 0 0 {0.75\height}, clip}
                {\includegraphics[width=\textwidth]{Images/NA/NATLayer_40kerneldensity_channel.png}}
                \hspace{-0.50em}
            \adjustbox{trim={0.75\width} 0 0 {0.75\height}, clip}
                {\includegraphics[width=\textwidth]{Images/NA/NATLayer_41kerneldensity_channel.png}}
        \end{subfigure}
        \hspace{-5.42em}
        % Layer 5 128
        \begin{subfigure}[b]{0.17\linewidth}
            \adjustbox{trim=0 {0.66\height} {0.66\width} 0, clip}
                {\includegraphics[width=\textwidth]{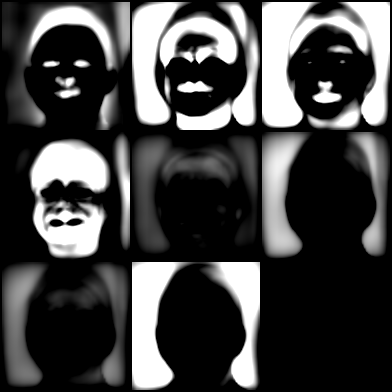}}
                \hspace{-0.50em}
            \adjustbox{trim=0 {0.66\height} {0.66\width} 0, clip}
                {\includegraphics[width=\textwidth]{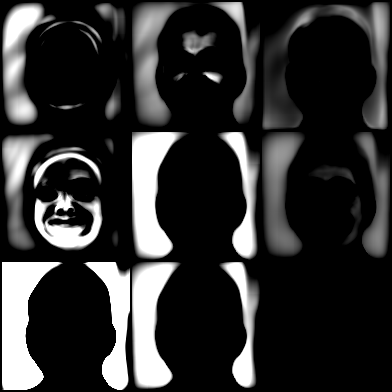}}
            \\
            \adjustbox{trim={0.33\width} 0 {0.33\width} {0.66\height}, clip}
                {\includegraphics[width=\textwidth]{Images/NA/NATLayer_50kerneldensity_channel.png}}
                \hspace{-0.50em}
            \adjustbox{trim={0.33\width} 0 {0.33\width} {0.66\height}, clip}
                {\includegraphics[width=\textwidth]{Images/NA/NATLayer_51kerneldensity_channel.png}}
        \end{subfigure}
        \hspace{-3.17em}
        % Layer 6 256
        \begin{subfigure}[b]{0.14\linewidth}
            \adjustbox{trim=0 {0.50\height} {0.50\width} 0, clip}
                {\includegraphics[width=0.95\textwidth]{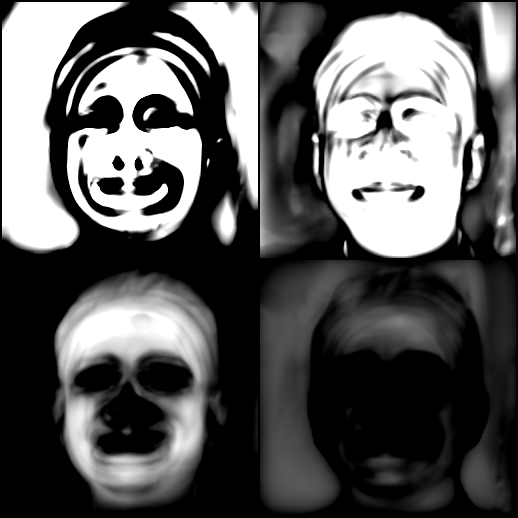}}
                \hspace{-0.50em}
            \adjustbox{trim=0 {0.50\height} {0.50\width} 0, clip}
                {\includegraphics[width=0.95\textwidth]{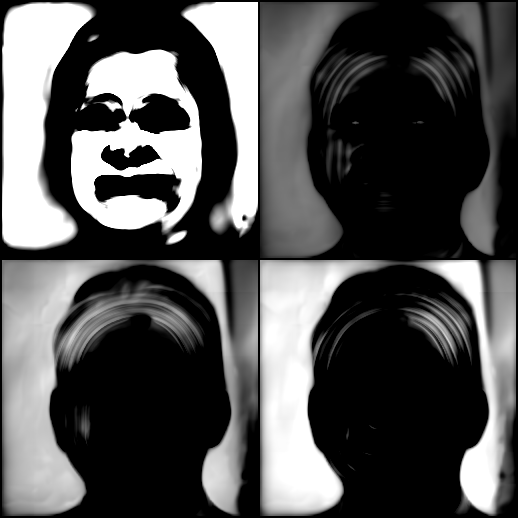}}
            \\
            \adjustbox{trim={0.5\width} 0 0 {0.5\height}, clip}
                {\includegraphics[width=0.95\textwidth]{Images/NA/NATLayer_60kerneldensity_channel.png}}
                \hspace{-0.50em}
            \adjustbox{trim={0.5\width} 0 0 {0.5\height}, clip}
                {\includegraphics[width=0.95\textwidth]{Images/NA/NATLayer_61kerneldensity_channel.png}}
        \end{subfigure}
        \hspace{-0.810em}
        % Layer 7 512
        \begin{subfigure}[b]{0.16\linewidth}
            \adjustbox{trim=0 {0.50\height} {0.50\width} 0, clip}
                {\includegraphics[width=0.95\textwidth]{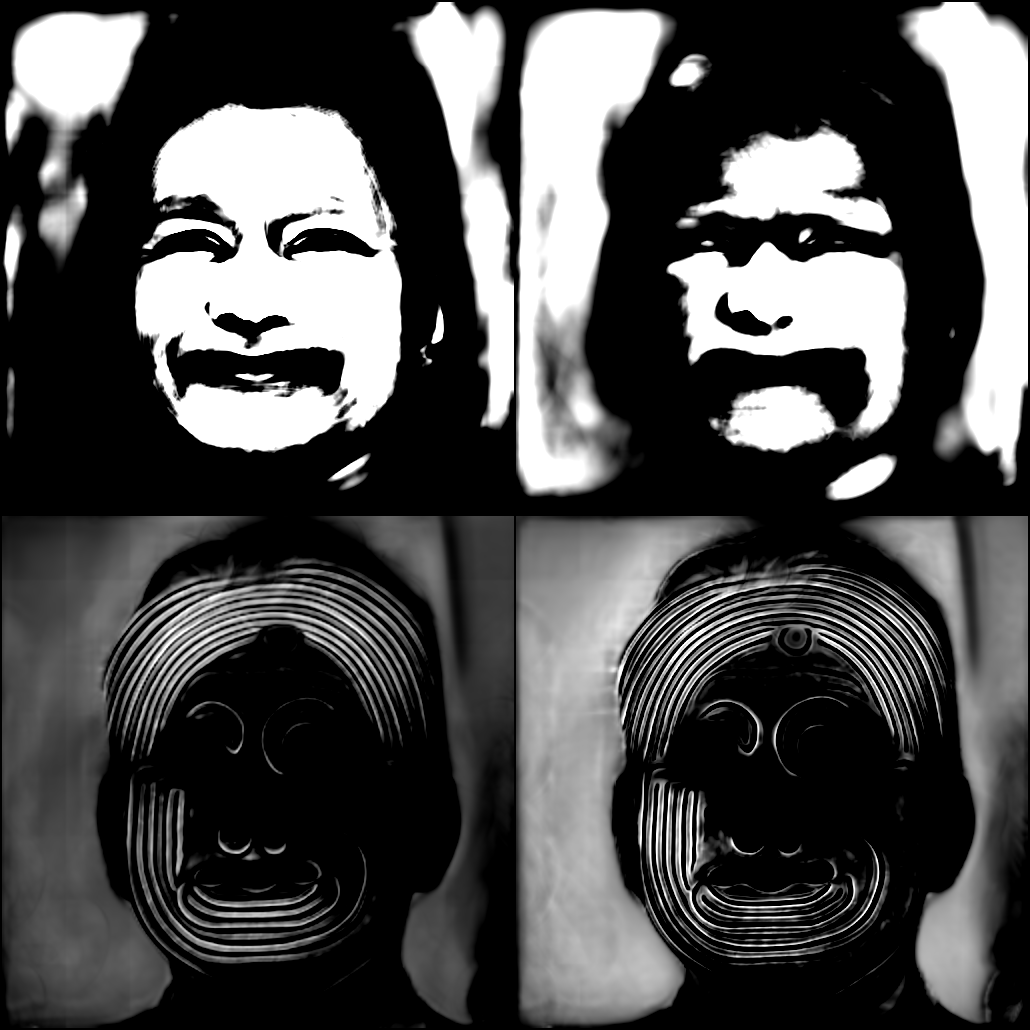}}
                \hspace{-0.50em}
            \adjustbox{trim=0 {0.50\height} {0.50\width} 0, clip}
                {\includegraphics[width=0.95\textwidth]{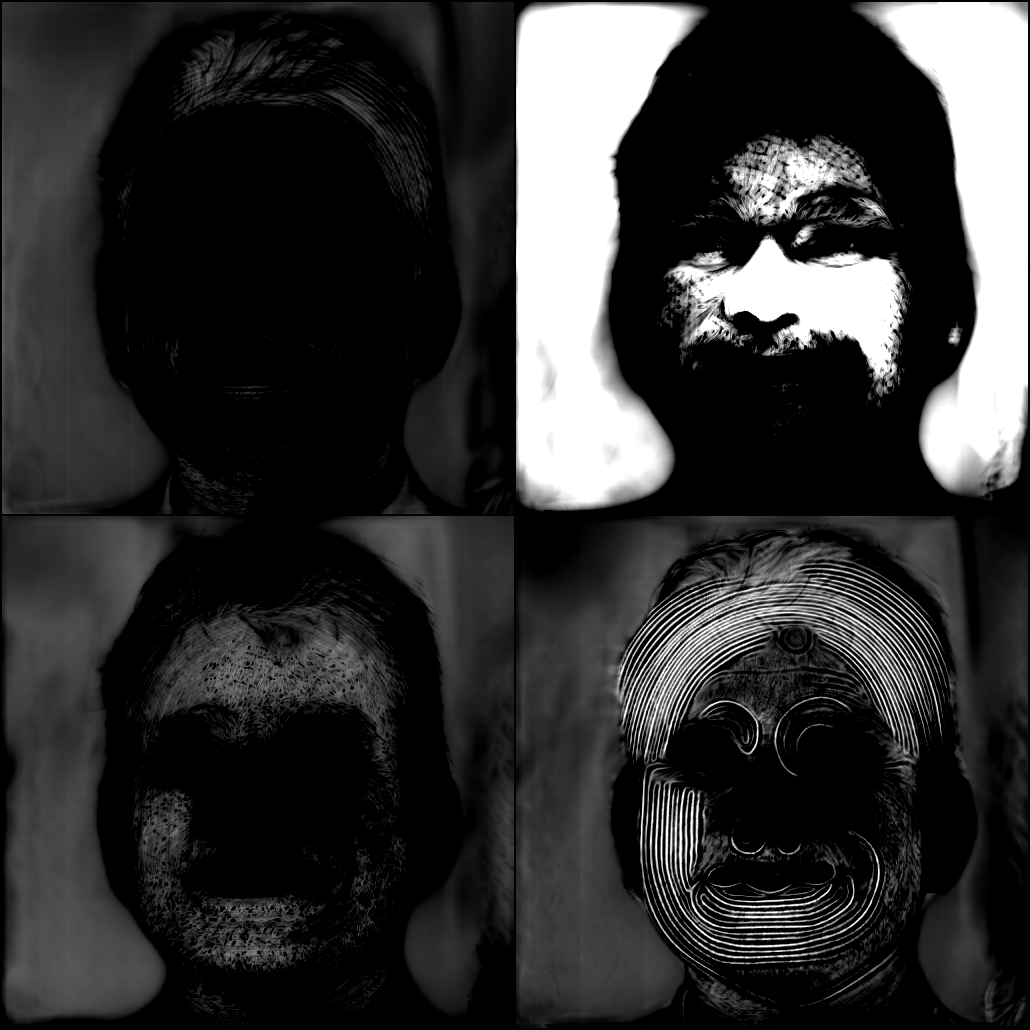}}
            \\
            %\vspace{-0.1em}
            \adjustbox{trim={0.5\width} 0 0 {0.5\height}, clip}
                {\includegraphics[width=0.95\textwidth]{Images/NA/NATLayer_70kerneldensity_channel.png}}
                \hspace{-0.50em}
            \adjustbox{trim={0.5\width} 0 0 {0.5\height}, clip}
                {\includegraphics[width=0.95\textwidth]{Images/NA/NATLayer_71kerneldensity_channel.png}}
        \end{subfigure}
        \hspace{-0.86em}
        % Layer 8 1024
        \begin{subfigure}[b]{0.18\linewidth}
            \adjustbox{trim=0 {0.50\height} {0.50\width} 0, clip}
                {\includegraphics[width=0.95\textwidth]{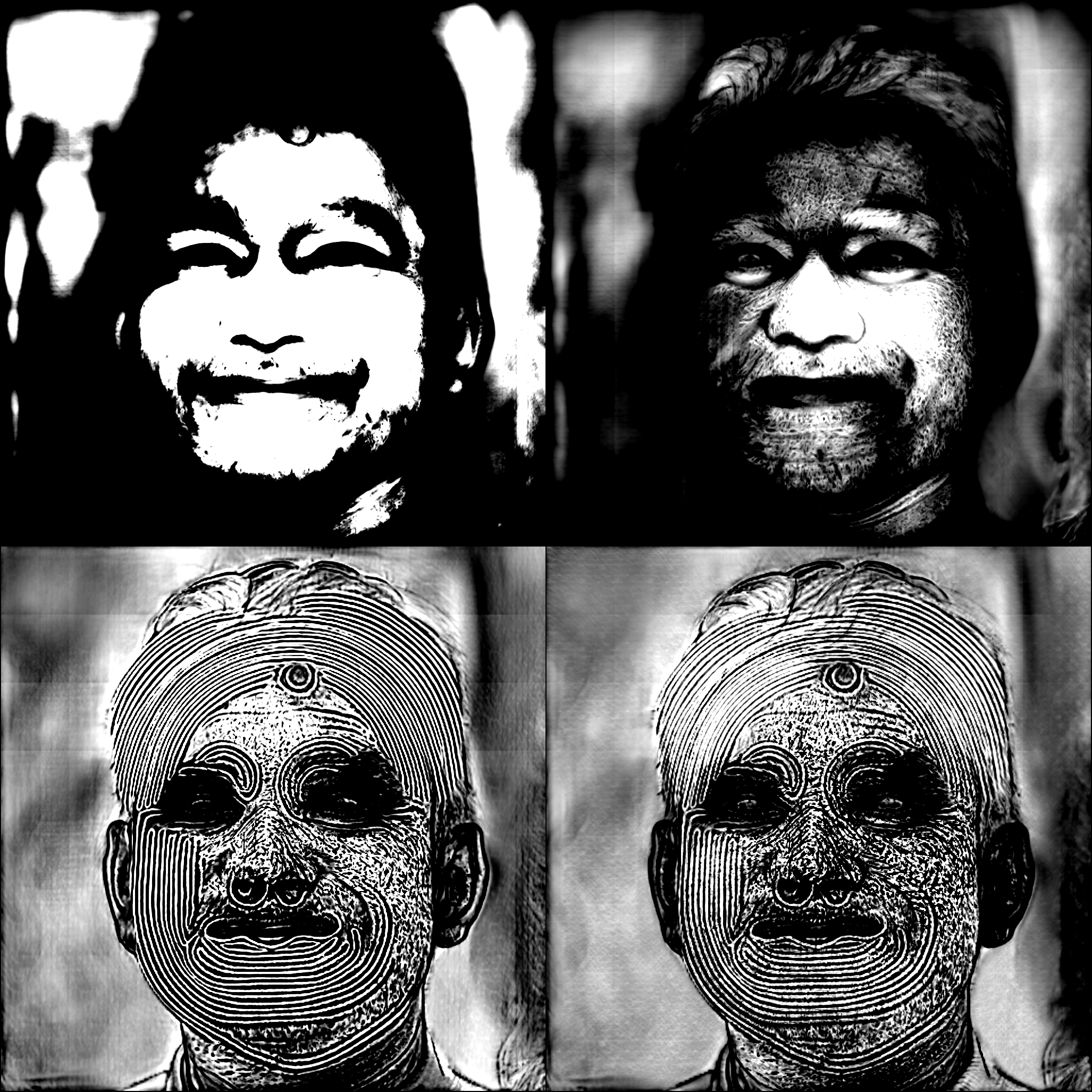}}
                \hspace{-0.8em}
            \adjustbox{trim=0 {0.50\height} {0.50\width} 0, clip}
                {\includegraphics[width=0.95\textwidth]{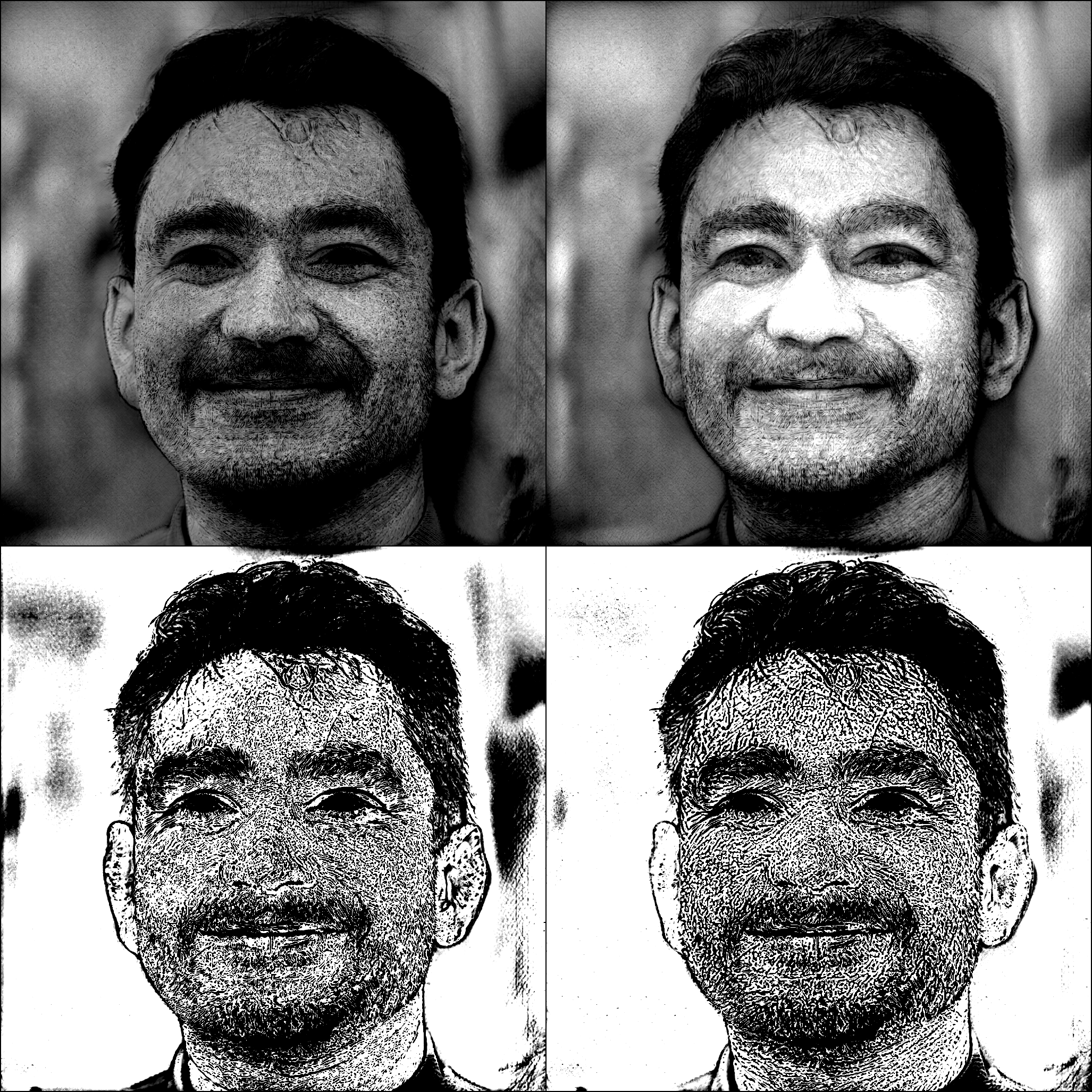}}
            \\
            %\vspace{-0.1em}
            \adjustbox{trim={0.5\width} 0 0 {0.5\height}, clip}
                {\includegraphics[width=0.95\textwidth]{Images/NA/NATLayer_80kerneldensity_channel.png}}
                \hspace{-0.8em}
            \adjustbox{trim={0.5\width} 0 0 {0.5\height}, clip}
                {\includegraphics[width=0.95\textwidth]{Images/NA/NATLayer_81kerneldensity_channel.png}}
        \end{subfigure}
        \caption{We progressively see the face form and notice the
                first head captures local features while the last head captures
                global features. Structural features appear early on while
                details are generated at higher resolutions.
        }\label{fig:progressiveAttnMaps_na}
    \end{subfigure}
    \\
    \begin{subfigure}[b]{\linewidth}
        \centering
        % Layer 2 16
        %\hspace{-0.4em}
        \begin{subfigure}[b]{0.14\linewidth}
            \adjustbox{trim=0 {0.75\height} {0.75\width} 0, clip}
                {\includegraphics[width=\textwidth]{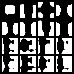}}
                \hspace{-0.49em}
            \adjustbox{trim=0 {0.75\height} {0.75\width} 0, clip}
                {\includegraphics[width=\textwidth]{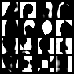}}
            \\
            \adjustbox{trim={0.75\width} 0 0 {0.75\height}, clip}
                {\includegraphics[width=\textwidth]{Images/swin/1024/StyleSwinTransformerBlock_20.png}}
                \hspace{-0.49em}
            \adjustbox{trim={0.75\width} 0 0 {0.75\height}, clip}
                {\includegraphics[width=\textwidth]{Images/swin/1024/StyleSwinTransformerBlock_21.png}}
        \end{subfigure}
        \hspace{-3.90em}
        %% Layer 3 32
        \begin{subfigure}[b]{0.17\linewidth}
            \adjustbox{trim=0 {0.75\height} {0.75\width} 0, clip}
                {\includegraphics[width=\textwidth]{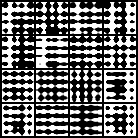}}
                \hspace{-0.50em}
            \adjustbox{trim=0 {0.75\height} {0.75\width} 0, clip}
                {\includegraphics[width=\textwidth]{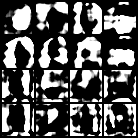}}
            \\
            \adjustbox{trim={0.75\width} 0 0 {0.75\height}, clip}
                {\includegraphics[width=\textwidth]{Images/swin/1024/StyleSwinTransformerBlock_30.png}}
                \hspace{-0.50em}
            \adjustbox{trim={0.75\width} 0 0 {0.75\height}, clip}
                {\includegraphics[width=\textwidth]{Images/swin/1024/StyleSwinTransformerBlock_31.png}}
        \end{subfigure}
        \hspace{-4.62em}
        % Layer 4 64
        \begin{subfigure}[b]{0.15\linewidth}
            \adjustbox{trim=0 {0.66\height} {0.66\width} 0, clip}
                {\includegraphics[width=\textwidth]{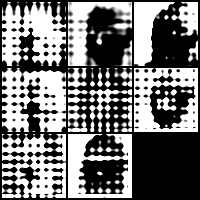}}
                \hspace{-0.50em}
            \adjustbox{trim=0 {0.66\height} {0.66\width} 0, clip}
                {\includegraphics[width=\textwidth]{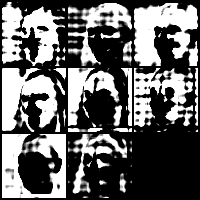}}
            \\
            \adjustbox{trim={0.33\width} 0 {0.33\width} {0.66\height}, clip}
                {\includegraphics[width=\textwidth]{Images/swin/1024/StyleSwinTransformerBlock_40.png}}
                \hspace{-0.50em}
            \adjustbox{trim={0.33\width} 0 {0.33\width} {0.66\height}, clip}
                {\includegraphics[width=\textwidth]{Images/swin/1024/StyleSwinTransformerBlock_41.png}}
        \end{subfigure}
        \hspace{-2.85em}
        % Layer 5 128
        \begin{subfigure}[b]{0.11\linewidth}
            \adjustbox{trim=0 {0.50\height} {0.50\width} 0, clip}
                {\includegraphics[width=\textwidth]{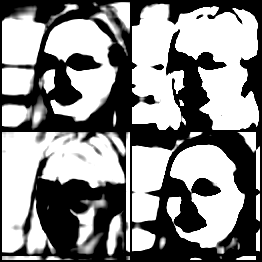}}
                \hspace{-0.50em}
            \adjustbox{trim=0 {0.50\height} {0.50\width} 0, clip}
                {\includegraphics[width=\textwidth]{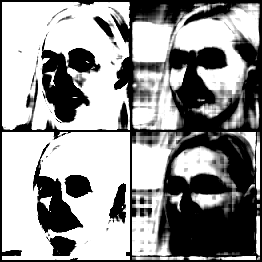}}
            \\
            \adjustbox{trim={0.5\width} 0 0 {0.5\height}, clip}
                {\includegraphics[width=\textwidth]{Images/swin/1024/StyleSwinTransformerBlock_50.png}}
                \hspace{-0.50em}
            \adjustbox{trim={0.5\width} 0 0 {0.5\height}, clip}
                {\includegraphics[width=\textwidth]{Images/swin/1024/StyleSwinTransformerBlock_51.png}}
        \end{subfigure}
        \hspace{-0.40em}
        % Layer 6 256
        \begin{subfigure}[b]{0.14\linewidth}
            \adjustbox{trim=0 {0.50\height} {0.50\width} 0, clip}
                {\includegraphics[width=0.95\textwidth]{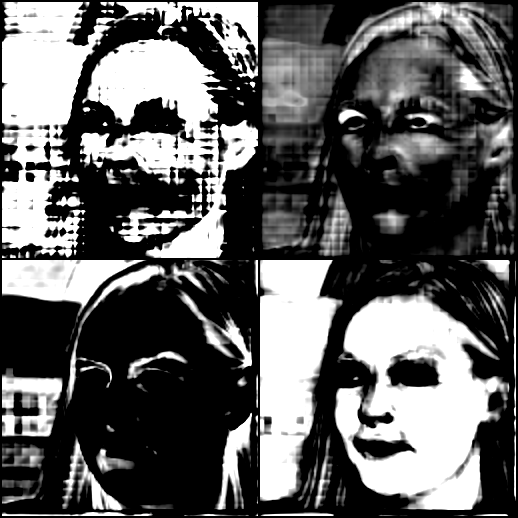}}
                \hspace{-0.50em}
            \adjustbox{trim=0 {0.50\height} {0.50\width} 0, clip}
                {\includegraphics[width=0.95\textwidth]{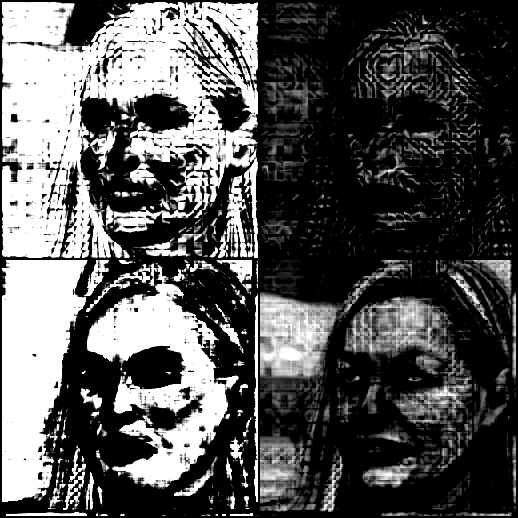}}
            \\
            \adjustbox{trim={0.5\width} 0 0 {0.5\height}, clip}
                {\includegraphics[width=0.95\textwidth]{Images/swin/1024/StyleSwinTransformerBlock_60.png}}
                \hspace{-0.50em}
            \adjustbox{trim={0.5\width} 0 0 {0.5\height}, clip}
                {\includegraphics[width=0.95\textwidth]{Images/swin/1024/StyleSwinTransformerBlock_61.png}}
        \end{subfigure}
        \hspace{-0.810em}
        % Layer 7 512
        \begin{subfigure}[b]{0.16\linewidth}
            \adjustbox{trim=0 {0.50\height} {0.50\width} 0, clip}
                {\includegraphics[width=0.95\textwidth]{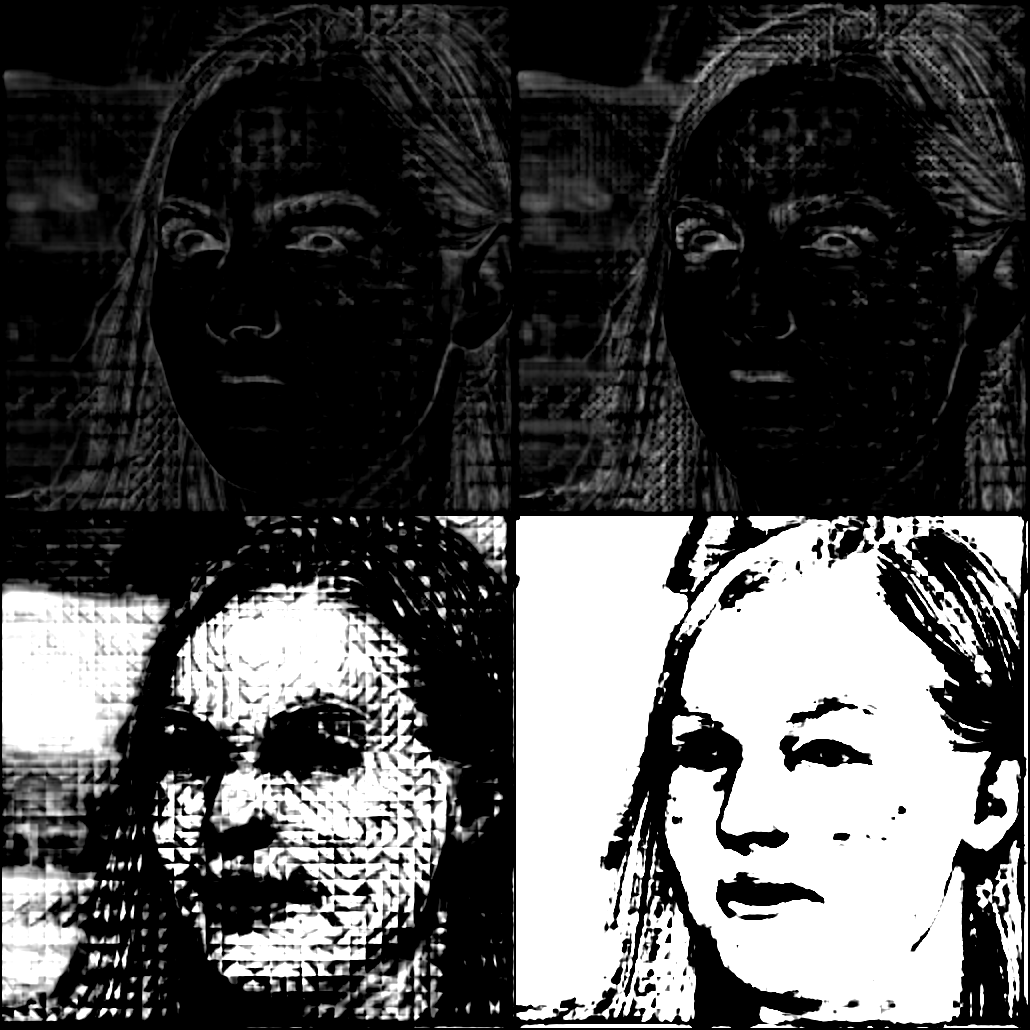}}
                \hspace{-0.50em}
            \adjustbox{trim=0 {0.50\height} {0.50\width} 0, clip}
                {\includegraphics[width=0.95\textwidth]{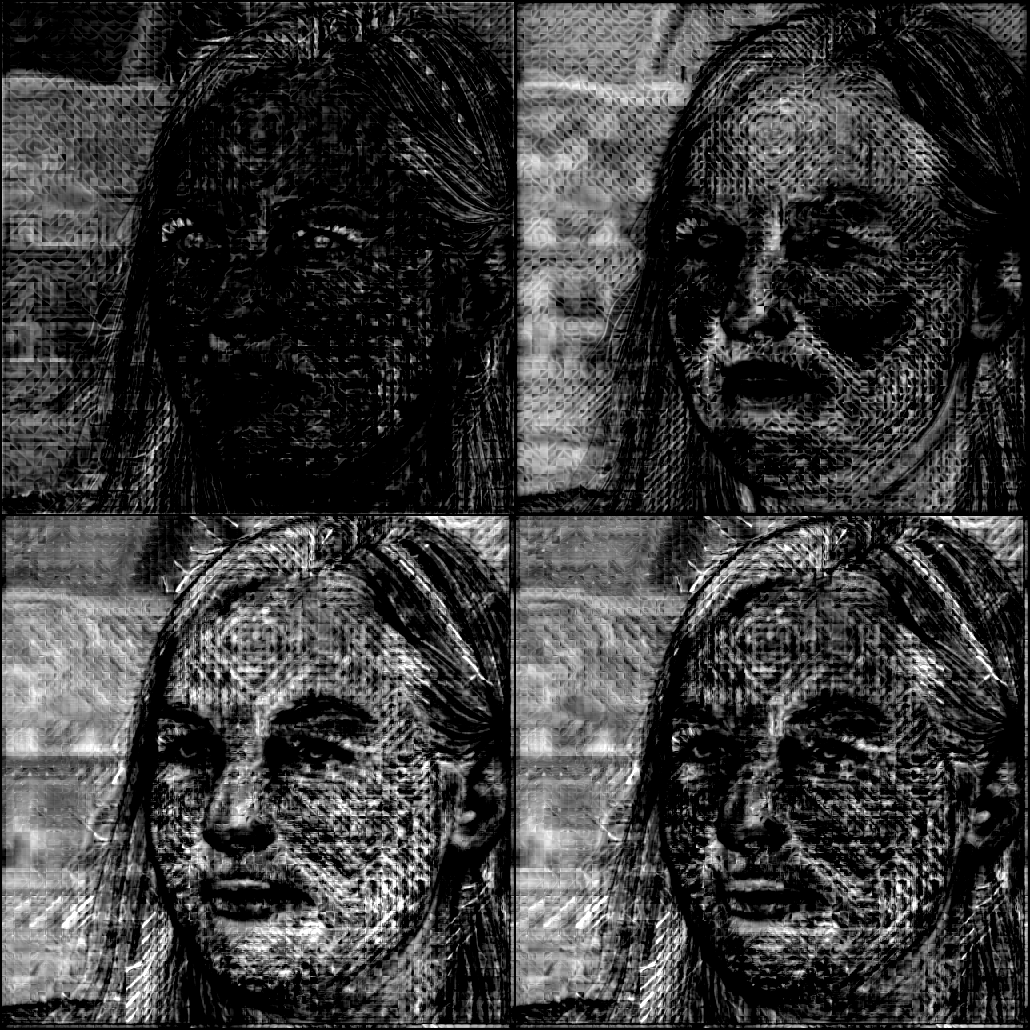}}
            \\
            %\vspace{-0.1em}
            \adjustbox{trim={0.5\width} 0 0 {0.5\height}, clip}
                {\includegraphics[width=0.95\textwidth]{Images/swin/1024/StyleSwinTransformerBlock_70.png}}
                \hspace{-0.50em}
            \adjustbox{trim={0.5\width} 0 0 {0.5\height}, clip}
                {\includegraphics[width=0.95\textwidth]{Images/swin/1024/StyleSwinTransformerBlock_71.png}}
        \end{subfigure}
        \hspace{-0.86em}
        % Layer 8 1024
        \begin{subfigure}[b]{0.18\linewidth}
            \adjustbox{trim=0 {0.50\height} {0.50\width} 0, clip}
                {\includegraphics[width=0.95\textwidth]{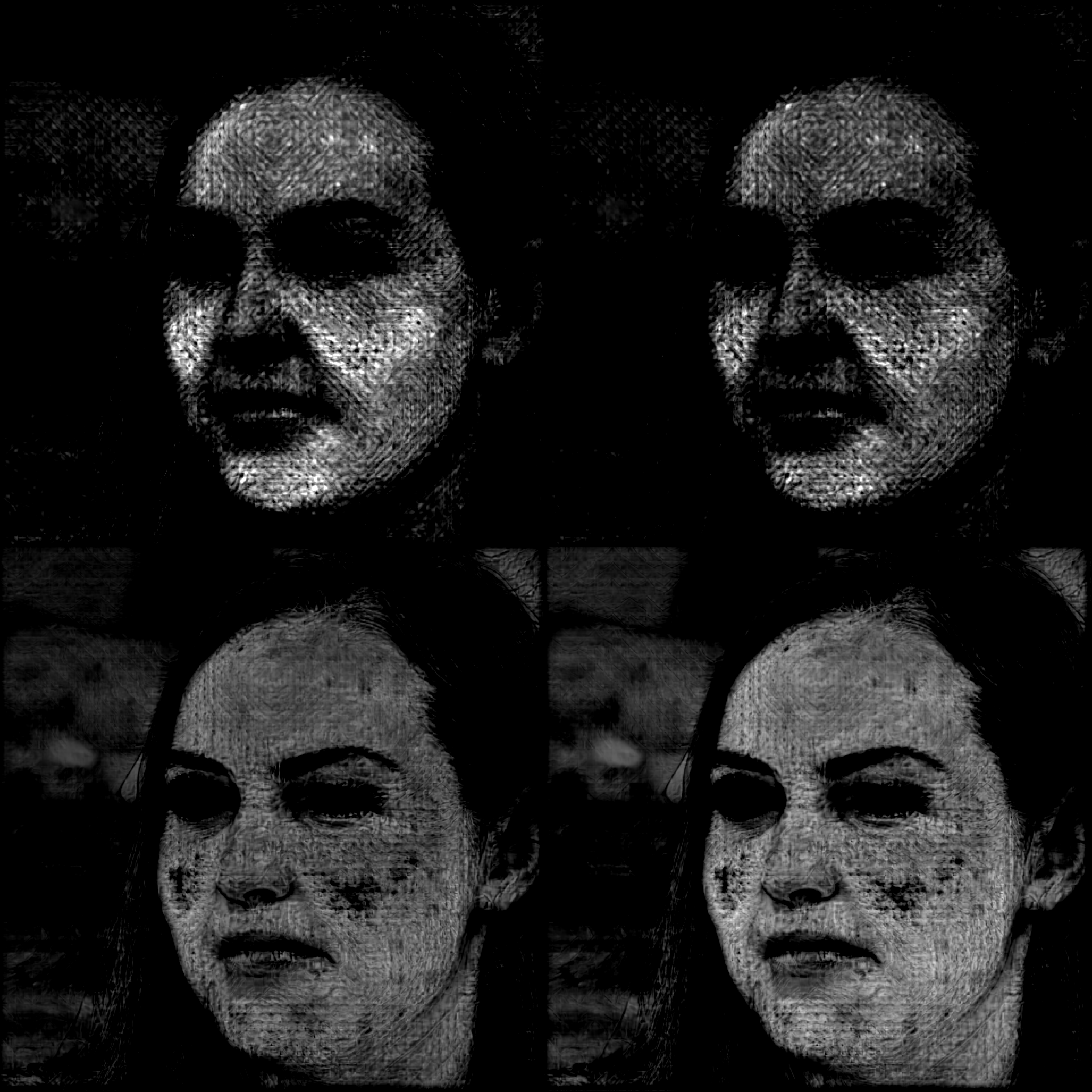}}
                \hspace{-0.8em}
            \adjustbox{trim=0 {0.50\height} {0.50\width} 0, clip}
                {\includegraphics[width=0.95\textwidth]{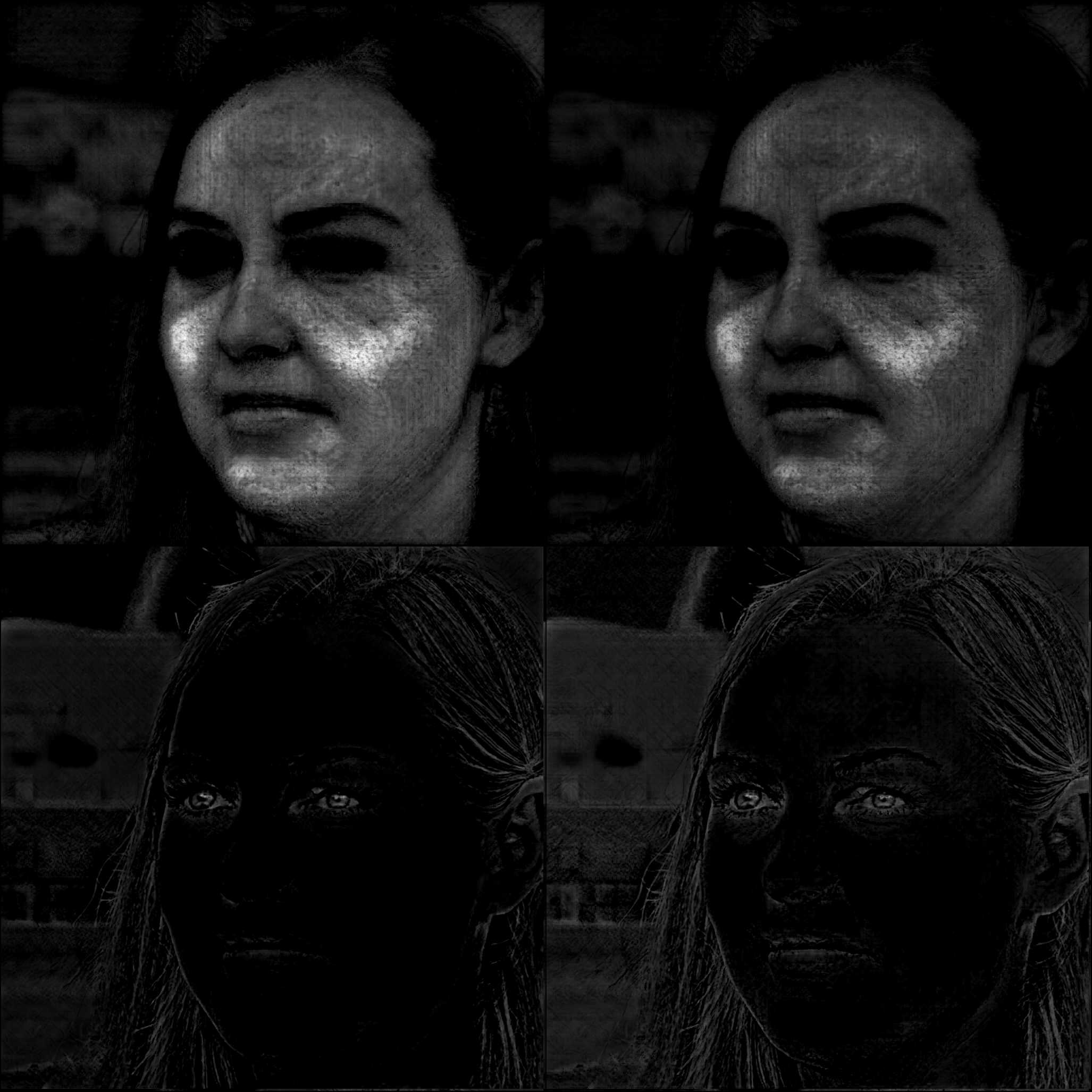}}
            \\
            %\vspace{-0.1em}
            \adjustbox{trim={0.5\width} 0 0 {0.5\height}, clip}
                {\includegraphics[width=0.95\textwidth]{Images/swin/1024/StyleSwinTransformerBlock_80.png}}
                \hspace{-0.8em}
            \adjustbox{trim={0.5\width} 0 0 {0.5\height}, clip}
                {\includegraphics[width=0.95\textwidth]{Images/swin/1024/StyleSwinTransformerBlock_81.png}}
        \end{subfigure}
        \caption{Low resolutions show decoherence and artifacts are not removed
            in progressive resolutions. We do not observe strong
                differentiation between local and global features. These
                maps explain the blocking artifacts discussed in section 1
                of StyleSwin~\cite{Zhang_2022_CVPR}.
        }\label{fig:progressiveAttnMaps_swin}
    \end{subfigure}
    \caption{Visualization of the first and last attention head progressing
        through StyleNAT. We start at a resoluton of $16\times16$ and grow to 
            $1024\times1024$. We generate 50 samples from each network and
            choose the best image from the sample to make comparisons as fair as 
            possible. The top row shows the first attention head, with 2
            transformers per resolution level. The bottom row shows the last
            attention head. \cref{fig:progressiveAttnMaps_na} visualizes for
            StyleNAT (\textbf{ours}) and \cref{fig:progressiveAttnMaps_swin}
        follows StyleSwin~\cite{Zhang_2022_CVPR}.
    }\label{fig:progressiveAttnMaps}
\end{figure*}

%% file: sec/2_rw/StyleGAN.tex
\subsection{StyleGAN Based Networks}\label{sec:rw-sg-architectures}

%\subsubsection{StyleGAN}\label{sec:rw-stylegan}

Style-based GANs~\cite{Karras_2019_CVPR,Karras_2020_CVPR,NEURIPS2020_8d30aa96,NEURIPS2021_076ccd93} have been the \emph{de facto} image generating architecture
over the last decade, only recently replaced by diffusion
models~\cite{NEURIPS2020_4c5bcfec}.
These networks operate by progressively growing noise from a small tensor into
the desired image resolution~\cite{karras2018progressive}.
They contain two sub-networks, the \emph{style}-network (mapping) and the
\emph{generator}-network (synthesis).\footnote{Papers refer to these as style and generator but in code they are
often called mapping and synthesis.}
We visualize this for StyleNAT and StyleSwin in~\Cref{fig:progressiveAttnMaps}.

The \emph{style} sub-network allows for control over the image synthesis, allowing
for the semantic image editing, similar to how text to image generators
work~\cite{esser2024scalingrectifiedflowtransformers,imagenteamgoogle2024imagen3,betker2023improving}.
This sub-network is a linear network that injects information just prior to each
convolution~\cite{Karras_2019_CVPR}, using an adaptive instance normalization
(AdaIN)~\cite{huang2017arbitrarystyletransferrealtime,dumoulin2017a}.
A StyleGAN's \emph{generator} sub-network is initialized from a small tensor, e.g. 
$512\times4\times4$, while a diffusion model~\cite{NEURIPS2020_4c5bcfec} 
uses an input the same size as the target resolution.
Latent diffusion models~\cite{Rombach_2022_CVPR} utilize a backbone based on a 
denoising autoencoder~\cite{vincent10a} with residual layers, forming a
U-Net~\cite{ronneberger2015unetconvolutionalnetworksbiomedical}.
The generator in a StyleGAN network does not have the encoder network,
requiring the generator must must perform more information gain at each
upsampling step.

%% file: sec/2_rw/Attention.tex
\subsection{Attention Mechanisms}\label{sec:rw-attention}

Attempts to integrate attention mechanisms into
GANs~\cite{NEURIPS2021_98dce83d,pmlr-v97-zhang19d} have met with mixed success,
especially compared to other
models~\cite{song2021denoising,Sukthanker_2022_CVPR,lu2023compensationsamplingimprovedconvergence}.
Convolutional models benefit from the progressively growing architecture, as
`global' features can be converted to `local' features at low resolutions.
Unfortunately, a convolution's local inductive biases often result in generation
issues, such as heterochromia (eyes with different colors), differing pupil
sizes, or misaligned teeth when generating
faces~\cite{Karras_2020_CVPR,goetschalckx2019ganalyze}, providing motivation for
introducing transformers which capture global features.
At high resolutions the memory requirements make integrating a
transformer that attends to the pixel space intractable.
While works like
FlashAttention~\cite{dao2022flashattention,dao2023flashattention} reduce
complexity, this is insufficient.
Instead, these networks use channel based attention~\cite{pmlr-v97-zhang19d}, to
make them computationally tractable.
Alternatively, restricted attention mechanisms have been proposed to reduce the
computational complexity of attention.

While transformer based GANs have shown some advantages the largest gains have
frequently been by introducing more parameters~\cite{NEURIPS2021_98dce83d}.
This provides additional flexibility to networks, allowing for smoother latent
representations and thus easier learning~\cite{schaeffer2023emergent,pmlr-v119-zhang20h}.

%\subsubsection{Swin Transformer}\label{sec:rw-swin}
%\subsection{SASA, SWSA, WSA: Restricted Attention}\label{sec:rw-swin}
\subsection{Restricted Attention: SASA, SWSA, WSA}\label{sec:rw-swin}

Stand-alone Self-Attention (SASA)~\cite{NEURIPS2019_3416a75f} was one of the
first to apply a local windowed attention to vision models, restricting the
attention calculation to a localized window. 
Later, Liu~\etal proposed Window Self-Attention (WSA) and Shifted-Window Self-Attention
(SWSA)~\cite{Liu_2021_ICCV,Liu_2022_CVPR}, creating the \textbf{S}hifted \textbf{Win}dow Transformer.
Like others~\cite{beltagy2020longformerlongdocumenttransformer,NEURIPS2019_3416a75f,Vaswani_2021_CVPR},
Swin uses a hierarchical method where images are initially divided up (WSA) and at
the next level these are shifted (SASA), wrapping around borders.
This has the advantage of incorporating global features and local features,
given enough permutations, finding success on many vision applications
\cite{Liu_2022_CVPR_video,10.1007/978-3-031-25066-8_9,9686686}.
%Swin found success on vision classification as well as proved adept in
%segmentation and detection.
%This led to a plethora of other Swin based vision
%methods~\cite{Liu_2022_CVPR_video,10.1007/978-3-031-25066-8_9,9686686}

Zhang~\etal proposed StyleSwin~\cite{Zhang_2022_CVPR}, which closely followed the
StyleGAN architecture, replacing the main convolution layer with a Swin
Transformer.
Similar to Zhao~\etal~\cite{NEURIPS2021_98dce83d} they splits the
attention mechanism across two different sets of heads and later concatenates
them.
This allowed StyleSwin to incorporate both WSA and SWSA into a single attention
mechanism, as opposed to different levels like Swin.
They hypothesize that this allows the attention mechanism to incorporate both 
global and local information, due to the overlapping partitions.
StyleSwin set a new state of the art for transformer based GANs, but not GANs
overall. 

\subsubsection{Neighborhood Attention}\label{sec:rw-nat}

Despite Swin and SASA's successes, SASA doesn't properly incorporate features at
the edges of images while Swin's partitioning introduces boundaries that mean
they are not equivariant to translation and rotations~\cite{Hassani_2023_CVPR}.
This can lead to blocking effects, as seen in \Cref{fig:progressiveAttnMaps} and
illustrated in the appendix of Hassani~\etal~\cite{Hassani_2023_CVPR}.
Hassani~\etal proposed Neighborhood
Attention~\cite{Hassani_2023_CVPR,hassani2022dilated,hassani2024fasterneighborhoodattentionreducing} to
resolve these issues, creating a mechanism that is both translationally
and rotationally equivariant.
When the window size is equal to the image size, it
is equivalent to the standard attention mechanism, whereas methods like SASA
fail at the edges.
Neighborhood can be expressed as follows

\begin{align}\label{eq:rw-nat}
    \mathbf{A}_i^k &= \begin{bmatrix}
            Q_i
            K^T_{\rho_0(i)}
            + B_{i,\rho_0(i)}\\
            \vdots\\
            Q_i
            K^T_{\rho_k(i)}
            + B_{i,\rho_k(i)}\\
    \end{bmatrix}
            \\
    \label{eq:rw-nat2}
    \mathbf{V}_i^k &= \begin{bmatrix}
        V_{\rho_0(i)}^T, \cdots, V_{\rho_k(i)}^T
    \end{bmatrix}^T
    \\
    \label{eq:rw-nat3}
    NA_{k(i)} &= \text{Softmax}
        \left(
            \frac{\mathbf{A}_i^k}{\sqrt{d}}
        \right)
        \mathbf{V}_i^k
\end{align}

$\mathbf{A}_i^k$ is the attention weights for the $i^\text{th}$ input with
neighborhood size $k$ (kernel).
Similarly, $\mathbf{V}_i^k$ are the neighboring values with rows being $k$
nearest neighboring projections for the $i^\text{th}$ input.
$\mathbf{Q},\mathbf{K},\mathbf{V}$ are used in the traditional 
sense, $\mathbf{B}(i,j)$ is the relative positional biases.
$\rho_j(i)$ is the 
$j^\text{th}$ nearest neighborhood of the $i^\text{th}$ input query, 
and $\sqrt{d}$ is the softmax temperature~\cite{dabre-fujita-2021-investigating}.

%% file: sec/3_methods.tex
\section{Methodology}\label{sec:methodology}

Prior works have demonstrated the utility of restricted attention mechanisms,
allowing them to achieve state of the art or highly competitive results on tasks 
such as instance Classification, Detection, and
Segmentation~\cite{Jain_2023_CVPR,zhan2022triocc}.
NA has also shown success in reducing computational the complexity of diffusion
models~\cite{crowson2024scalable}.
But these have not found success in models like GANs, which are less stable.

We hypothesize that we can increase the information gain of architectures like
NA by liberating each attention head to attend to independent receptive fields.
By doing this, we can allow a \emph{single} attention mechanism to incorporate
both local information \emph{and} global information.

\input{Includes/Figures/architecture}

This new variation can be expressed as

\begin{align}\label{eq:mth-hnat}
    \mathbf{A}_{i,h}^k &= \begin{bmatrix}
        Q_{i,h(w,d)}
            K^T_{\rho_0(i),h(w,d)}
            + B_{i,\rho_0(i),w}\\
            \vdots\\
            Q_{i,h(w,d)}
            K^T_{\rho_k(i),h(w,d)}
            + B_{i,\rho_k(i),w}\\
    \end{bmatrix}
            \\
    \label{eq:mth-hnat2}
    \mathbf{V}_{(i,h)}^k &= \begin{bmatrix}
        V_{\rho_0(i),h(w,d)}^T, \cdots, V_{\rho_k(i),h(w,d)}^T
    \end{bmatrix}^T
    \\
    NA_{k(i)} &= \text{Softmax}
        \left(
            \frac{\mathbf{A}_{i,h}^k}{\sqrt{d}}
        \right)
        \mathbf{V}_{i,h}^k
\end{align}

%The difference from~\Cref{eq:rw-nat,eq:rw-nat2} is that we introduce $w,d$ which represent
%the windows and dilations corresponding to specific heads ($h$), respectively.
The equations are quite similar to those in
\cref{eq:rw-nat,eq:rw-nat2,eq:rw-nat3}, the key difference being that attention
heads, $h$, of the
$\mathbf{Q}$, $\mathbf{K}$, $\mathbf{V}$, and $B$ parameters are dependent
on the window size, $w$, and dilation, $d$.\footnote{The relational bias, $B$,
is not dependent on dilation.}
In NA~\cite{Hassani_2023_CVPR} and Dilated NA (DiNA)~\cite{hassani2022dilated}
these are fixed per attention, but we remove this requirement.
To perform this variant, dubbed \emph{Hydra-NA}, we perform the same algorithm
on a \emph{per-head} basis, concatenating the result prior to being processed by
the transformer's MLP layer.
Like the mythical many headed hydra, this attention mechanism can concentrate on
independent locations and process the result together.
By increasing the information gain through Hydra-NA we can achieve higher
performance without the scaling that other methods like StyleGAN-XL require~\cite{sauer2022stylegan}.
The Python code to implement Hydra-NA is included in 
\Cref{appdx:hydracode}

Our method has no increase in the number of network parameters nor increases the 
MACs operations, compared to NA, and has
minimal computational increase, mostly due to the partitioning and lack of
a GPU kernel to execute these in parallel.
A proper CUDA kernel will result in increased throughput and reduce training
time.

%% file: Includes/Figures/architecture.tex
\begin{figure*}[htbp!]
    \centering
    % [trim={left bottom right top},clip]
    \includegraphics[trim={0.4cm 0.12cm 0.4cm 1.45cm},clip,width=0.80\textwidth]{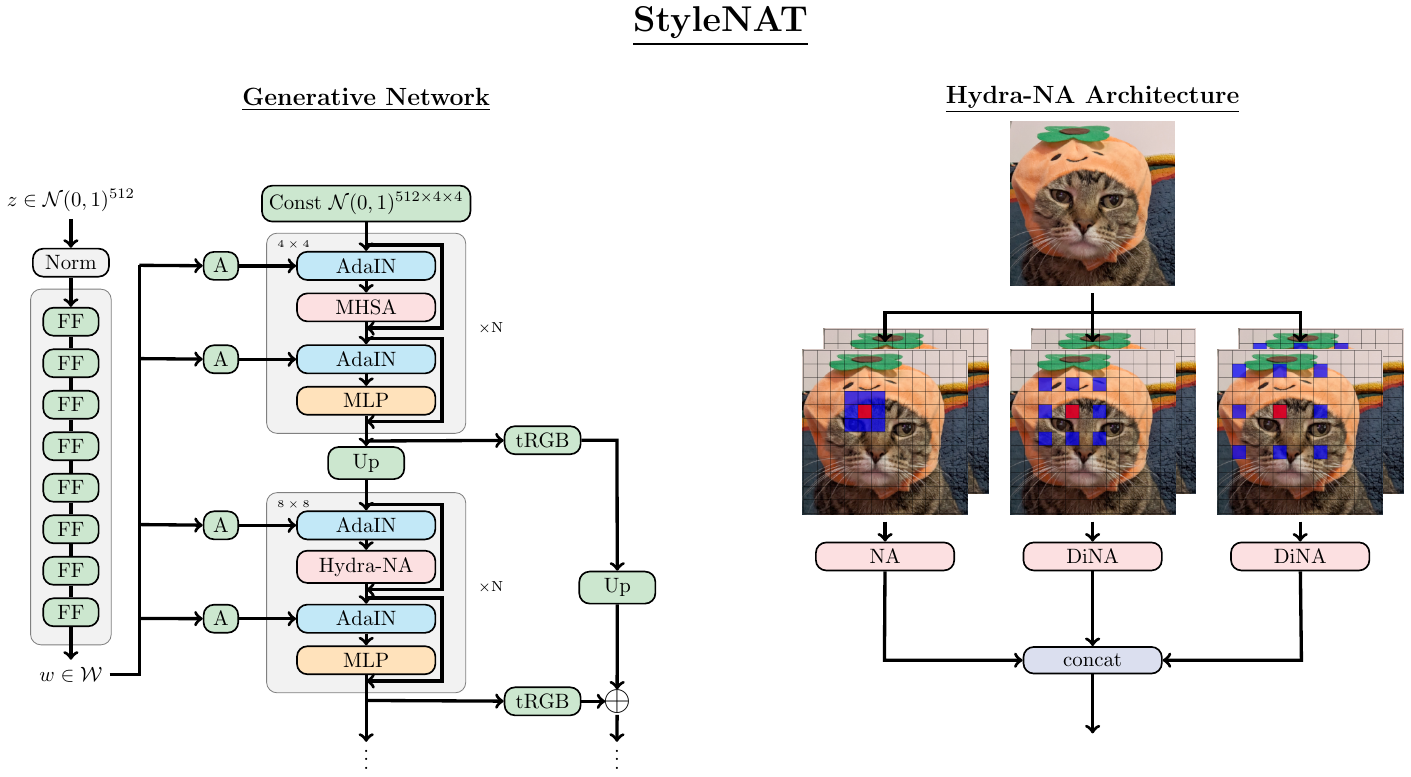}
    \caption{StyleNAT. We follow the design of StyleSwin, which closely follows 
    StyleGAN. The style-network shown at the left most, injecting information
    into the generator. N represents the number of resolutions per layer, fixed
    at 2 for all included experiments.
    We use Hydra-NA layers in place of convolutions at all resolutions except
    the smallest ($4\times4$) wherein a MHSA layer is used.
    Our Hydra-NA formulation is depicted to the right, illustrating how the
    input is partitioned across attention heads. This illustration would
    correspond to an attention layer with 6 heads, each with a kernel size of 3.
    The first 2 heads have a dilation of 1, followed by 2 heads with dilation 2,
    and 2 more heads with dilation 3.
    This gives our model additional flexibility over the receptive field while 
    maintaining computational efficiency. 
    The code for Hydra-NA is included \Cref{appdx:hydracode}.
    }\label{fig:method-architecture}
\end{figure*}

%% file: sec/4_experiments.tex
\section{Experiments}\label{sec:experiments}

Due to limitations in computational budgets we carefully design our experiments
to ensure we can evidence our hypothesis without brute force experimentation.
Our experiments are carefully constructed to ensure that we isolate experimental
variables and ensure the results are explainable by Hydra-NA.
Our total computational budget is lower than that of StyleGAN 3's ``early
exploration'' phase~\cite{NEURIPS2021_076ccd93}.

We focus on utilizing a GAN architecture for our experiments, due to their
difficulty of incorporating transformer architectures.
A StyleGAN is utilized both for familiarity and due to the increased
computational pressure required by the generative (decoding) network.
We leverage the fact that StyleGANs are well studied so that researchers are
better able to interpret results.
More specifically, we base our model off of StyleSwin~\cite{Zhang_2022_CVPR} as
it is the closest framework to our own. 

We purposefully do not explore other network modifications and minimize 
hyper-parameter optimization to we isolate experimental variables.
%and ensure that our results can be unambiguiously contributed to Hydra-NA.
Thus we will follow the same training procedure of StyleSwin, using the Two 
Time-Scale Update Rule (TTUR)~\cite{NIPS2017_8a1d6947}, a discriminator 
learning rate of $2\times10^{-4}$, utilize Balanced Consistency Regularization
(bCR)~\cite{zhao2021improved}, r1 regularization~\cite{pmlr-v80-mescheder18a}, 
and have a learning rate decay.
We only tune the start of the learning rate decay based on when we observed
saturation in our training.

We have open sourced our network, including all information used for training,
inference, calculating throughput, and calculating the attention maps.
We also embed all training parameters into our model checkpoints for redundancy
and clarity.
All experiments are performed on NVIDIA A100s.
FFHQ-256 and LSUN Church experiments are performed on a single A100 node. 
For FFHQ-1024 we only perform a \emph{single} training run, \emph{from scratch},
utilizing 4 A100 nodes.
We do not initialize from the lower resolution model to demonstrate the
stability and effectiveness of the architecture's ability to coherently capture
local and global features.
\emph{We do not fully train} this network due to our limited computational
budget and did not observe network saturation when we stopped training.
Similarly, this limits our dataset and architectural exploration, but we believe 
our experimental methodology allows us to evidence our research hypothesis.

We choose Flickr-Face-HQ Dataset (FFHQ)~\cite{Karras_2019_CVPR} 
and Large Scale image database (LSUN)~\cite{yu15lsun} Church.
These are the most common datasets with these architectures, making for easier
comparisons.
There are many limitations to evaluating image
quality~\cite{NEURIPS2019_0234c510,kynkaanniemi2023the,Parmar_2022_CVPR,pmlr-v119-naeem20a,NEURIPS2019_65699726,morozov2021on}, but we use the standard
Frechet Inception Distance (FID)~\cite{NIPS2017_8a1d6947}.
Evaluating generative models is notoriously
difficult~\cite{theis2016noteevaluationgenerativemodels,bińkowski2018demystifying}.
Stein~\etal~\cite{stein2023exposing} demonstrates the limitations of these
methods and conducted the largest human evauation study, comparing many
different metrics and demonstrates that these do not strongly correlate with
human preference.

\input{sec/3_methods/ffhq}

\input{sec/3_methods/church}

% Tables
\input{Includes/Tables/church_ablation}
\input{sec/3_methods/analysis}

%% file: sec/3_methods/ffhq.tex
\subsection{FFHQ}\label{sec:exp-ffhq}

Due to humans' innate ability to recognize faces and their strong aptitude in
recognizing even minor flaws, we focus on this dataset for evaluation.
With the aforementioned limitations of mathematical metrics it becomes
imparitive that we utilize a dataset in which the results can be deeply
scrutinized.
While FFHQ contains only a single class, it includes high levels of complexity
and details.
To aid in evaluation, we highlight common locations for generative failures.
For local features it failures often occur in hair, teeth, necks, ears, and at
the corners of eyes, especially when faces with glasses are generated.
Global decoherence is easier to spot due to lack of facial symmetry, with issues
like the size of pupils, light reflections, as well as symmetry of earrings or
eyeware.
Issues like heterochromia can indicate both global or local decoherence, with
differing eye colors indicating failure to capture global symmetry while
multiple colors being found in a single iris represent local decoherence.
Jewlrey and headware (hats, glasses, etc) present particular challenges as these
features occur at far lower rates within the data, making them harder to learn.

\input{Includes/Figures/ffhq256_samples}

We believe that the disadvantage of a human's aptitude for visual evaluation of
faces significantly outweighs the benefits that a single class provides to the
model.
To offset the limitations in mathematical evaluation, we further enable human 
evaluation by embedding all imagry at full resolution within
the PDF and encourage readers to zoom in and carefully study the generated
images of our work and that of others. 
We also include a deeper analysis and comparison in
\Cref{app:metrics,app:visual_analysis}.

For our initial experiments we resize the images to $256\times256$ and train for
940k iterations with a batch size of 8 (per GPU), for a total of 60.2 million
images.
Our LR decay begins at 740k iterations and our window size was 7.
The initial resolution level being $4\times4$ we used standard MHSA as 
this cannot be properly windowed.
For the second level, $8\times8$, we utilize no dilation, as this would go out
of bounds.
Otherwise we kept our window size of 7 and had half the heads as a dense kernel
and the other as the maximal dilation size: e.g. at resolution $16\times16$ has 
a dilation of 2, and thus its window size was effectively 14.
By keeping the window size at 7 we ensure our network is close to that of
StyleSwin.
By increasing the kernel size we may unfairly advantage our architecture.

\input{Includes/Figures/church_samples}
\input{Includes/Tables/ffhq-ablation}
\input{Includes/Tables/full_compare}

\Cref{tab:ffhq256-ablation} shows our ablation, testing the effectiveness of our
architecture methodology.
First by interchanging 
the StyleSwin's split-head transformer for NAT (FID=$2.81$), 
which does not have any split heads ($+$NA, $FID=2.74$).
This demonstrates the advantage provided by NA, isolating advantages of NA over
Swin, serving as a baseline comparison to our method.
We then introduce Hydra-NA, partitioning heads into two groups (like StyleSwin) 
that correspond to dense local receptive fields and the most sparse global 
($+$Hydra, FID=$2.24$).
This allows the network to incorporate local and global information, 
demonstrating the effectiveness of our architectural change.
This step represents the largest increase in performance.
StyleSwin noted that horizontal Flips did not increase performance within their
network, so we introduce them to ours to test the equivariant properties of NA
and due to transformer's preference of augmentation ($+$Flips, FID=$2.05$).
Finally we test partitioning on 4 heads, utilizing a progressive dilation,
incorporating intermediate features ($+$Prog Di ($4$), $FID=2.55$).
The result decreases, but demonstrates that even improper configuration benefits
generation.
Our result represents the state of the art performance due to architectural
changes.
%A concurrent work by Takida~\etal~\cite{takida2023san} is able to outperform us,
%but utilizes a novel training method, utilizes significantly more parameters,
%and has lower throughput (based on StyleGAN-XL).

For FFHQ-1024 we retrain our model from scratch with near identical settings to
our 256 experiments.
\emph{Only a single training run was performed for this resolution.}
We decrease the batch size to 4, increase to 4 A100 nodes, and train the model 
for 900k iterations (28.8M images).
We observe an initial saturation around 500k iterations and pick that as the
start to our learning rate decay.
We early stop the training with a final FID score of 4.17.

%% file: Includes/Figures/ffhq256_samples.tex
\begin{figure}
    \centering
    \includegraphics[width=0.43\textwidth]{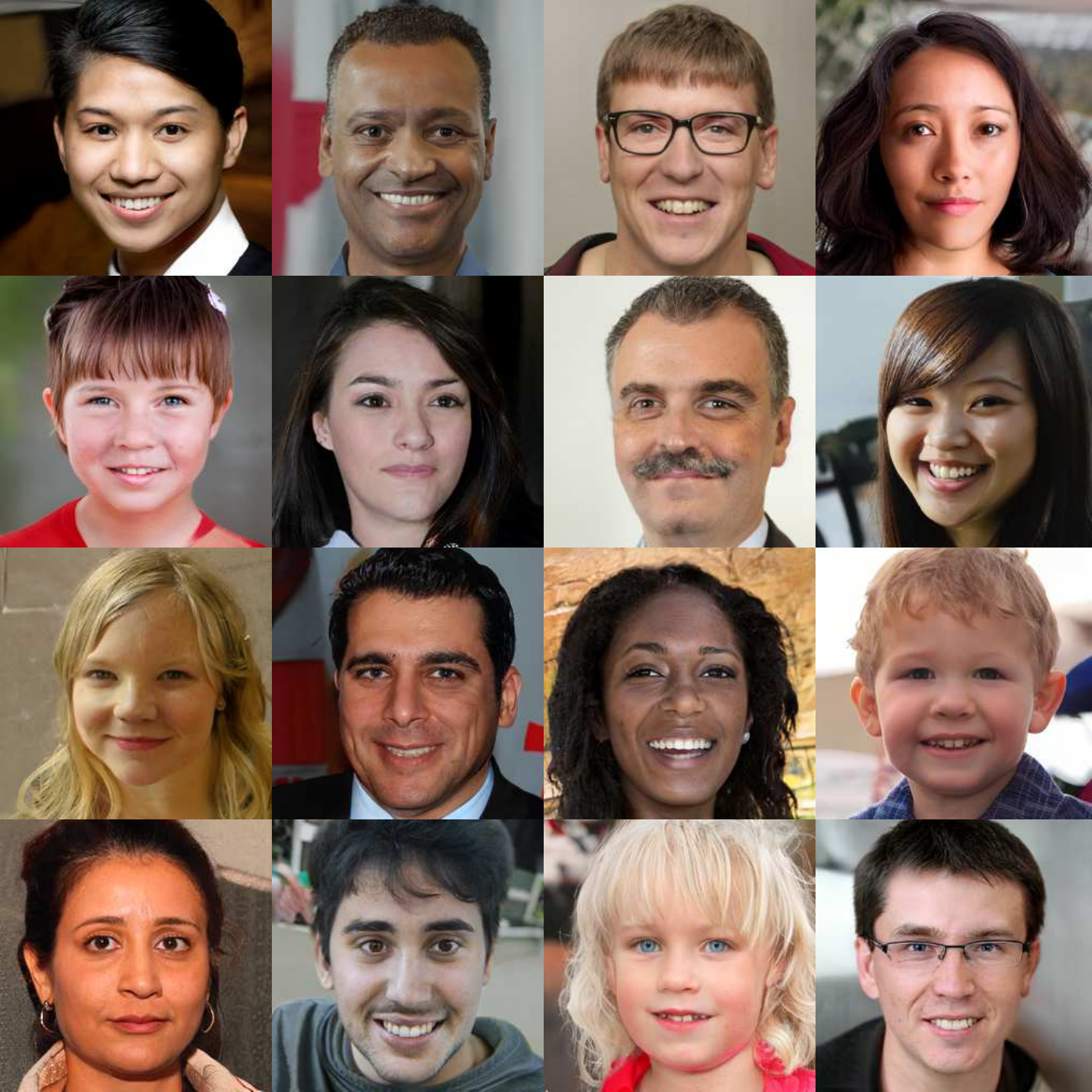}
    % Sampling does not use truncation or any other tricks to increase fidelity.}
    \caption{Samples from FFHQ-256 with FID50k of 2.05.
        }\label{fig:exp-ffhq256-samples}
\end{figure}

%% file: Includes/Figures/church_samples.tex
\begin{figure}
    \centering
    \includegraphics[width=0.43\textwidth]{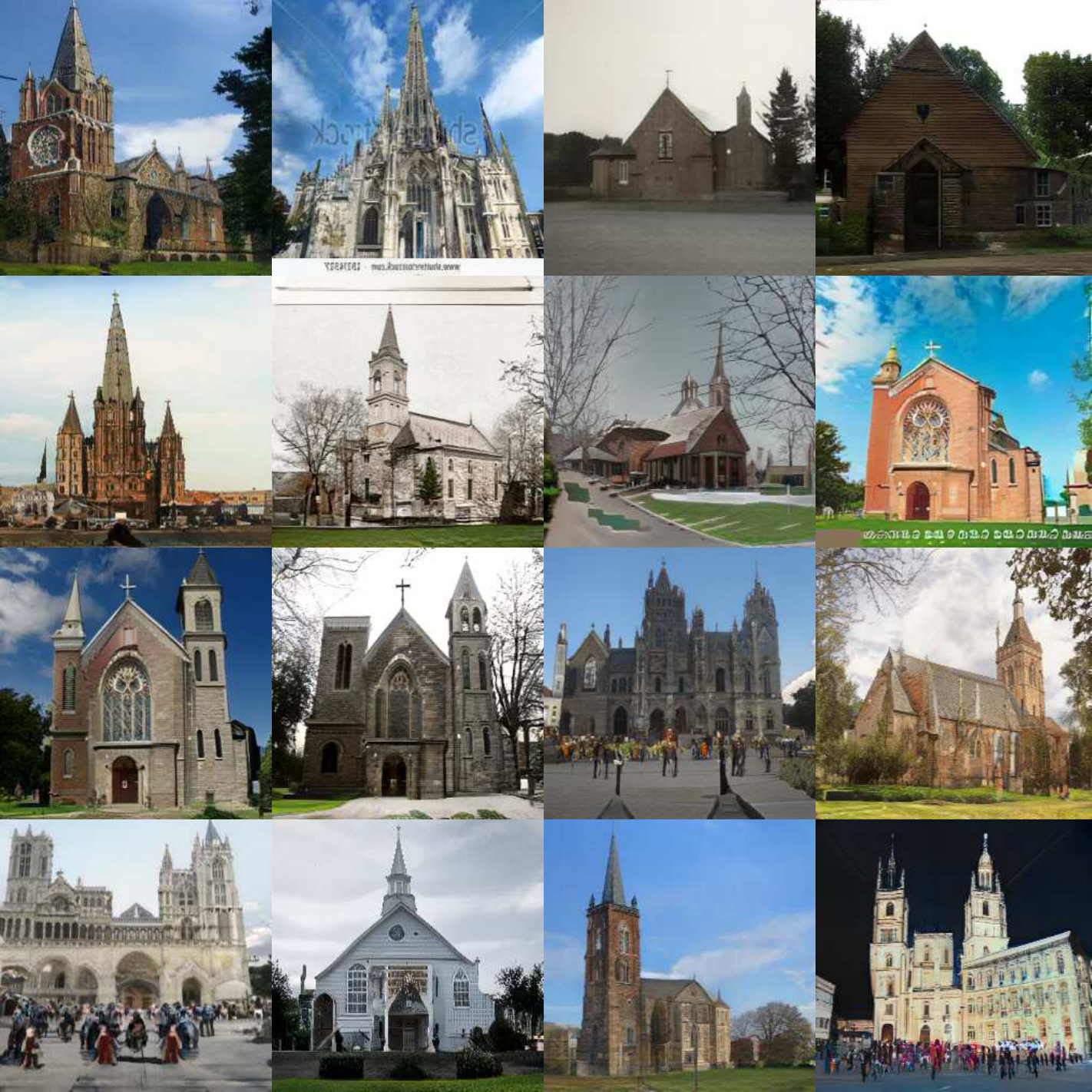}
    %\caption{Random uncurated samples from LSUN Church results with FID50k of \fidfullchurch.}
    % Sampling does not use truncation or any other tricks to increase fidelity.}
    \caption{Samples from LSUN Church with FID50k of 3.40
        }\label{fig:exp-church-samples}
\end{figure}

%% file: Includes/Tables/ffhq-ablation.tex
\begin{table}[!ht]
    \centering
    \begin{tabular}{l|c|r}
    \toprule
    Ablation & FID $\downarrow$ & diff $\downarrow$\\
    \hline
    StyleSwin & 2.81 & --\phantom{00} \\
    + NA & 2.74 &  \gain{-0.07}\\
    + \textbf{Hydra-NA} & 2.24 & \textbf{\gain{-0.50}}\\
    + Flips & \textbf{2.05} & \gain{-0.19}\\
    + Prog Di (4) & 2.55 & \loss{+0.50}\\
    \bottomrule
    \end{tabular}
    \caption{Ablation study comparing models on FFHQ-256 dataset.
        }\label{tab:ffhq256-ablation}
\end{table}

%% file: Includes/Tables/full_compare.tex
\begin{table*}[htpb]
    \centering
    \begin{tabular}{l|l|cc|c|cc}
        \toprule

        {\multirow{2}{*}{\textbf{Architecture}}} &
        {\multirow{2}{*}{\textbf{Model}}} 
            & \multicolumn{2}{|c|} {\textbf{FFHQ FID} $\downarrow$} 
            & \textbf{Church}
            & \multicolumn{2}{|c}
            {\textbf{Usage Metrics (256)}}\\
                & & \textbf{256} & \textbf{1024} & \textbf{256} 
                & \textbf{Throughput (img/s)} & \# \textbf{Params (M)} \\ \hline
        {\multirow{5}{*}{Convolutions}}
            &StyleGAN2~\cite{Karras_2020_CVPR} & 3.83 & 2.84 & 3.86 
                & 84.85 & 30.03 \\
            &StyleGAN3-T~\cite{NEURIPS2021_076ccd93} & - & 2.70 & - 
                & 108.84$^\star$ & 23.32$^\star$ \\
            &Projected GAN~\cite{NEURIPS2021_9219adc5} & 3.39 & - & 1.59 
                & 143.64 & 105.84\\
            &INSGen~\cite{NEURIPS2021_4e0d67e5} & 3.31 & - & - 
                & 89.00 & 24.94 \\
            &StyleGAN-XL~\cite{sauer2022stylegan} & 2.19 & 2.02 & -
                & 14.29 & 67.93\\
        \hline
        {\multirow{7}{*}{Transformers}}
            &GANFormer~\cite{hudson2021ganformer} & 7.42 & - & - 
                & - & 32.48\\
            &GANFormer2~\cite{hudson2021ganformer2} & 7.77 & - & -
                & - & -\\
            &HiT-S~\cite{NEURIPS2021_98dce83d} & 3.06 & - &  - 
                & 86.64$^\dagger$ & 38.01$^\dagger$ \\
            &HiT-B~\cite{NEURIPS2021_98dce83d} & 2.95 & - & - 
                & 52.09$^\dagger$ & 46.22$^\dagger$ \\
            &HiT-L~\cite{NEURIPS2021_98dce83d} & 2.58 & 6.37 & -
                & 20.67$^\dagger$ & 97.46$^\dagger$ \\
            &StyleSwin~\cite{Zhang_2022_CVPR} & 2.81 & 5.07 & 2.95
                & 62.48 & 48.93 \\
            &StyleNAT (\textbf{ours}) & 2.05 & 4.07 & 3.40 
                & 59.90 & 48.92\\
            %&TransGAN
        \hline
        {\multirow{7}{*}{Diffusion}}
            &DDPM~\cite{NEURIPS2020_4c5bcfec} & - & - & 7.89
                    & - & 256.00\\
            &D.StyleGAN2~\cite{wang2023diffusiongan} & - & 2.83 & 3.17
                & - & \phantom{0}23.94\\\
            &D.Proj.Gan~\cite{wang2023diffusiongan} & - & - & 1.85
                & - & 105.85\\
            &LDM~\cite{Rombach_2022_CVPR} & 4.98 & - & 4.02
                & 1.28 & 329.32\\
            &LFM~\cite{dao2023flow} & 4.55 & - & 5.54
                & 4.18 & 457.06\\
            &UDM~\cite{pmlr-v162-kim22i} & 5.54 & - & - 
                & - & \phantom{0}65.58\\
            &Unleashing~\cite{bond2022unleashing} & 6.11 & - & 4.07 
                & 6.65 & 159.96\\
        \bottomrule
    \end{tabular}
    \caption{FID50k results. Usage Metrics are evaluated at
        $256\times256$ resolution for fair comparison and were collected
        ourselves.
        StyleNAT does not utilize any FID enhancing processing, such as
        StyleGAN's truncation trick.
        $^\dagger$HiT-L was optimized for TPU and there is no existing pytorch
        version to compare. There are also no public checkpoints so come from 
        the paper~\cite{NEURIPS2021_98dce83d} and are reported on a V100. 
        For StyleGAN3 we ran without loading a checkpoint. 
        While most architectures are built off of the official StyleGAN models,
        they may not all be able to utilize the custom CUDA kernels, which
        can significantly increase throughput~\cite{Karras_2020_CVPR}.
    }\label{table:exp-maintable}
\end{table*}

%% file: sec/3_methods/church.tex
\subsection{LSUN Church}

We include LSUN Church, which is composed of churches, towers,
cathedrals, temples, and many varying features in the backgrounds.
This proves a challenging dataset for many generators and is particularly troublesome for evaluation metrics due to the high variance in image details as
well as many metrics being biased towards shape and
texture.~\cite{NEURIPS2020_db5f9f42,geirhos2018imagenettrained}.
Part of this is due the high rates of biological images in ImageNet~\cite{kynkaanniemi2023the} as
well as the natural texture bias for CNNs~\cite{geirhos2018imagenettrained}.
This results in high rates of generative artifacts even in state of the art 
models like Projected GAN~\cite{NEURIPS2021_9219adc5} which have significantly
better FID scores, demonstrating the metric limitations.

This experiment starts with the same configuration as the FFHQ experiments.
Initial results showed the model quickly diverged.
We increased the number of partitions to 4, evenly spacing the dilations,
incorporating data between local and global, which resulted in significant
improvement.
Increasing the number of final heads to 8, which decreases the effective
dimension per head, demonstrated minor improvement as well as when increasing
the number of partitions.
The results of this can be seen in \Cref{tab:exp-church-heads}.
While we were unable to fully optimize our hyperparameters, this result still
shows strong performance and the resultant imagery is competitive with other
top-performing models.
Interestingly, we observe our model is able to generate the Shutterstock
watermark, common in the dataset, and even able to generate convincing IDs.
This behavior was observed even in training runs where we obtained higher FID
scores.
Samples of such images can be found in \Cref{fig:exp-church-overfit} and we
observed a strong tendency for these images to be of higher quality when
compared to other samples.
We were unable to observe this behavior in other GANs trained on Church despite
generating large numbers of samples.
We believe such samples potentially demonstrate overfitting and provide a deeper 
analysis with attention maps in \Cref{app:attn_maps}.
While this effect may demonstrate that more care may be needed in training more
diverse datasets, such as Church, it does demonstrate the capacity for our model
to learn high quality representations.

\input{Includes/Figures/short_church_overfit}
\input{Includes/Figures/short_attn_map}

%% file: Includes/Figures/short_church_overfit.tex
\begin{figure}
    \centering
    % \begin{subfigure}[b]{0.47\linewidth}
    \adjustbox{trim={.5\width} 0 0 0,clip}%
  {\includegraphics[width=0.79\textwidth]{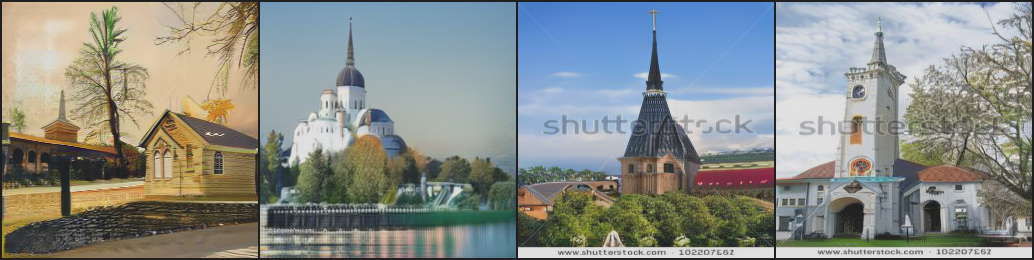}}
    \caption{Examples of Church samples overfitting and showing text features. 
             Clear watermark and source address where ``shutterstock'' is 
             clearly readable. FID 4.22
        }\label{fig:exp-church-overfit}
\end{figure}

%% file: Includes/Figures/short_attn_map.tex
\begin{figure}[ht]
    \centering
    \begin{subfigure}[b]{0.43\linewidth}
    \includegraphics[width=\linewidth]{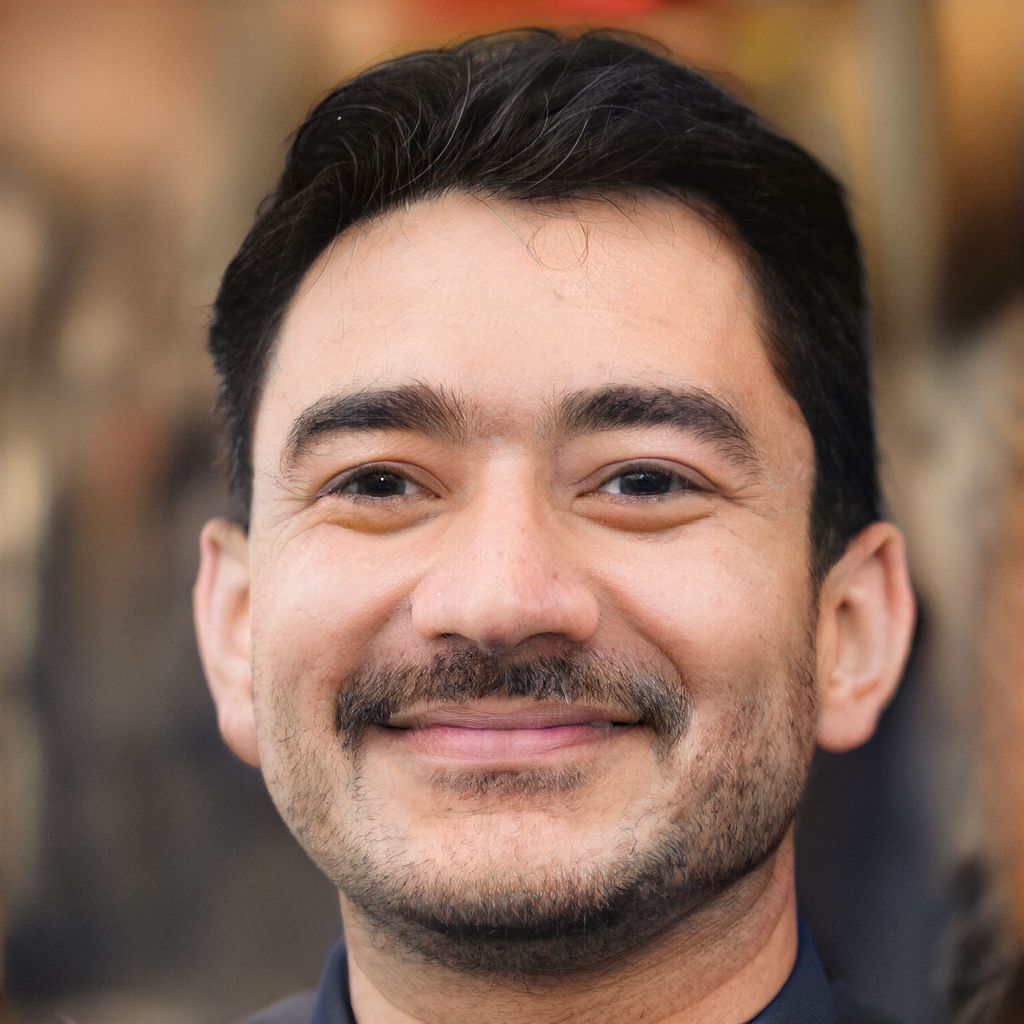}
        \caption{StyleNAT
            }\label{fig:main_snatsamp}
    \end{subfigure}
    \begin{subfigure}[b]{0.43\linewidth}
    \includegraphics[width=\linewidth]{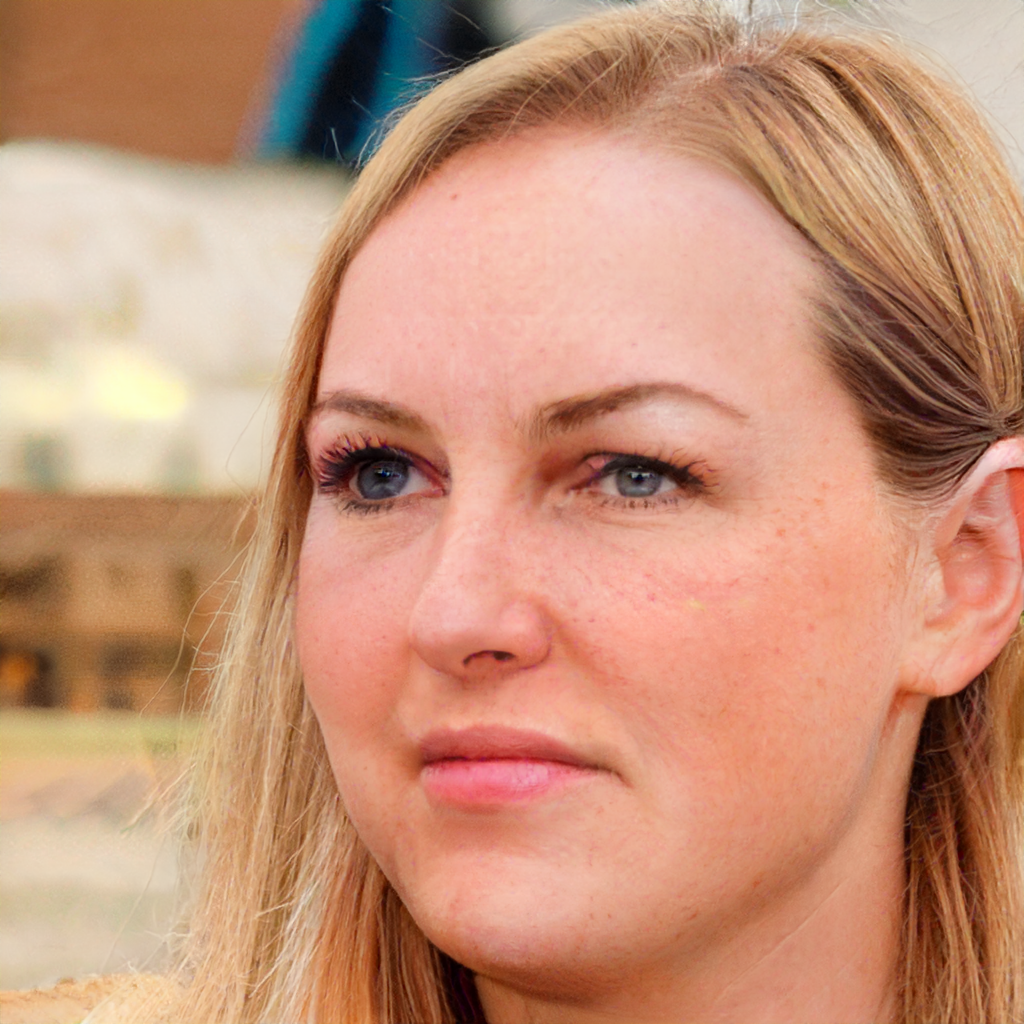}
        \caption{StyleSwin
            }\label{fig:main_ssamp}
    \end{subfigure}
    \\
    \begin{subfigure}[b]{0.43\linewidth}
        \includegraphics[width=\linewidth]{Images/NA/NATLayer_80kerneldensity_channel.png}
        \caption{1024 Level, Layer 1
            }\label{fig:main_l8a0}
    \end{subfigure}
    \begin{subfigure}[b]{0.43\linewidth}
        \includegraphics[width=\linewidth]{Images/NA/NATLayer_81kerneldensity_channel.png}
        \caption{1024 Level, Layer 2
            }\label{fig:main_l8a1}
    \end{subfigure}
    \\
    \begin{subfigure}[b]{0.43\linewidth}
     \includegraphics[width=\linewidth]{Images/swin/1024/StyleSwinTransformerBlock_80.png}
        \caption{1024 Level, Layer 1
            }\label{fig:main_styleswin_l8a0}
    \end{subfigure}
    \begin{subfigure}[b]{0.43\linewidth}
        \includegraphics[width=\linewidth]{Images/swin/1024/StyleSwinTransformerBlock_81.png}
        \caption{1024 Level, Layer 2
            }\label{fig:main_styleswin_l8a1}
    \end{subfigure}

    \caption{Attention maps for FFHQ-1024 demonstrating
        StyleNAT~\Cref{fig:main_snatsamp} and StyleSwin~\Cref{fig:main_ssamp}.
            For StyleNAT top row represents kernels of size 7 and no dilation.
            Bottom rows are similarly sized kernels with dilation 128. For
            StyleSwin the top row is the 4 way partitioned windowing and shifted
            windows on bottom.
            See \Cref{app:attn_maps} for more details.
        }\label{fig:exp-attn-maps}
\end{figure}

%% file: Includes/Tables/church_ablation.tex
\begin{table}[!ht]
    \centering
    \begin{tabular}{l|c|c|r}
    \toprule
    \textbf{Partitions} & \textbf{Min Heads} & \textbf{FID} $\downarrow$ & \textbf{Diff} $\downarrow$\\
    \hline
    2 & 4 & 23.33 & --\phantom{00} \\
    4 & 4 & 6.08 &  \gain{-17.25}\\
    6 & 8 & 5.50 & \gain{-0.58}\\
    8 & 8 & 3.40 & \gain{-2.10} \\
    %8 & 16 & \stilltraining{16.92} & \gain{-} \\
    \bottomrule
    \end{tabular}

    \caption{Comparison for number of head partitions when learning LSUN Church.
        Min heads represents the minimum number of heads in our transformer.
        }\label{tab:exp-church-heads} 
\end{table}

%% file: sec/3_methods/analysis.tex
\subsection{Analysis}\label{sec:exp-analysis}

A full comparison of our results can be found in~\Cref{table:exp-maintable}.
In addition to FID calculations we include measurements of image throughputs and
the number of parameters for each model for a more holistic comparison.
On FFHQ-256 StyleNAT not only achieves the best FID but also does so with fewer
parameters and higher throughput than models with similar performance.

To better understand the results, we generalize the technique for attention
mapping, extending it to work for windowed attention such as for Swin and NA.
To our best knowledge this is the first instance of attention maps for either
Swin or NA.
For NA this requires extracting the queries and keys, taking the means of the
queries and flattening keys, and taking the matrix product.
For Swin we unroll half the heads, reorienting them to the correct alignment
before performing the same procedure.
We find that this process clearly shows the blocking artifacts discussed by the
StyleSwin authors~\cite{Zhang_2022_CVPR}.
An example of which can be seen in~\Cref{fig:exp-attn-maps} and
in~\Cref{fig:progressiveAttnMaps_swin} it can be seen that these artifacts
appear early on within the network.

Stein~\etal~\cite{stein2023exposing} recently performed the largest human based
evaluation of generative models, in an effort to find which metrics corresponded
to human preference and to determine if they were correlated to a human's
ability to distinguish a real and synthetic based image.
This study used our model\footnote{the authors of this work have no affiliation
with Stein~\etal's work} which had the highest human error rate for FFHQ, a 20\%
improvement from the next best work.
This independent result helps illustrate the benefits provided by Hydra-NA.

%% file: sec/5_conclusion.tex
\section{Conclusion}\label{sec:conclusion}

In this work we present StyleNAT, which shows a highly flexible generative
network able to accomplish generation on multiple types of datasets.  
The Hydra-NA design allows for an arbitrary number of partitions of heads 
to utilize different kernels and/or dilations.
This design allows for a single attention mechanism to combine various 
viewpoints and combine this knowledge to create a much more powerful 
architecture while remaining computationally efficient. 
Our work demonstrates that this style of architecture is able to achieve state 
of the art performance on image generation tasks.  
It also demonstrates that attention based GANs are able to outperform
convolution based ones.
We also include our generalized attention map visualization technique that 
works for local attentions such as NA and Swin, along with a detailed 
investigation of the visual fidelity and origin of generative artifacts.
Additionally, we note that the parameters in our method need not be fixed and
can be learned.
We believe such an investigation would be of interest to researchers with larger
computational budgets and will likely lead to better results but increased
instability.

%% file: sec/6_appendix.tex
\appendix
\section{Appendix}\label{appendix}

In our Appendix we cover issues of reproducibility and focus on image analysis
which requires visually inspecting figures.
All images are embedded using vector graphics, to allow the reader to zoom in,
but note that the source images are not vectored. 
To organize our Appendix we first include sections to make our work
reproducible.
\Cref{app:model-arch} covers the model architecture, \Cref{app:paramsAndThroughputs} 
includes important details about the parameters and throughput measurements, 
in \Cref{app:metrics} we have a small discussion on the limitations of metrics
to help clarify the motivation for our visual analysis, which is coverend in
\Cref{app:visual_analysis}.
The visual analysis includes a discussion of the attention maps
(\Cref{app:attn_maps}), including maps at multiple resolutions for both StyleNAT
and StyleSwin, and for both FFHQ~(\Cref{app:swin_ffhq,app:na_ffhq}) and
Church~(\Cref{sec:swin_church_attnmap,sec:snat_church_attnmap}) datasets.

\input{sec/6_appendix/ModelArch}

\input{sec/6_appendix/ParamThroughput}
\input{sec/6_appendix/Metrics}
\input{sec/6_appendix/Visual}
\input{sec/6_appendix/AttnMaps}

%% file: sec/6_appendix/ModelArch.tex
\section{Model Architecture}\label{app:model-arch}

\input{Includes/Tables/ModelArch_2part}

\input{Includes/Tables/ModelArch_progressive}

%\Cref{fig:method-architecture} includes a depiction of the first 2 resolution 
Figure 4 includes a depiction of the first 2 resolution 
levels of the StyleNAT architecture.
At each resolution we use $N=2$ transformer blocks, but this is configurable for
scale 
%\todo{(e.g. StyleGAN3~\cite{NEURIPS2021_076ccd93} uses $N=14$, and
%StyleGAN-XL~\cite{sauer2022stylegan} uses a variable amount).}
For all experiments the layers have the same attention parameters, but there is
nothing preventing one from making these distinct.
Similarly, we use a constant kernel (window), $k$, size of 7.
In all experiments we use a combination of no dilation (1) to create local dense
windowing, and larger but more sparse receptive fields (dilation $>1$).
There are no restrictions in the architecture that require this, but we felt
that these choices would be most clear in demonstrating the effectiveness of our
work, due to the closeness to StyleSwin's kernel size of 8.

We include~\Cref{tab:2split_model_arch}~and~\Cref{tab:pyramid_model_arch} which 
show the parameters for the ``split head'' experiments and the progressive, or 
``pyramid dilation'', used in the ablation study for LSUN Church
%experiments~(\Cref{tab:exp-church-heads}).
experiments~(Table 2).
For ``pyramid dilation'' we always use a growth based on powers of $2$, but this
is not a requirement of the architecture.

For our Church experiments we found a more progressive dilation solution to work better.
We provide an example of our ``pyramid'' style dilation, wherein we continually 
grow progression mixing several levels of dilation.

%%%%%%%%%%%%%%%%%%%%%%%%%%%%%%

\subsection{Hydra-NA Code}\label{appdx:hydracode}

\input{Includes/HydraCode}

We include some PyTorch code for implementing
Hydra-NA in~\Cref{fig:hydracode}.
Note that this code is not optimized and that all the natten operations could be
performed in parallel for increased performance.
%%%%%%%%%%%%%%%%%%%%%%%%%%%%%%%

\subsection{Other Configurations}
The configurations are truly variadic, and allows for a large number of possible
combinations. 
The primary reason for our parameter choices was to minimize the compounding
factors in our experiments and minimize the influence of our parameter choices
to the results, compared to our nearest competitor, and while fitting within our
computational budget.

During our exploratory phase, we performed a few short experiments where we 
replaced the Swin Attention in StyleSwin with Hydra-NA.
In general we found increased performance under the conditions that we did not
use kernels sized $3$ and that we did not use a very large and very small
kernel when replacing the last layer.
For example, using a split head design with one kernel sized 3 and the other 
kernel sized 45, both with dilation 1, showed worse performance when compared to
StyleSwin.
But we did find that there was increased performance if instead we used kernels
size 45 for both partitions.
The trade-off is that the larger kernels requires significantly more GPU memory
and computation time.
For example, while StyleSwin would utilize $\approx36$ GB of DRAM per GPU, when
using a kernel size of 45 we used $\approx76$GB of DRAM per GPU and
approximately 6x wall time (measured at a resolution of 1k iterations).
Our final StyleNAT architecture uses slightly less DRAM than StyleSwin and
$\approx30$s more per 1k iterations.
Note that these DRAM requirements are not due exclusively to the model itself
but these also include the EMA model, batch images, pre-fetches, and other such
additional data that may be included when training.
We show some results in~\Cref{tab:replace_last_layer}, comparing to our training
of StyleSwin and include comparisons with our final StyleNAT result.
These results give evidence that the model can scale and that its performance
may be proportional to the effective computation.
These runs were never taken to convergence and thus we believe such experiments 
would be worthwhile, but are outside the compute budgets of our lab.
We also performed similar replacement experiments with other resolution levels,
but did not explore larger kernels.
Without dilation we still found incremental improvements over StyleSwin.

\begin{table}[ht]
    \centering
    \begin{tabular}{c|c|c|c}
        \toprule
        %& \multicolumn{3}{|c|}{\textbf{FID}}\\
        kernel size & 10k & 25k & 50k\\
        \hline
        StyleSwin & 84.63 & 29.89 & 18.85\\
        3/45 & 85.17 & 31.95 & 22.25\\
        7/45 & 86.32 & 41.15 & 25.66\\
        45/45 & 68.65 & 28.36 & 17.95\\
        StyleNAT & 145.57 & 20.10 & 10.81\\
        \bottomrule
    \end{tabular}

    \caption{FID results, showing effects of replacing last resolution level's Swin Attention with 
    a split head NA. First column shows the kernel sizes. No dilation was used.
    StyleSwin and StyleNAT included for comparison.
        }\label{tab:replace_last_layer} 
\end{table}

\subsection{Potential Configurations}

In this section we discuss the potential configurations for Hydra-NA.
We believe this discussion will help others determine how to optimize this
architecture and better understand the potential flexibility of the network.
Our hyperparmeters were fixed in favor of ensuring higher interoperability of
results given compute constraints.
Optimization of the network will require further search, but we demonstrate that
these are reasonable choices to start with.
This discussion will also help practitioners optimize not just for the final
fidelity of the images, but for computational constraints, noting that smaller
kernels will increase inference speeds and reduce GPU memory. 
%Our choice of GANs as a platform for experimentation was also chosen to focus on
%the potential information gain from this formulation.
%We believe that Hydra-NA will still provide benefits to other architectures and
%domains.

Our kernels, $k$, can range from a size of $3$ to the nearest odd integer 
smaller than the resolution, $\mathcal{R}$.
This means that there will be $\frac{\mathcal{R}}{2}-1$ potential kernels.
We can also note that any kernel sized $\geq\frac{\mathcal{R}}{2}$ cannot have
any dilation without exceeding the image size\footnote{It is technically
possible to extend NA/DiNA to work for kernels larger than the image size but we
will operate under this condition for practical purposes.}.
There are $\frac{\mathcal{R}}{4}$ such kernels.
The max dilation size for a given kernel is 
$\left\lfloor\frac{\mathcal{R}}{k}\right\rfloor$.
Thus the total number of configurations, per head, is
\begin{align}\label{eq:num_configs}
    N_c &= \sum_{i=1}^{\mathcal{R}/2-1} 
                \left\lfloor
                    \frac{\mathcal{R}}{2i + 1}
                 \right\rfloor
                 \\
        \label{eq:simp_num_configs}
        &= \frac{\mathcal{R}}{4} + 
                \sum_{i=1}^{\mathcal{R}/4-1}
                \left\lfloor
                     \frac{\mathcal{R}}{2i+1}
                 \right\rfloor
\end{align}
This, of course, is assuming that we are using square kernels, square images,
and images with an even number of pixels.
None of these are actual constraints to our formulation, so this could actually
double\footnote{For an image with an odd number of pixels there are
$\left\lfloor\frac{\mathcal{R}}{2}\right\rfloor$ kernels,
$\left\lceil\frac{\mathcal{R}}{4}\right\rceil$ kernels with no potential
dilations}.

For our FFHQ-256 configuration we have $16$ heads for resolutions $4\times4$ to
$64\times64$, $8$ heads for resolution $128\times128$, and $4$ heads for all other
resolutions, and $2$ transformers per resolution, this results in
$2\times\left(16\times\left(4 + 14 + 37 + 97\right) + 8\times237 +
4\times565\right) = 13176$
potential configurations!
($\approx47$k for $1024\times1024$ resolution images)
With $8$ minimum heads in our LSUN Church experiments that allows for $17696$ potential 
configurations.
This is an extraordinary number of possible configurations for our architecture,
with allows for potentially high rates of expressibility.
In our experiments we manually selected parameters to ensure similarity to our
best comparator \emph{but these hyper-parameters can be learned}.

%% file: Includes/Tables/ModelArch_2part.tex
\begin{table}[ht]
    \centering
    \begin{tabular}{c|c|c|c}
        \toprule
        Level & Kernel Size & Dilation & Dilated Size \\
        \hline
        $\phantom{111}4$ & - & \phantom{--}- & \phantom{--}- \\
        $\phantom{111}8$ & 7 & \phantom{11}1 & \phantom{11}7 \\
        $\phantom{11}16$ & 7 & \phantom{11}2 & \phantom{1}14\\
        $\phantom{11}32$ & 7 & \phantom{11}4 & \phantom{1}28 \\
        $\phantom{11}64$ & 7 & \phantom{11}8 & \phantom{1}56\\
        $\phantom{1}128$ & 7 & \phantom{1}16 & 112 \\
        $\phantom{1}256$ & 7 & \phantom{1}32 & 224 \\
        $\phantom{1}512$ & 7 & \phantom{1}64 & 448\\
        $1024$           & 7 & 128           & 896\\
        \bottomrule
    \end{tabular}

    \caption{StyleNAT 2-Partition Model Architecture. First level uses
        Multi-headed Self Attention and not DiNA. This model is used for all 
        FFHQ results, at all resolutions.
        }\label{tab:2split_model_arch} 
\end{table}

%% file: Includes/Tables/ModelArch_progressive.tex
\begin{table}[ht]
    \centering
    \begin{tabular}{c|c|l}
    \toprule
    Level & Kernel Size & Dilations \\
    \hline
    $\phantom{111}4$    & -  & -  \\
    $\phantom{111}8$    & 7 & 1 \\
    $\phantom{11}16$   & 7 & 1,2\\ 
    $\phantom{11}32$   & 7 & 1,2,4\\ 
    $\phantom{11}64$   & 7 & 1,2,4,8\\ 
    $\phantom{1}128$  & 7 & 1,2,4,8,16\\ 
    $\phantom{1}256$  & 7 & 1,2,4,8,16,32\\ 
    $\phantom{1}512$  & 7 & 1,2,4,8,16,32,64\\ 
    $1024$ & 7 & 1,2,4,8,16,32,64,128\\
    \bottomrule
    \end{tabular}
    \caption{Example of progressive dilation with 8 heads minimum, referred to
        ``pyramid dilation."
    }\label{tab:pyramid_model_arch}
\end{table}

%% file: Includes/HydraCode.tex
%%%%%%%%%%%%%%%%%%%
\definecolor{codegreen}{rgb}{0,0.6,0}
\definecolor{codepurple}{rgb}{0.58,0,0.82}
\definecolor{codeblue}{rgb}{0, 0, 0.6}
\definecolor{codered}{rgb}{0.6, 0, 0}
\definecolor{codeyellow}{rgb}{0.6, 0.6, 0}
\definecolor{background}{rgb}{0.9, 0.9, 0.9}

\lstdefinelanguage{pytorch}
{
    morekeywords={class, def, super, len, for, range, len, if, else, zip, apply, return, __init__, append},
    keywordstyle=\color{codepurple}\textbf,
    keywords=[2]{self,},
    keywordstyle=[2]\color{codegreen}\textbf,
    keywords=[3]{in, True, False, None, or, \%, },
    keywordstyle=[3]\color{codeblue}\textbf,
    keywords=[4]{nn, torch, Module, Linear, cat, Parameter, ParameterList, chunk, shape, zeros, Dropout, squeeze, reshape, permute, trunc_normal_, softmax, forward, dim},
    keywordstyle=[4]\color{codered}\textbf,
    keywords=[5]{natten2dqkrpb,natten2dav,HydraNeighborhoodAttention},
    keywordstyle=[5]\color{codeyellow}\textbf,
}

\begin{figure*}[htbp]
%\begin{mdframed}[backgroundcolor=background, hidealllines=true,]
% Text sizes
% tiny, scriptsize, footnotesize, small, normalsize, large, Large, LARGE, huge, Huge
% Too large and you have to break up lines more
% tiny is probably largest while still fitting on single page, single column
% Larger and you need both columns
\begin{lstlisting}[language=pytorch,
                   basicstyle=\ttfamily\linespread{1.1}\footnotesize,
                   backgroundcolor=\color{background},
                   ]
class HydraNeighborhoodAttention(nn.Module):
    def __init__(self, dim : int, kernel_sizes : list[int], num_heads : int, 
                 qkv_bias : bool=True, qk_scale : Optional[float]=None, attn_drop : float=0., 
                 proj_drop : float=0., dilations : list[int]=[1]) -> None:
        super().__init__()
        self.num_splits, self.num_heads = len(kernel_sizes), num_heads
        self.kernel_sizes, self.dilations = kernel_sizes, dilations
        self.head_dim = dim // self.num_heads
        self.scale = qk_scale or self.head_dim ** -0.5
        self.window_size = []
        for i in range(len(dilations)):
            self.window_size.append(self.kernel_sizes[i] * self.dilations[i])
        self.qkv = nn.Linear(dim, dim * 3, bias=qkv_bias)
        if num_heads % len(kernel_sizes) == 0:
            self.rpb = nn.ParameterList([nn.Parameter(
                    torch.zeros(num_heads//self.num_splits, (2*k-1), (2*k-1))) 
                for k in kernel_sizes])
            self.clean_partition = True
        else:
            diff = num_heads - self.num_splits * (num_heads // self.num_splits)
            rpb = [nn.Parameter(torch.zeros(
                    num_heads//self.num_splits, (2*k-1), (2*k-1))) 
                for k in kernel_sizes[:-diff]]
            for k in kernel_sizes[-diff:]:
                rpb.append(nn.Parameter(torch.zeros(
                    num_heads//self.num_splits + 1, (2*k-1), (2*k-1))
                ))
            self.rpb = nn.ParameterList(rpb)
            self.clean_partition = False
            self.shapes = [r.shape[0] for r in rpb]
        [trunc_normal_(rpb, std=0.02, mean=0.0, a=-2., b=2.) for rpb in self.rpb]
        self.attn_drop = nn.Dropout(attn_drop)
        self.proj = nn.Linear(dim, dim)
        self.proj_drop = nn.Dropout(proj_drop)
        
    def forward(self, x:torch.Tensor) -> torch.Tensor:
        B, H, W, C = x.shape
        qkv = self.qkv(x)
        qkv = qkv.reshape(B, H, W, 3, self.num_heads, self.head_dim)
        q,k, v = qkv.permute(3, 0, 4, 1, 2, 5).chunk(3,dim=0)
        q = q.squeeze(0) * self.scale
        k,v = k.squeeze(0), v.squeeze(0)
        if self.clean_partition:
            q = q.chunk(self.num_splits, dim=1)
            k = k.chunk(self.num_splits, dim=1)
            v = v.chunk(self.num_splits, dim=1)
        else:
            i, _q, _k, _v = 0, [], [], []
            for h in self.shapes:
                _q.append(q[:, i:i+h, :, :])
                _k.append(k[:, i:i+h, :, :])
                _v.append(v[:, i:i+h, :, :])
                i = i+h
            q, k, v = _q, _k, _v
        attention = [natten2dqkrpb(_q, _k, _rpb, _kernel_size, _dilation)
                         for _q, _k, _rpb, _kernel_size, _dilation in \
                         zip(q, k, self.rpb, self.kernel_sizes, self.dilations)]
        attention = [self.attn_drop(a.softmax(dim=-1)) for a in attention]
        x = [natten2dav(_attn, _v, _k, _d) for _attn, _v, _k, _d 
                in zip(attention, v, self.kernel_sizes, self.dilations)]
        x = torch.cat(x, dim=1).permute(0, 2, 3, 1, 4).reshape(B, H, W, C)
        return self.proj_drop(self.proj(x))
    \end{lstlisting}
    %\end{mdframed}
    \caption{Full code for StyleNAT's Hydra-NA module. Type hinting included for
             added clarity. \emph{Requires NATTEN package}. Using NATTEN
             v0.14.6, subsequent versions may need modifications. Code is 
             unoptimized, intended for research and clarity.
        }\label{fig:hydracode}
\end{figure*}
%%%%%%%%%%%%%%%%%%%

%% file: sec/6_appendix/ParamThroughput.tex
\section{Parameters and Throughputs}\label{app:paramsAndThroughputs}

We measured all throughputs via the official code bases and their respective
sampling practices.
Modifications to code were only done when necessary for equivalent evaluation
and we do not believe these would result in meaningful differences.
When checkpoints were provided we used those and followed the run formulas
provided in a project's README file.
Several models did not provide checkpoints, so we used the configurations most
similar to those that would be used during training.
For parameter counts we used the open source tool
\emph{graftr}\footnote{\href{https://github.com/lmnt-com/graftr}{https://github.com/lmnt-com/graftr}}, to directly probe
checkpoints.

Many StyleGAN models provide checkpoints as pickle files, which we loaded using
the official ``legacy.\_LegacyUnpickler'' method and converted to PyTorch
checkpoints.
Parameters were based on the generator's Exponential Moving Average (EMA)
models if provided, otherwise the generator. 
We did not count parameters included in the discriminator or elsewhere, focusing
on what parameters are required for synthesis. 
For Unleashing Transformers~\cite{bond2022unleashing} we include the total
counts for the VQGAN (83.12M) and the Absorbing EMA's denoising function
(76.84M).
For LDM~\cite{Rombach_2022_CVPR} we find 603M parameters within the state\_dict,
but this includes the EMA model.
The ``model'' contains 274M parameters, matching Table 12 in their Appendix E.1, 
but we include all non-EMA parameters, believing these are still necessary for
synthesis. 
We also note that LDM's training time and memory load can likely be significantly 
improved due to the implementation for loading the EMA model, which requires
more DRAM than necessary.

For HiT GAN we were unable to gather throughputs and relied upon the paper's
numbers.
HiT GAN was written in tensorflow v1 and we were unable to find python wheels
that would satisfy our system's requirements.
In Table 4 of the HiT GAN paper~\cite{NEURIPS2021_98dce83d}, they report that 
on FFHQ-256 HiT-L has a throughput of 20.67 imgs/s and 97.64M parameters when 
run on a single NVIDIA V100 GPU.
They similarly note that StyleGAN2 achieves 95.79 imgs/s using 30.03M 
parameters.
The original StyleGAN2 paper~\cite{Karras_2020_CVPR} reports only reports 
results for FFHQ-1024 (Section 6), at 61 imgs/s.
Their results were gathered on a NVIDIA DGX-1 with 8 V100 GPUs and during
training.
We note that their throughput measurements for StyleGAN2 are a bit over $10\%$
higher than our measurements.
We experienced similar issues when attempting to measure performance for
GANformer and their GitHub page was ambiguous as to corresponding checkpoints.
We did not reach out to the authors but similar questions were raised on their 
GitHub issues page, which appears to be inactive.

All measurements were performed on a single NVIDIA A100 GPU using Python
3.10.13, PyTorch 2.1.0, and CUDA 12.1 (installed with the official PyTorch
instructions. The system CUDA version was 12.0).
All measurements are normalized to batch size to make them independent of memory
constraints.
We utilize 50 rounds of warmup before sampling 100 rounds, which we average
over.
For diffusion models we used the highest batch size we could fit in memory since
this yielded the best results.
We took the highest throughput from multiple measurements, noting that variance
was generally quite low.

For performance we enabled TF32 through torch's backends and utilized torch's
inference\_mode as opposed to no\_grad.
We used no other optimizations and made no attempt to use PyTorch's compile
feature or NVIDIA's TensorRT, which should boost performance for all models.
Several models have cuDNN enabled, and we left this at the default value.

We also note that variance in Style-based models can vary greatly due to their
dependence on custom CUDA kernels written for the bias activation, conv2d
gradient fix and resampling, fused-multiple-add (FMA), grid sampling gradient
fix, and
upfirdn2d\footnote{\href{https://github.com/NVlabs/stylegan2-ada-pytorch/tree/main/torch_utils/ops}{https://github.com/NVlabs/stylegan2-ada-pytorch/torch\_utils/ops}}
Karras~\etal noted that these implementations could account for upwards of 40\%
improvement in throughputs.
For example, ProjectedGAN~\cite{NEURIPS2021_9219adc5} makes use of all of these
as well as introduces some additional CUDA
kernels\footnote{\href{https://github.com/autonomousvision/projected-gan/tree/main/torch_utils/ops}{https://github.com/autonomousvision/projected-gan/torch\_utils/ops}}.
Additionally, StyleGAN3~\cite{NEURIPS2021_076ccd93} (and consequently
StyleGAN-XL~\cite{sauer2022stylegan}) introduced additional
CUDA kernels. 
In Appendix D Karras~\etal note that the speedup is $\approx20-40\times$ when 
comparing against native PyTorch operations, with an
overall training speedup of $\approx10\times$ but do not specify if there are
any differences from the StyleGAN2 implementations. 
On the other hand, GANformer, StyleSwin, and StyleNAT rely far less upon these
implementations and thus likely results from suboptimal throughput performance.
Note that StyleNAT also has a custom CUDA implementation for NA/DiNA, but not
the head splitting parts of the architecture.
Please refer to~\Cref{fig:hydracode} for more clarity.

For throughput measurements we use the procedure shown in~\Cref{fig:throughput}.
This code is also available on our GitHub project.

%%%%%%%%%%%%%%%%%%%
\definecolor{codegreen}{rgb}{0,0.6,0}
\definecolor{codepurple}{rgb}{0.58,0,0.82}
\definecolor{codeblue}{rgb}{0, 0, 0.6}
\definecolor{codered}{rgb}{0.6, 0, 0}
\definecolor{codeyellow}{rgb}{0.6, 0.6, 0}
\definecolor{background}{rgb}{0.9, 0.9, 0.9}

\lstdefinelanguage{pytorch}
{
    morekeywords={class, def, super, len, for, range, len, if, else, zip, apply, return, __init__, append},
    keywordstyle=\color{codepurple}\textbf,
    keywords=[2]{self,},
    keywordstyle=[2]\color{codegreen}\textbf,
    keywords=[3]{in, True, False, None, or, \%, },
    keywordstyle=[3]\color{codeblue}\textbf,
    keywords=[4]{nn, torch, Module, Linear, cat, Parameter, ParameterList, chunk, shape, zeros, Dropout, squeeze, reshape, permute, trunc_normal_, softmax, forward, dim},
    keywordstyle=[4]\color{codered}\textbf,
    keywords=[5]{natten2dqkrpb,natten2dav,HydraNeighborhoodAttention},
    keywordstyle=[5]\color{codeyellow}\textbf,
}
\begin{figure}[h]
%\begin{mdframed}[backgroundcolor=background, hidealllines=true,]
\begin{lstlisting}[language=pytorch,
                   basicstyle=\ttfamily\linespread{1.1}\footnotesize,
                   backgroundcolor=\color{background},
                   ]
@torch.inference_mode()
def calculate_throughput(args):
    torch.backends.cuda.matmul.allow_tf32 = True
    times = torch.empty(rounds)
    noise = torch.randn((batch_size, latent_dim), 
                        device="cuda")
    for _ in range(warmup):
        imgs = generator(noise)
    for i in range(rounds):
        starter = torch.cuda.Event(
                    enable_timing=True)
        ender   = torch.cuda.Event(
                    enable_timing=True)
        starter.record()
        imgs = generator(noise)
        ender.record()
        torch.cuda.synchronize()
        times[i] = starter.elapsed_time(ender) \
                   / 1000
    imgsPerSecond = total_imgs / total_time \
                    / batch_size
    print(f"{torch.std_mean(imgsPerSecond)}")
\end{lstlisting}
%\end{mdframed}
\caption{Pseudo code for throughput measurements}
\label{fig:throughput}
\end{figure}

%% file: sec/6_appendix/Metrics.tex
\section{Metrics and Image Quality}\label{app:metrics}

As the quality of our synthesized images increases we more quickly approach the
limitations our our metrics.
We note that if one compares the FID of the first 10k images of FFHQ-256 to the
bottom 60k this yields a FID of 2.25.
While this is not a completely fair comparison to how measurements are formed
when comparing synthesised images to training images\footnote{this is 10k vs
60k which does change results}, it illustrates limitations in FID and the
natural variance within the data itself.
A true metric measuring the visual fidelity to humans still remains
elusive~\cite{stein2023exposing,NEURIPS2019_0234c510,kynkaanniemi2023the,Parmar_2022_CVPR,pmlr-v119-naeem20a,NEURIPS2019_65699726,morozov2021on,bińkowski2018demystifying}. 

Additionally, to measure the preformance of models we aggregate over samples,
and such results will not convey the difference between the quality of a single
image and that of the average image.
These make truly evaluating the quality of generative networks difficult and
illustrates the need to further develop metrics and for evaluators to pay close
attention to samples from the networks. 

A recent work by Stein~\etal~\cite{stein2023exposing} performed on of the, if
not the, largest human study to date, investigating the predictive power of
various metrics when compared to a human's ability to distinguish synthesized
images from real images.
Their results include StyleNAT, but we note that the authors of this work have
no affiliation with Stein~\etal nor have contacted them in any way.
Their results demonstrate that FID and other metrics are generally unreliable
when predicting the human error rate for distinguishing synthesized images from
real ones.
We refer to their work for a full discussion of metrics and for an independent 
evaluation of differing metrics.
We note that their model selection has a significant overlap with those
in~\Cref{table:exp-maintable}.
%in Table 3.
Additionally, we note that StyleNAT is a outlier in their results, and its
samples are the most difficult for humans to distinguish when compared to other
models.
While this result correlates with FID, they show that this is not true for many
other models and datasets.

%% file: sec/6_appendix/Visual.tex
\section{Visual Analysis and Artifacts}\label{app:visual_analysis}

Given the limitations to metrics, as discussed in~\Cref{app:metrics}, we
investigate the visual quality of synthesized images and look for different
biases that different networks exhibit.
We encourage the reader to not completely rely on the examples within this work
and to generate their own samples.
For this section we embed the highest quality images that we can, using pdf
embeddings, but note that the source images are in JPEG.
This allows readers to zoom in if using an electronic PDF reader.
Note that some artifacts may not be obvious at the standard resolution, but if
one zooms into 300\% or more then these artifacts will become much more apparent
and most readers will still be able to identify them after returning to their
normal reading settings.
We believe that such investigations are essential for the evaluation of
generative models and encourage the community to perform similar such studies. 

\subsection{StyleGAN3 vs StyleSwin vs StyleNAT}\label{app:visual_StyleGANoff}

First we perform a comparison between StyleGAN3, StyleSwin, and StyleNAT on
FFHQ-1024 samples.
We use FFHQ because there is universal familiarity with faces and humans 
are biologically primed to recognize faces, meaning that we are
more likely to notice subtle details where this may not be as true for images of
other classes of objects.
We believe that this makes it the best choice for identifying generation errors.
We also note that the FIDs of these networks are 2.79, 5.07, and 4.17, 
respectively.
If we rely exclusively on FID then we expect StyleGAN3 to be significantly
better than StyleNAT and StyleSwin, and for StyleNAT to be moderately better
than StyleSwin.
This is difficult to truely measure and would require a costly human study,
similar to Stein~\etal's~\cite{stein2023exposing}.
Instead we try to focus on the \emph{best possible} samples, and investigate the
visual artifacts.

We remind the reader that common locations for visual artifacts can be found
when looking at ears, eyes (including eyebrows, eyelashes, and pupils), the
neck, and hair.
In particular one can often quickly identify synthetic images by closely looking
at eyes.
We believe that all networks always produce artifacts that humans can easily
identify given sufficient training, but leave the subjective conclusions to the
reader. 
We note that we find a strong bias for high fidelity images to have simple
backgrounds with a Bokeh effect.
That is, where the image subject is sharp and in focus while the background is 
out of focus. 
This is achieved by using a high f-stop or large aperture size when taking a
photo.
We believe this is likely a dataset bias as this aesthetic is common for
professional portraits.
We will not focus on the background as when images like these are used for
creating deep fakes it is not uncommon for these to be replaced.
They are also a common source of errors, which the Bokeh style often makes
harder to identify.

\subsubsection{StyleGAN3}\label{app:va_stylegan3}

Karras~\etal~\cite{NEURIPS2021_076ccd93} has made public a set of curated
images\footnote{\href{https://nvlabs-fi-cdn.nvidia.com/stylegan3}{https://nvlabs-fi-cdn.nvidia.com/stylegan3}}, 
which we searched through.
We searched for the best quality image we could find and provide the link for
independent analysis. 
%As referenced in~\Cref{sec:rw-sg-architectures}, the authors note that a
The authors note that a
significant part of their work was in removing aliasing from images.
Aliasing being defined as overlapping frequency components that lead to
distortion and other potential spatial artifacts.

\begin{figure}
    \centering
    \begin{subfigure}[b]{\linewidth}
        \includegraphics[width=\linewidth]{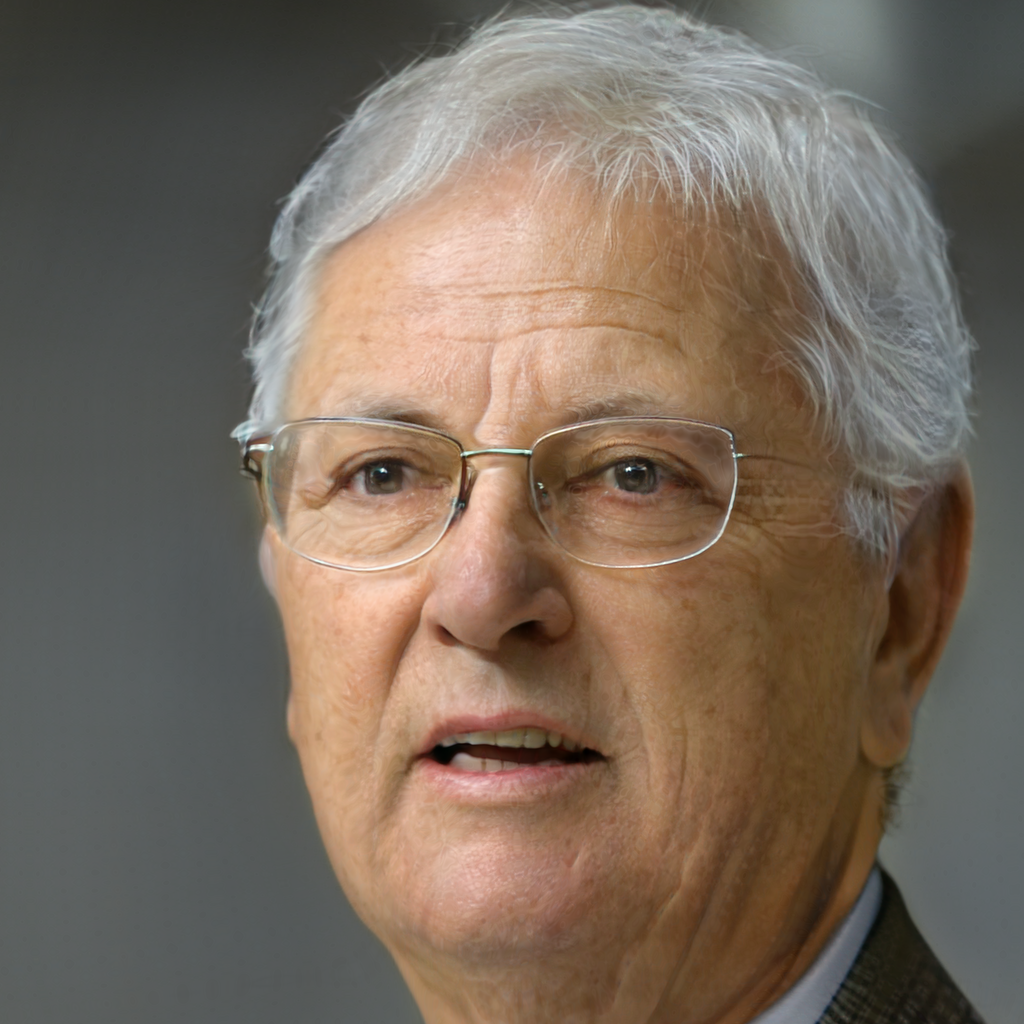}
        \caption{1024 FFHQ Sample from StyleGAN3}
        \label{fig:sg3samp}    
    \end{subfigure}
    \\
    \begin{subfigure}[b]{0.49\linewidth}
        \includegraphics[width=\linewidth]{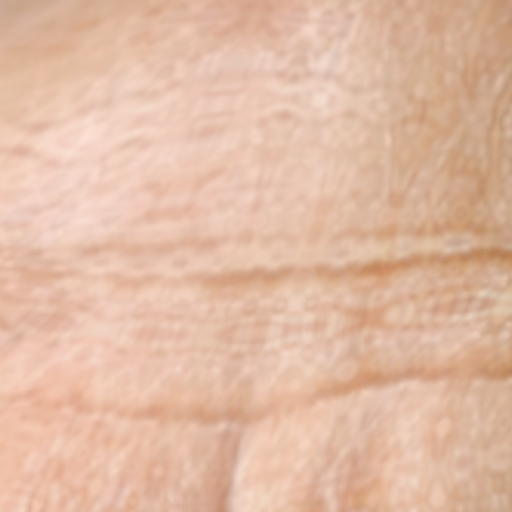}
        \caption{Forehead bead pattern. Two bands at top and bottom third.\phantom{We need more words}}
        \label{fig:sg3band}    
    \end{subfigure}
    \hfill
    \begin{subfigure}[b]{0.49\linewidth}         
        \includegraphics[width=\linewidth]{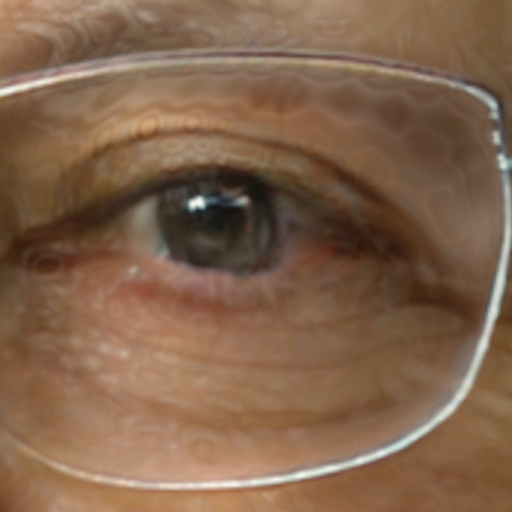}
        \caption{Glasses with hexagonal artifacts around edges. Upper right of glasses.}
        \label{fig:sg3glass}    
    \end{subfigure}
    \caption{Artifacts from StyleGAN3 FFHQ 1024 samples (sample 0068 from link). We show the banding effect that is common in StyleGAN3 photos, especially on foreheads, as well as hexagonal patterns that happen in glasses.}
    \label{fig:sg3}
\end{figure}

In~\Cref{fig:sg3} we show our best found sample and in
subfigures~\Cref{fig:sg3band}~and~\Cref{fig:sg3glass} we provide zoomed in
sections where we identify frequency based artifacts.
The most obvious artifact is the banding, which we show a zoomed in section
in~\Cref{fig:sg3band}, but this appears throughout the face.
Another horizontal band can be easily identified by the eye on the right side,
near the glasses (under the disappearing Temple) as well as vertical banding
across the neck.
In~\Cref{fig:sg3glass} we show a zoomed in section of the right glasses, where
there are clear hexagonal patterns.
Similar patterns can be found on the other side.
By careful inspection of the eyes we can notice that the pupils are irregular
and that neither the iris nor pupil is not circular, being more pronounced in 
the eye on the left side.

While these were not obvious at first glance to many some of our colleagues,
all were able to see if they zoomed in, and would continue to see them after
returning to their normal level of zoom.
We were unable to identify a single image within the curated collection that did
not exhibit similar artifacts. 
We were able to identify similar artifacts for all images we looked at,
irrespective of the resolution or dataset, including non-human images.
Thus we believe that this can serve as a reliable ``fingerprint'' for
identifying StyleGAN3 based models.

\subsubsection{StyleSwin}\label{app:va_styleswin}

\begin{figure}
    \centering
    \begin{subfigure}[b]{\linewidth}
        \includegraphics[width=\linewidth]{Images/swin/1024/original.png}
        \caption{1024 FFHQ Sample from StyleSwin}
        \label{fig:ssamp}    
    \end{subfigure}
    \\
    \begin{subfigure}[b]{0.49\linewidth}
        \includegraphics[width=\linewidth]{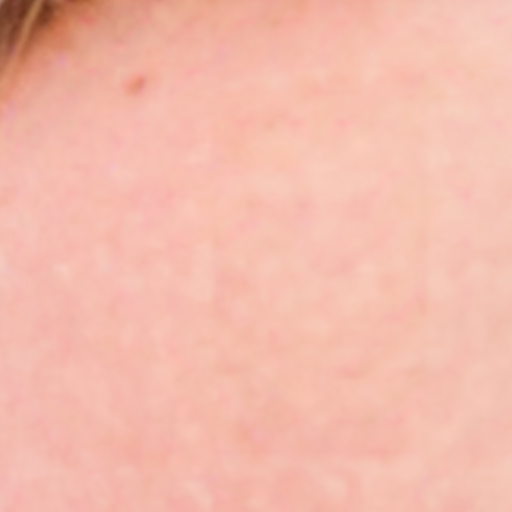}
        \caption{Forehead squares}
        \label{fig:ssband}    
    \end{subfigure}
    \hfill
    \begin{subfigure}[b]{0.49\linewidth}         
        \includegraphics[width=\linewidth]{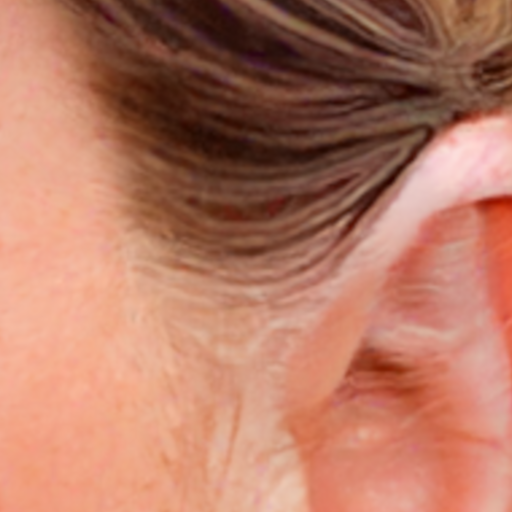}
        \caption{Right ear texture}
        \label{fig:ssear}    
    \end{subfigure}
    \caption{Artifacts from StyleSwin FFHQ 1024. (We generated these.)}
    \label{fig:va_styleswin}
\end{figure}

For StyleSwin~\cite{Zhang_2022_CVPR} we generated 50 samples from their official
checkpoint and picked the best sample, discarding any samples where there were
obvious large scale artifacts (colloquially refereed to as ``GAN monsters'').
These samples can be generated from our codebase using the StyleSwin
checkpoints.
Our selection is show in~\Cref{fig:va_styleswin}, where artifacts are more
apparent than in StyleGAN3.
We notice clear ``block'' like shapes throughout the face, as seen
in~\Cref{fig:ssband}. 
These are significantly different from those noted 
by~Zhang~\etal~\cite{Zhang_2022_CVPR}, which are more similar to pixelization.
Interestingly we do not observe the same pixelization visible in the samples
shown in Figures 3 and 5 of their paper, but ours look more similar to those 
shown in their header or in Figures 7, 10, 11, or 12.

In addition we notice more continuous patterns most easily seen by the ear,
\Cref{fig:ssear}, but also observable in the chin and near the eyes (not to be
confused with ``crow's feet'').
We also notice a large discrepancy between the eyes.
The image exhibits clear heterochromia (the iris are different colors), as well
as significantly different sizes.
Upon close inspection, it can be seen that the reflection within the eyes would
suggest the person is looking at two different scenes, with different lighting
conditions. 
Additionally, we notice a high rate of speckling in faces, with the easiest to
view one being the light yellow spot on the cheek.
Such artifacts are less obvious and may be confused with common skin blemishes
(e.g. sun spots).

These artifacts show a clear demonstration where the transformer does not
provide long-range coherence.
We also observe these patterns within the images shown in the paper, including
the aforementioned figures.
We provide further explination of these artifacts in 
~\Cref{app:attn_maps}.

\subsubsection{StyleNAT}\label{app:va_stylenat}

\begin{figure}
    \centering
    \begin{subfigure}[b]{\linewidth}
        \includegraphics[width=\linewidth]{Images/NA/original.png}
        \caption{1024 FFHQ Sample from StyleNAT}
        \label{fig:snatsamp}    
    \end{subfigure}
    \\
    \begin{subfigure}[b]{0.49\linewidth}
        \includegraphics[width=\linewidth]{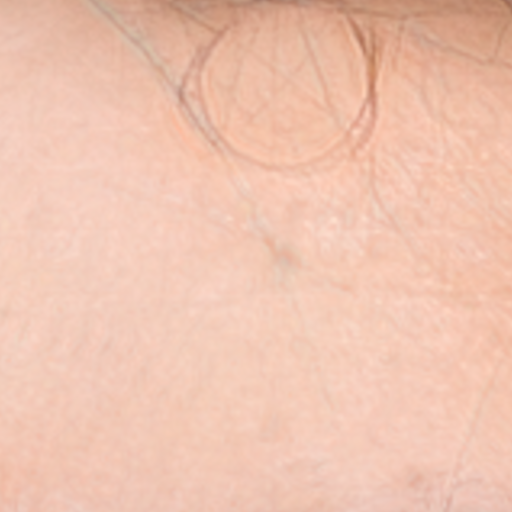}
        \caption{Forehead lines}
        \label{fig:snathead}    
    \end{subfigure}
    \hfill
    \begin{subfigure}[b]{0.49\linewidth}         
        \includegraphics[width=\linewidth]{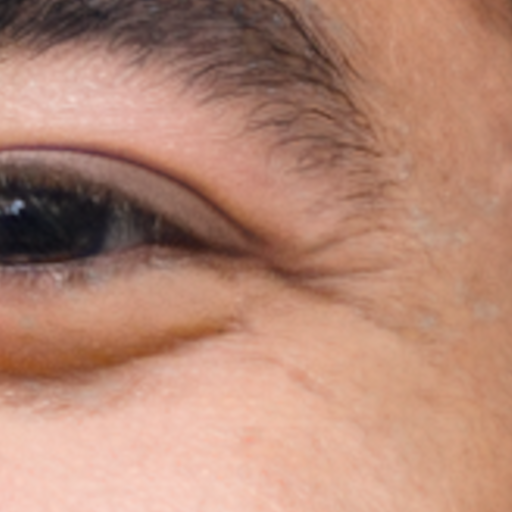}
        \caption{Right eye spotting}
        \label{fig:snateye}    
    \end{subfigure}
    \caption{Artifacts from StyleNAT FFHQ 1024. (We generated these.)}
    \label{fig:va_stylenat}
\end{figure}

For our image selection we follow the same procedure as with
StyleSwin~\Cref{app:va_styleswin}.
Our selection is shown in~\Cref{fig:va_stylenat}, and other examples can be seen
%by zooming in on our header figure or in~\Cref{fig:exp-ffhq256-samples} for
by zooming in on our header (\Cref{fig:header}) or in 
%Figure 5 
\Cref{fig:exp-ffhq256-samples}
for
samples from the lower resolution network.
Similar artifacts are visible within these images and some maybe more easily
seen in those examples.
For example, the subject's iris color is dark and may make investigation of the
iris and pupils more difficult for some.
We find heterochromia and distorted pupils less common among our samples but
observe that when it happens it is more likely to be Sectoral Heterochromia,
where the iris has multiple colors (as is referenced by the popular YouTuber
3Blue1Brown).
This can be seen in the central image of the header, in the eyes if the blond
hair girl.
Her eyes are predominantly blue, but show bits of brown, and there is an
unrealistic color closer to cerulean blue in the bottom corners, similar to
those of the fictional Fremen in Dune.
We find full heterochromia is quite rare, but an instance can be observed in the
header in the right most column of the FFHQ-256 samples.
Several other images have slightly distorted pupils.
Despite these artifacts, we believe that our samples perform better than others
within these areas and demonstrates our model's ability to learn long range
coherence. 

We do identify other artifacts, which we believe can be used to visually
fingerprint our model.
We also observe a banding like pattern, but that these are smoother lines and
may be easily confused with strands of hair or indentations of the skin.
We show a zoomed in section of the forehead in~\Cref{fig:snathead}.
Similar artifacts are more apparent on the left eye and near the smile lines of
the cheek (clearer on the right side).
We also observe some spotting artifacts, that are predominantly blue in color.
We illustrate this in~\Cref{fig:snateye}, showing the right eye, but this is
also visible between the eyebrows.
Similar artifacts appear in the man's beard, but may be easily confused with
gray hairs.
We also notice this chromatic aberrations within the strands of hair in the
forehead.
This can also be seen a bit in~\Cref{fig:snathead} but may require zooming in
depending on the screen used for viewing.
The speckling artifacts being similar to StyleSwin may be a result of the
underlying architecture, but further investigation is necessary.

We believe that these artifacts can serve as means to visually fingerprint our
model and distinguish it from others.
Notably, are lines appear to be an artifact of the method, and we are able to
view these within the attention maps~\Cref{app:attn_maps}.

%% file: sec/6_appendix/AttnMaps.tex
\section{Attention Maps}\label{app:attn_maps}
To help explain the observations we see throughout this work, we visualized the attention maps for both StyleSwin and StyleNAT.
We note that neither of these networks can have attention maps generated in the usual manner.
Our code for this analysis will also be included in our GitHub.
For both versions we can't extract the attention map from the forward network, as would be usually done, but instead extract both the query and key values.
Exact methods are explained in the respective FFHQ sections.
We believe that these maps demonstrate the inherent biases of the network and specifically demonstrate why StyleNAT, and critically the Hydra attention, result in superior performance.
Swin's shifted windows demonstrate a clear pooling, which may be beneficial in classification tasks, but not as much for generative tasks, which are more sensitive and unstable.
They also provide explanations upon where both networks may be improved within future works and we believe this tool will be valuable to other researchers in other domains.

We will look at both FFHQ and LSUN Church to try to determine the differences and biases of the networks and attention mechanisms.
For all of these we will generate a random 50 samples and select by hand representative images for the give tasks.
It is important to take care that there is a lot of subjectivity here and that these maps should only be used as guides into understanding our networks rather than explicit interpretations.
Regardless, the attention maps are still a helpful tool in determining features and artifacts in generation, as we will see below.
The patterns discussed were generally seen when looking at each of the sampled images during our curation.

Our attention maps suggest that these networks follow a fairly straight forward and logical method in building images.
In general we see that lower resolutions focus on locating the region of the main objects within the scene while the higher resolutions have more focus on the details of the images.
We see progressive generation of the images, that each resolution implicitly learns the final image in progressively detail.
This suggests that the progressive training seen in StyleGAN-XL may also benefit both of these networks.
The Style-based networks generally have two main feature layers (or blocks) per resolution level, which we similarly follow.
Our maps also suggest that a logical generation method is performed at the resolution level. 
Where the first layer generating the structure, realigning the image after the previous up-sampling layer.
The second layer generates more details at the resolution level.
This may suggest that a simple means to increasing fidelity would be to make each resolution level deeper, which is also seen in StyleGAN-XL.
In other words, fidelity directly correlates to the number of parameters, and thus it is necessary to incorporate that within our evaluation.
The goals of this work is on architecture and the changes that they make, rather than overall fidelity.
We leave that to larger labs with larger compute budgets.

To make reading easier we have placed all maps at the end of the document and provide detailed descriptions in each caption.

\subsection{FFHQ}
For FFHQ we will look at specifically the 1024 dataset and we will select our best sample.
We are doing this to help determine the differences in artifacts that we saw in~\Cref{app:model-arch}.
Since many of these features are fine points we will want to see the high resolution attention maps to understand what the transformers are concentrating on and how the finer details are generated.
Specifically, we use the same images that were used within the previous section to help us identify the specific issues we discussed.

\input{sec/6_appendix/Swin_FFHQ}

\input{sec/6_appendix/NA_FFHQ}

\subsection{Church Attention Maps}
To help us understand the differences in performances specifically in the LSUN Church dataset we also wish to look at the attention maps to help give us some clues.
We know that FID has limitations being that Inception V3 is trained on ImageNet-1k and uses a CNN based architecture.
ImageNet-1k is primarily composed of biological figures and so does not have many objects that have hard corners like LSUN Church.
Additionally, CNNs have a biased towards texture~\cite{geirhos2018imagenettrained,NEURIPS2020_db5f9f42}, which can potentially make the metric less meaningful, especially on datasets like this.
Since we had noticed that the Swin FFHQ attention maps had a bias to create blocky shapes and StyleNAT had a bias to create rounder shapes, we may wish to look into more detail to determine if these are biases of the architecture or that of the dataset.
We find that this is true for StyleSwin but not of StyleNAT.

% To make a fair comparison, as well as determine how we can use these maps to see how our networks are failing, we will again create 50 random samples from each network but this time we will pick a good and a bad sample from each.
To understand why this dataset provides larger difficulties for these networks we not only select a good sample, but also a bad sample, hoping to find where the model loses coherence.
We find that in general this happens fairly early on, with the networks having difficulties placing the ``subjects'' within the scene.
We see higher fidelity maps in the better samples but find that overall these struggle far more than on the FFHQ task.

We believe that our results here show that FID is not reliable for the LSUN Church dataset, as well as demonstrates that Church is a significantly harder generation problem for these models than FFHQ is.
This is claim is consistent with many of the aforementioned works, which present stronger cases and make similarly arguments for other metrics.
These demonstrate the need to perform visual analysis as well as feature analysis to ensure that the model is properly aligned with the goal of high quality synthesis rather than with the biases of our metrics.
Evaluation unfortunately remains a difficult task, where great detail and care is warranted.

Specifically, at low resolutions both networks have difficulties in capturing the general concept of the scene.
We believe that this is due to the increased variance and diversity of this dataset, compared to FFHQ.
While human centered faces share a lot of general features, such as a large oval centered in the image, this generalization is not true for the Church dataset, which a wide variety of differing building shapes, many different background objects to include (which we say FFHQ prefers simple backgrounds), and that the images are taken from many different distances.

StyleSwin samples are shown in~\Cref{fig:swin_church} and the StyleNAT samples are show in~\Cref{fig:snat_church}.
We believe that both these samples look on par with the quality of that of ProjectedGAN~\cite{NEURIPS2021_9219adc5}, which currently maintains SOTA on LSUN Church with an FID of 1.59, and thus are sufficiently ``good'' samples. 
We will look at the blocks sized 32 to 256, as we believe this is sufficient to help us understand the problems, but we could generate smaller maps.

It is unclear at this point if the fidelity could be increased simply by increasing the number of training samples or if additional architecture changes need to be made in order to resolve this (as suggested above).
We will specifically note that even SOTA generation on this dataset, ProjectedGAN~\cite{NEURIPS2021_9219adc5}, does not produce convincing fakes, while this task has been possible on FFHQ for some time, albeit not consistently. 
This is extra interesting considering that the SOTA FID on LSUN Church is 1.59, with 3 networks being below a 2.0 while SOTA FFHQ-256 (the same size) is 2.05 (this work) and scores as high as 3.8~\cite{NEURIPS2020_8d30aa96} frequently produce convincing fakes.
We also remind the reader that while ProjectedGAN performs well on Church (1.59), it does not do so on FFHQ (3.46), 
%see Tables 1 and 2.
%\Cref{table:ffhq,tab:church}.
%\Cref{tab:exp-maintable}.
Table 3.

\input{sec/6_appendix/Swin_Church}
\input{sec/6_appendix/NA_Church}

\input{Includes/Figures/na_ffhq_attn_maps}
\input{Includes/Figures/swin_ffhq_attn_maps}
\input{Includes/Figures/na_church_attn_maps}
\input{Includes/Figures/swin_church_attn_maps}

%% file: sec/6_appendix/Swin_FFHQ.tex
\subsubsection{(FFHQ) StyleSwin}\label{app:swin_ffhq}

For StyleSwin we extract the query and key values from each forward layer (note
that there are two attentions per resolution level for StyleGAN based
networks). We perform this for each split window which has shape
$[B\frac{H}{w_s}\frac{W}{w_s}, n_h, w_s^2, C']$, where $B$ is the batch, $W,H$
are the height and width, $w_s$ is the window size, $n_h$ is the number of
heads, and $C'$ is the number of channels. We concatenate along the split heads
and then reverse the windowing operation by re-associating the windows with the
height and width. Once this is done we can mean the pixel dimensions for the
query and flatten them for the key (q is unsqueezed for proper shaping). We then
can obtain a normal attention map where we have an image of dimensionality 
$B, n_h, H, W$.

We will look at this attention map for the same sample as in
\Cref{fig:va_styleswin}. We break these into multiple figures so that they fit
properly with \Cref{fig:swin_ffhq_1024} representing the $1024\times1024$
resolution, \Cref{fig:swin_ffhq_512} the $512\times512$,
\Cref{fig:swin_ffhq_256128} representing both the $256\times256$ and
$128\times128$, \Cref{fig:swin_ffhq_6432} the $64\times64$ and $32\times32$,
and finally \Cref{fig:swin_ffhq_168} representing the $16\times16$ and
$8\times8$ resolutions. Note that the second half of the heads represents a
shifted window, per the design specified in their paper. In these feature
maps we see consistent blocking happening, which is indicative of the issues
with the Swin Transformer~\cite{Liu_2021_ICCV}. This also confirms the
artifacts and texture issues we saw in the previous section. These artifacts
can even be traced down to the 64 resolution level,
\Cref{fig:swin_ffhq_6432}. We believe that this is a particularly difficult
resolution for this network as it has more blocking in the second layer than
others.

We also notice that in the earlier feature maps that StyleSwin has difficulties
in picking up long range features, such as ears and eyes. This likely confirms
the authors' observations of frequent heterochromism (common in GANs),
mismatched pupil sizes, and differing ear shapes. 
%
% With respect to the errors seen in \Cref{fig:ssamp} we see some corresponding
% attention focusing on the same forehead region in
% \Cref{fig:swin_ffhq_512,fig:swin_ffhq_256128}. We also see similar focusing on
% the ears and hair in the same figures.

% We also notice that the different heads have significantly differing levels of
% attention.
At higher resolutions (128 and above) we find that the network struggles with
texture along the face despite establishing the general features. This is seen
by half the heads being dark and the other half being bright, as is seen in
\Cref{fig:swin_ffhq_512,fig:swin_ffhq_256128}. This does suggest some
under-performance from the network, with one set of heads doing significantly
more work when compared to the others.
% We also see that in the low level features the first layer is often
% blocky~\Cref{fig:swin_ffhq_512,fig:swin_ffhq_6432}. This likely explains why
% StyleSwin is unable to produce high quality samples at scale, as the features
% start to break down when scaling to the next resolution level.
We also see higher blocking, especially in the first layer, at lower
resolutions, indicating difficulties in acquiring the general scene structure.
This warrants more flexibility, such as that offered by Hydra.

%% file: sec/6_appendix/NA_FFHQ.tex
\subsubsection{(FFHQ) StyleNAT}\label{app:na_ffhq}

For StyleNAT we perform a similar operation as to StyleSwin.
% We also generate 50 samples and pick the best one.
We similarly extract the queries and keys, mean over the query's pixel
dimensions (unsqueezing), and flattening the key's pixel dimensions. We
similarly get back an image of shape $[B, n_h, H, W]$, with similar dimension
definitions. We break these into multiple figures so that they fit properly with
\Cref{fig:nat_ffhq_1024} representing the $1024\times1024$ resolution,
\Cref{fig:nat_ffhq_512} the $512\times512$, \Cref{fig:nat_ffhq_256128}
representing both the $256\times256$ and $128\times128$,
\Cref{fig:nat_ffhq_6432} the $64\times64$ and $32\times32$, and finally
\Cref{fig:nat_ffhq_168} representing the $16\times16$ and $8\times8$
resolutions. The first half of the heads has no dilation and the second half
has dilations corresponding with the architecture specified in
\Cref{tab:2split_model_arch}.

We see that in the high resolution images that NA is learning textures and long
range features across the face. This supports the claim that transformer
mechanisms are adequately learning these long range features, as should be
expected, and would support more facial symmetry that would be seen in human
faces. 

In the first layer we notice that more local features are being learned, which
explains the better textures seen in our samples. Particularly we notice in
\Cref{fig:nat_ffhq_1024,fig:nat_ffhq_512} that the first 2 heads learn feature
maps on the main part of the face. Noting that the first two heads represent
$7\times7$ kernels that are not dilated. We also noticed some swirling patterns
in the second half of heads, which correspond to dilated neighborhood attention
mechanisms. These correspond to the soft lines we saw above, and act like edge
detectors. The second layer does a better job at finer detail and removes many
of these, tough they are still visible on the chin. We notice that these
particularly appear around hair and may be reasoning that the hair quality of
our samples perform well. The long range, dilated, features all do tend to learn
long range features and aspects like backgrounds, as we would expect.
% We did also observe that backgrounds of samples frequently were better than
% other models, but this is difficult to measure.

In the lower resolutions we see that these attention maps learn more basic
features such as noses, ears, and eyes, which helps resolve many of the issues
faced by CNN based GANs. Details such as eyes and mouth can be identified even
at the 16 resolution image~\Cref{fig:nat_ffhq_168}! The authors noticed that
while generating they observed lower rate of heterochromia (different eye
        colors), which are common mistakes of GANs. This is difficult to
quantify as it would unlikely be caught by metrics such as FID but we can see
from the attention maps that the early focus on eyes suggests that this
observation may not be purely speculation. 

We believe that these attention maps demonstrate a strong case for StyleNAT and
more specifically our Hydra Neighborhood Attention. That small kernels can
perform well on localized features, like CNNs, but that our long range kernels
can incorporate long range features that we'd want from transformers. We can
also see from the feature maps that the mixture of heads does support our desire
for added flexibility. This is done in a way that is still efficient
computationally, having high throughputs, training speed, and a low requirement
on memory.

% While our analysis is limited to FFHQ we note that this same procedure is also useful for other types of data and may help users determine the proper configurations.
% While StyleNAT's Hydra layer added more flexibility, we also demonstrate a way to interpret the hyperparameter choices and how this can be a highly desirable tool for many users.

%%%%%%%%%%%%%%%%%%%%%%%%%%%%%%%%%%%%%%%%%%%%%%%%%%%%%%%%%%%%%%%%%%%%%%%%%%%%%%%%
%%%%%%%%%%%%%%%%%%%%%%%%%%%%%%%%%%%%%%%%%%%%%%%%%%%%%%%%%%%%%%%%%%%%%%%%%%%%%%%%
%%%%%%%%%%%%%%%%%%%%%%%%%%%%%%%%%%%%%%%%%%%%%%%%%%%%%%%%%%%%%%%%%%%%%%%%%%%%%%%%
%%%%%%%%%%%%%%%%%%%%%%%%%%%%%%%%%%%%%%%%%%%%%%%%%%%%%%%%%%%%%%%%%%%%%%%%%%%%%%%%

%% file: sec/6_appendix/Swin_Church.tex
\subsubsection{(Church) StyleSwin}
\label{sec:swin_church_attnmap}

\begin{figure*}[ht]
 \centering
    \begin{subfigure}[b]{0.49\linewidth}         
        \includegraphics[width=\linewidth]{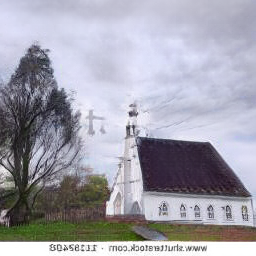}
        \caption{Good}
        \label{fig:swin_church_good}    
    \end{subfigure}
    \begin{subfigure}[b]{0.49\linewidth}         
        \includegraphics[width=\linewidth]{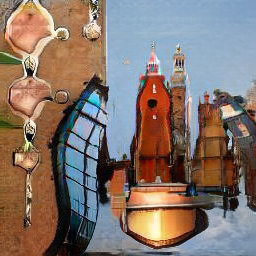}
        \caption{Bad}
        \label{fig:swin_church_bad}    
    \end{subfigure}
    \caption{Church good and bad samples from StyleSwin. Good example has a clearly visible church and tree with a good distinction of foreground and background. Good example has a fairly legible citation but no other watermarks. Bad example has lost global structure but does maintain church like features.}
    \label{fig:swin_church}
\end{figure*}

For the StyleSwin generated images, \Cref{fig:swin_church}, we can see that the
good image looks nearly like a Shutterstock image, almost reproducing a mirrored
image and where the text is almost legible. But in this we also see large
artifacts, like the floating telephone poll, the tree coming out of a small
shed, or other distortions. We believe that this telephone poll may actually be
part of a watermark, but are unsure. In the bad image, we see that there was a
mode failure and specifically that the generation lost track of the global
landscape. Even with this failure, we still do see church like structures, such
as a large distorted window, making this image more of a surrealist
interpretation of a church than a photograph. These images will thus provide
good representations for understanding these two modes, of why good church
images are still hard to generate and why they completely fail.

In short, we find that at all levels, there is higher blockiness and less detail
captured by the attention mechanisms when compared to FFHQ. We see either very
high or very low activations with the inability to focus on singular tasks. At
low resolutions we see difficulties capturing structure and that this error
propagates through the model.

\Cref{fig:swin_church_256} shows our full resolution images, and in the
attention maps we can again see the same blocky/pixelated structures that we
found in the FFHQ investigation, but at a higher rate. For the good images, in
the first layer we see that the first head is performing an outline detection on
the scene, almost like a Sobel filter. We also see that the last head is
delineating the boundary between the foreground and background, particularly the
sky. Interestingly this appears almost like Pointillism, which we see in many
following maps as well. This is likely bias from the shifted windows. In the
second layer we see more fine grained structure, but interestingly we do not see
as much as we saw in the FFHQ version at the same resolution,
\Cref{fig:swin_ffhq_256128}, which suggests that this is a more difficult task
for this network. We can also see that this level is concentrating on the text
at the bottom of the image, which is not as clearly visible in the smaller
resolution maps. Notably, we do not clearly see the floating telephone pole or
the wires in the sky. These could be formed from another part of the network,
such as the RGB or MLP layers, but we have not investigated this. Further
investigation is needed to understand the contributions of these layers.
%
% Interestingly we see some of the same patterns within the bad image framework,
% where even the total brightness of the maps per head are similar, suggesting
% that these individual heads are learning specific tasks. While the higher
% resolution attention maps are interested at seeing finer detail, they likely
% do not give a good explanation at the collapse we see here.

Moving down to the 128 resolution images in~\Cref{fig:swin_church_128} we see
that the attention maps overall get much messier. For the good image we can see
that the roof of the church is picked up by many of the heads. In the second
layer, on the second head, we also see a clear filter looking at the tree and
roof of the church, which we can also see a less clear selection in head 6 and
the first layer at head 5. These same heads provide decent filters for the bad
images, the layer 1 head 5 and layer 2 head 2 seeming to do the best at object
filtering.

Looking at the 64 resolution~\Cref{fig:swin_church_64} and 32
resolutions~\Cref{fig:snat_church_32} we can more clearly see where the problems
are happening. In FFHQ the 64 resolution maps, \Cref{fig:swin_ffhq_6432}, is
where we start to first see our main object with relative details and the 32
resolution has a decent depiction of broad shape. We do not have as good of an
indication within these maps, where the 64 resolution images do not have clearly
identifiable building textures, let alone building shapes. This is even worse at
the 32 resolution image level.

Interestingly, at this resolution it is difficult to distinguish which version
would generate the good or bad image, which doesn't seem distinguishable till at
least the 128 resolution. These aspects suggest that the generation of this data
is substantially harder for this model. This is extra interesting considering
that the FID scores are fairly close for both of these datasets, with FFHQ being
2.81 and Church being 2.95. With more difficulties in capturing general
structure the network then struggles to increase detail and this systematic
issue cannot be resolved. This network saw trained on 1.5M iterations.

%% file: sec/6_appendix/NA_Church.tex
\subsubsection{(Church) StyleNAT}\label{sec:snat_church_attnmap}

\begin{figure*}[ht]
 \centering
    \begin{subfigure}[b]{0.49\linewidth}         
        \includegraphics[width=\linewidth]{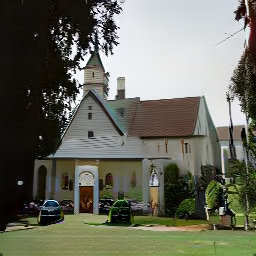}
        \caption{Good}
        \label{fig:snat_church_good}    
    \end{subfigure}
    \begin{subfigure}[b]{0.49\linewidth}         
        \includegraphics[width=\linewidth]{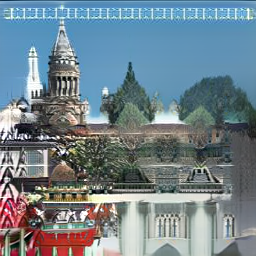}
        \caption{Bad}
        \label{fig:snat_church_bad}    
    \end{subfigure}
    \caption{Church good and bad samples from StyleNAT. Good sample has a clearly visible church, with cars out front and a clear distinction between foreground and background. While the bad sample distinguishes foreground and background, it is unable to correctly connect a coherent image of a church.}
    \label{fig:snat_church}
\end{figure*}

StyleNAT performs significantly worse at LSUN Church, and it isn't clear why this is.
Determining if this is a limitations of the metric, which may be biased to these features~\cite{kynkaanniemi2023the}, we must explore a bit deeper.
For these attention maps we will use the model that generated visible text and what the authors thought were higher quality.
This network uses smaller kernel sizes of 3 and has a max dilation rate of 8.
Thus the dilations are [[1],[1][1,2],[1,2,4],[1,2,4],[1,2,4],[1,2,4,8]].
This is the same configuration as when higher overfitting was observed.
Since higher overfitting tends to correspond to higher fidelity we want to investigate what went right, to improve the work.
The good image here is on par with that of Swin, and SOTA works, but the bad image is again a surrealist work wherein we see a agglomeration of a ``Church."
The good image appears pixelated, has scan lines, and some other distortions such as the car being reflected and turned into a bush.
The bad image seems to incorporate nearly every feature within the dataset, including churches, towers, temples, cathedrals, as well as many different trees all smashed together Cronenberg style.

In short, we find that StyleNAT is in fact able to generate hard lines, as this dataset is biased towards, but does tend to prefer smoother features.
We also see that at even the early stages that the scene has difficulties capturing global coherence.
This likely explains the instabilities we faced and why training often diverged fairly early on, with nearly a fifth of the number of iterations as FFHQ and nearly 10\% of StyleSwin.

The 256 resolution attention maps,~\Cref{fig:snat_church_256}, images we immediately see that some of the attention heads to not have rounder features, indicating that our network does not have a significant bias towards biological shapes.
In the first level, the first three attention heads have what appear to be scan lines, which we do see manifest in the full image.
We also see traces of this in the next three heads, as well as most of the heads in the second layer.
It appears that in this case, this level is looking a lot at texture, similar to that in FFHQ~\Cref{fig:nat_ffhq_256128}.
An interesting feature here, clearer in the first layer in heads 4-8, is that the we see what looks like the skeleton of a tree with branches coming out, almost centered at where the actual tree is in the main picture (left).
What is notable here is that neither the trunk nor the branches are visible in the generated image, and that the ``imagined'' trunk is a bit translated from where we may predict it would be on the ``actual'' tree.

In other maps that we generated, that aren't shown, we noticed this pattern is extremely frequent when trees exist in the scene and there exists identical structure when the tree does not have foliage.
This includes the circular shapes adjacent to the trunks.
We did not notice this feature when only the foliage is visible, where the tree may look more like a bush, such as in the bad sample.
We did not notice such skeletons as prevalent in the Swin version, although the best example can be seen in head 4 of the 256 layer in~\Cref{fig:swin_church_256}, but this appeared to be an exception rather than the norm.
We are careful to make a conclusion that the network has classified trees and understands their skeletal structure and note that a reasonable alternative explanation is that this trunk looking figure can just as easily be a guide for distinguishing the location of the tree.

There is also a notable difference in the attention maps between levels.
In general we believe these show that the first layer is working more on the general structure of the scene while the second layer is improving detail.
We also saw such correlations within the FFHQ analysis.
We believe that this is a reasonable guess because the first level follows an upsampling layer and thus the network needs to first re-establish structure of the scene before it can provide detail.
We also believe that this happens within the Swin based generator as well.
This can mean that potentially higher fidelity generators can add additional layers, and that this is more necessary at higher resolutions. 

As for the reasons for the low quality generations, we notice that the scenes in the 32 and 64 resolution, \Cref{fig:snat_church_32,fig:snat_church_64}, levels have potentially suggestive attention maps.
Particularly we notice that the bad quality image has much more chaotic attention maps.
Interestingly, we also see the scan lines

%% file: Includes/Figures/na_ffhq_attn_maps.tex
% NAT FFHQ 1024
\begin{figure*}[ht]
    \centering
    \begin{subfigure}[b]{0.49\linewidth}         
        \includegraphics[width=\linewidth]{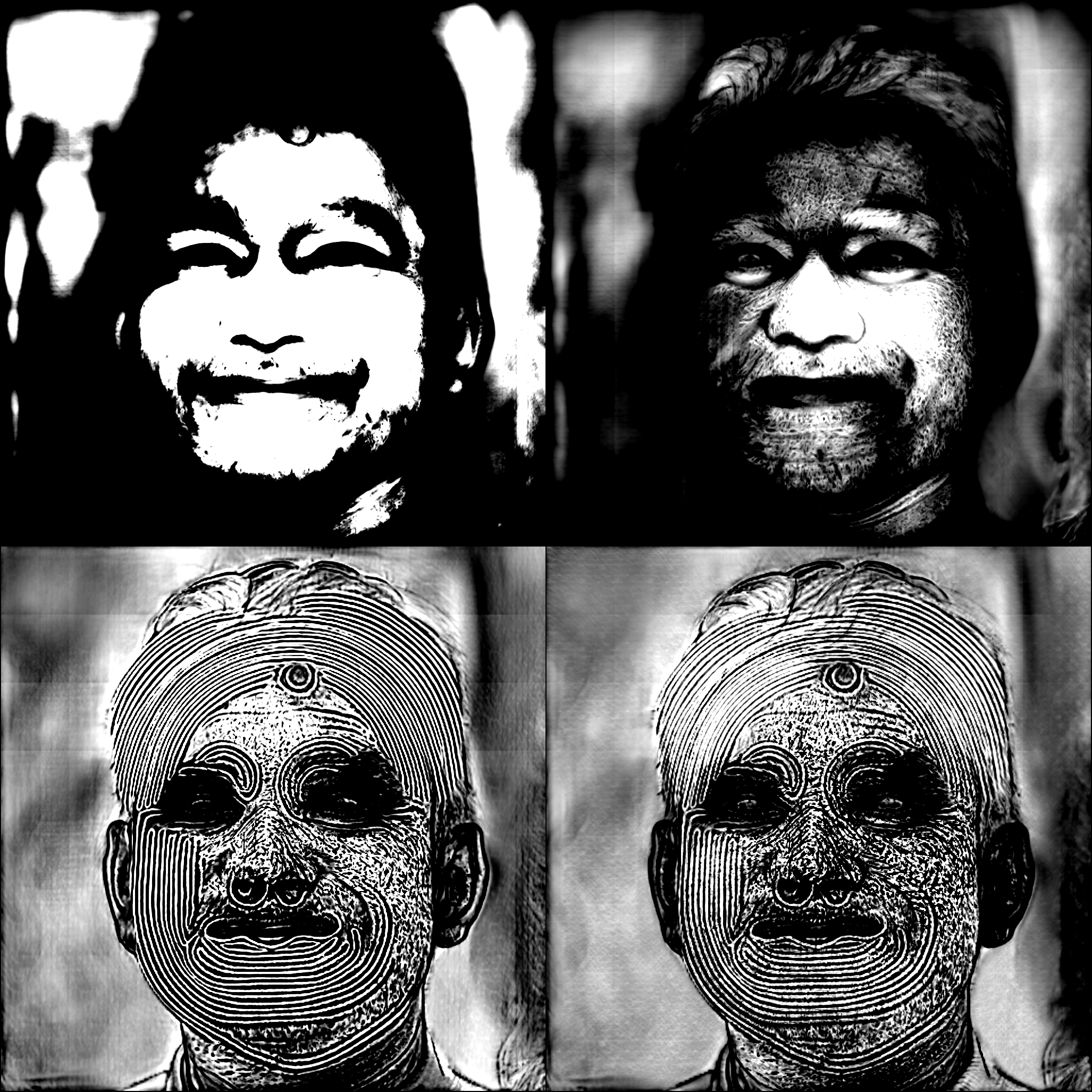}
        % \caption{First Attention: 4 heads}
        \caption{1024 Level, Layer 1}
        \label{fig:l8a0}    
    \end{subfigure}
    % \\
    \begin{subfigure}[b]{0.49\linewidth}         
        \includegraphics[width=\linewidth]{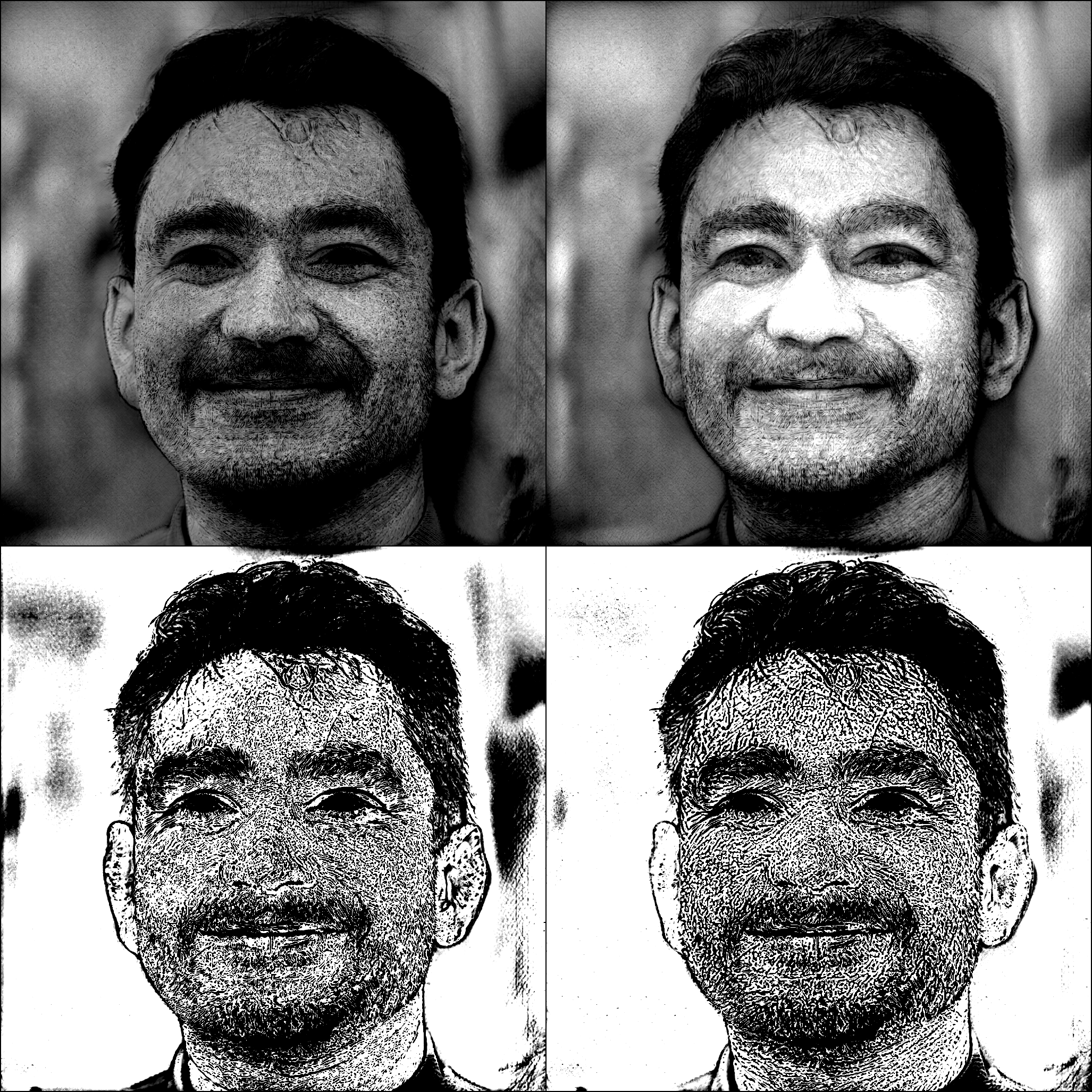}
        % \caption{Second Attention: 4 heads}
        \caption{1024 Level, Layer 2}
        \label{fig:l8a1}    
    \end{subfigure}
    % \caption{1024 Resolution Level for StyleNAT. Each has 4 heads with kernel size of 7 but the last 2 heads have a dilation of 128}
    \caption{FFHQ StyleNAT attention maps at the level with 1024 resolution. Every layer has 4 heads with kernel size of 7, but the last 2 heads have a dilation of 128. First layer concentrates more on structure and second more on texture. Heads without dilations appear to focus more on texture and the face.}
    \label{fig:nat_ffhq_1024}
\end{figure*}
% 512
\begin{figure*}[ht]
    \centering
    \begin{subfigure}[b]{0.49\linewidth}         
        \includegraphics[width=\linewidth]{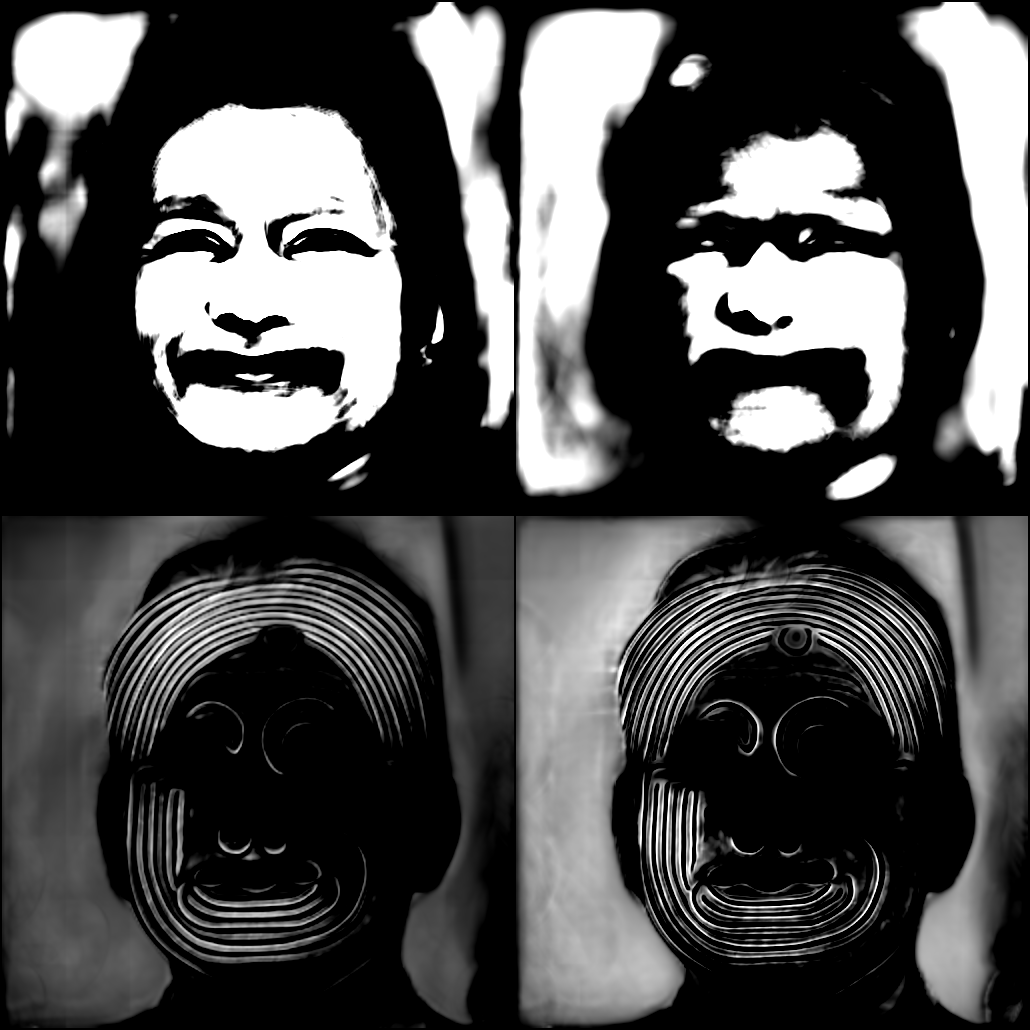}
        % \caption{First Attention: 4 heads}
        \caption{512 Level, Layer 1}
        \label{fig:l7a0}    
    \end{subfigure}
    % \\
    \begin{subfigure}[b]{0.49\linewidth}         
        \includegraphics[width=\linewidth]{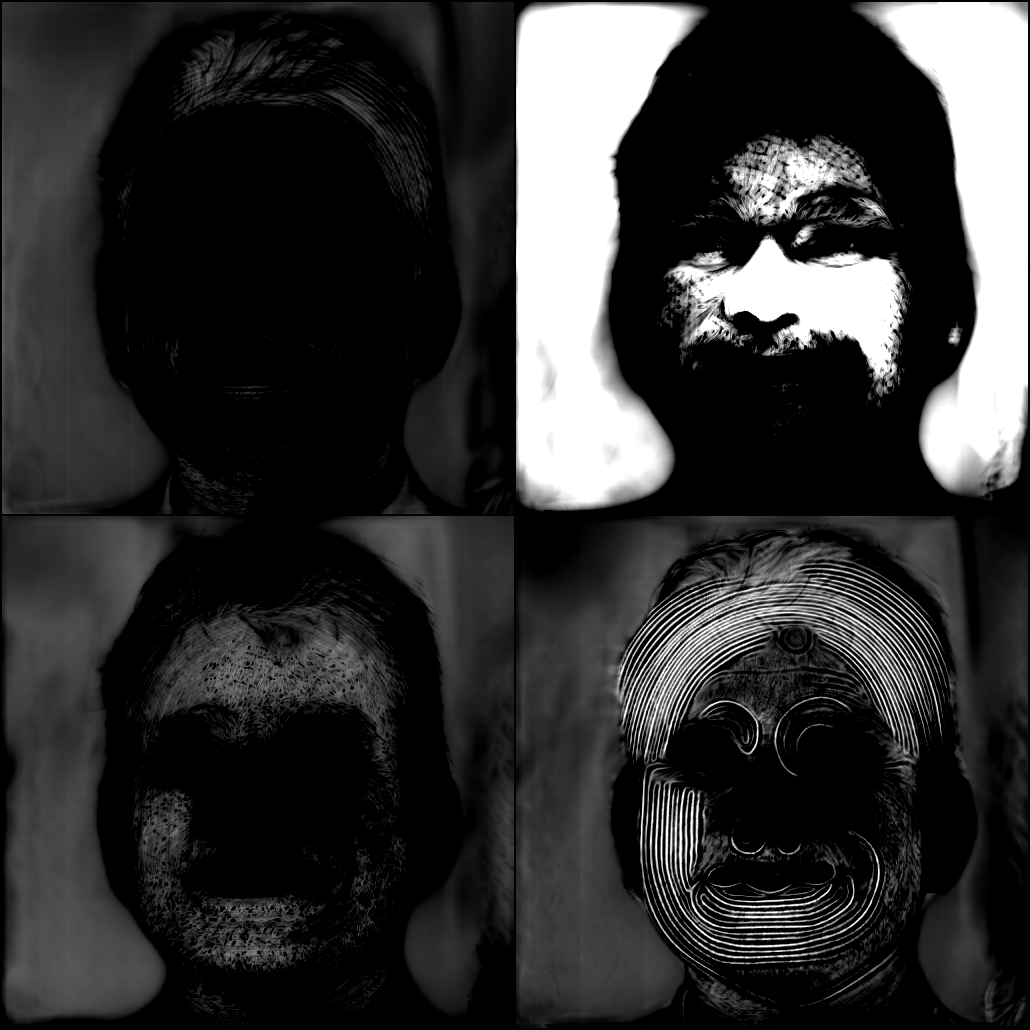}
        % \caption{Second Attention: 4 heads}
        \caption{512 Level, Layer 2}
        \label{fig:l7a1}    
    \end{subfigure}
    % \caption{512 Level for StyleNAT. Each has 4 heads with kernel size of 7 but the last 2 heads have a dilation of 64}
    \caption{FFHQ StyleNAT attention maps at the level with 512 resolution. Every layer has 4 heads with kernel size of 7, but the last 2 heads have a dilation of 64. First layer appears to focus more on structure, with no dilations concentrating on the face. Dilated heads focus on head shape.}
    \label{fig:nat_ffhq_512}
\end{figure*}
\clearpage

% 256 x 128
\begin{figure*}[ht]
    \centering
    \begin{subfigure}[b]{0.49\linewidth}         
        \includegraphics[width=\linewidth]{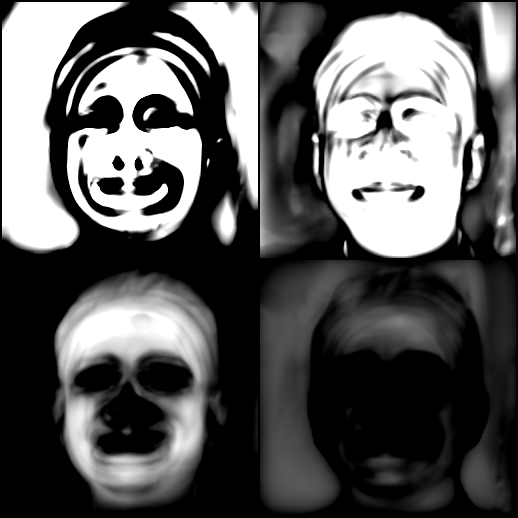}
        % \caption{256 First Attention: 4 heads}
        \caption{256 Level, Layer 1}
        \label{fig:l6a0}    
    \end{subfigure}
    \begin{subfigure}[b]{0.49\linewidth}         
        \includegraphics[width=\linewidth]{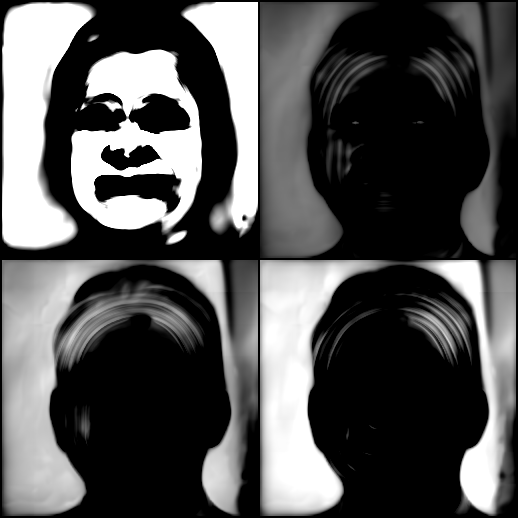}
        % \caption{256 Second Attention: 4 heads}
        \caption{256 Level, Layer 2}
        \label{fig:l6a1}    
    \end{subfigure}
    \\
    \begin{subfigure}[b]{0.49\linewidth}         
        \includegraphics[width=\linewidth]{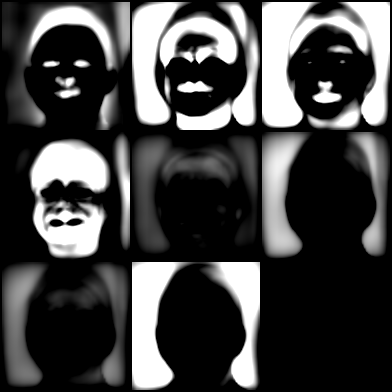}
        % \caption{128 First Attention: 4 heads}
        \caption{128 Level, Layer 1}
        \label{fig:l5a0}    
    \end{subfigure}
    \begin{subfigure}[b]{0.49\linewidth}         
        \includegraphics[width=\linewidth]{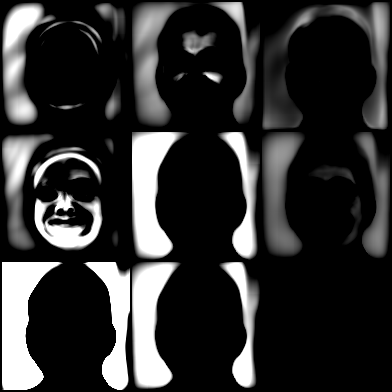}
        % \caption{128 Second Attention: 4 heads}
        \caption{128 Level, Layer 2}
        \label{fig:l5a1}    
    \end{subfigure}
    \caption{FFHQ StyleNAT attention maps at levels with 128 and 256 resolution. Every layer in each 32\texttimes{}32 level has 16 heads, and every layer in each 64\texttimes{}64 level has 8 heads, all with kernel size of 7. The second half of the heads  have dilations 16 and 32 respectively. General structure is visible and it can be seen we capture long range features.}
    \label{fig:nat_ffhq_256128}
\end{figure*}
\clearpage
% 64 x 32
\begin{figure*}[ht]
    \centering
    \begin{subfigure}[b]{0.49\linewidth}         
        \includegraphics[width=\linewidth]{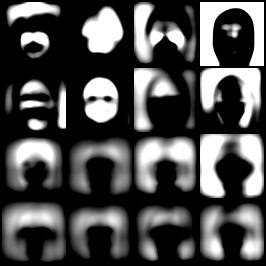}
        \caption{64 Level, Layer 1}
        \label{fig:l4a0}    
    \end{subfigure}
    \hfill
    \begin{subfigure}[b]{0.49\linewidth}         
        \includegraphics[width=\linewidth]{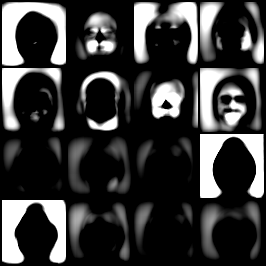}
        \caption{64 Level, Layer 2}
        \label{fig:l4a1}    
    \end{subfigure}
    \\
    \begin{subfigure}[b]{0.49\linewidth}         
        \includegraphics[width=\linewidth]{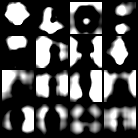}
        \caption{32 Level, Layer 1}
        \label{fig:lXa0}    
    \end{subfigure}
    \hfill
    \begin{subfigure}[b]{0.49\linewidth}         
        \includegraphics[width=\linewidth]{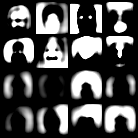}
        \caption{32 Level, Layer 2}
        \label{fig:lXa1}    
    \end{subfigure}
    \caption{FFHQ StyleNAT attention maps at levels with 32 and 64 resolution. Every layer in each level has 16 heads with kernel size of 7. The second half of the heads have dilations 4 and 8, respectively. Main structure visible in these resolutions, including eyes and the separation of face and hair.}
    \label{fig:nat_ffhq_6432}
\end{figure*}
\clearpage

% 16 x 8
\begin{figure*}[ht]
    \centering
    \begin{subfigure}[b]{0.49\linewidth}         
        \includegraphics[width=\linewidth]{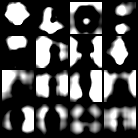}
        \caption{16 Level, Layer 1}
        \label{fig:l3a0}    
    \end{subfigure}
    \begin{subfigure}[b]{0.49\linewidth}         
        \includegraphics[width=\linewidth]{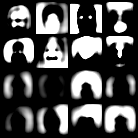}
        \caption{16 Level, Layer 2}
        \label{fig:l3a1}    
    \end{subfigure}
    \\
    \begin{subfigure}[b]{0.49\linewidth}         
        \includegraphics[width=\linewidth]{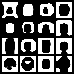}
        \caption{8 Level, Layer 1}
        \label{fig:l2a0}    
    \end{subfigure}
    \begin{subfigure}[b]{0.49\linewidth}         
        \includegraphics[width=\linewidth]{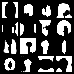}
        \caption{8 Level, Layer 2}
        \label{fig:l2a1}    
    \end{subfigure}
    \caption{FFHQ StyleNAT attention maps at levels with 8 and 16 resolution. Every layer in each level has 16 heads with kernel size of 7. The second half of the heads have dilations 1 and 2, respectively. The 8 resolution image looks to be focusing on placement of object within the scene, taking the general round shape and distinguishing subject from background.}
    \label{fig:nat_ffhq_168}
\end{figure*}
\clearpage

%% file: Includes/Figures/swin_ffhq_attn_maps.tex
% Swin FFHQ 1024
\begin{figure*}[ht]
    \centering
    \begin{subfigure}[b]{0.49\linewidth}         
        \includegraphics[width=\linewidth]{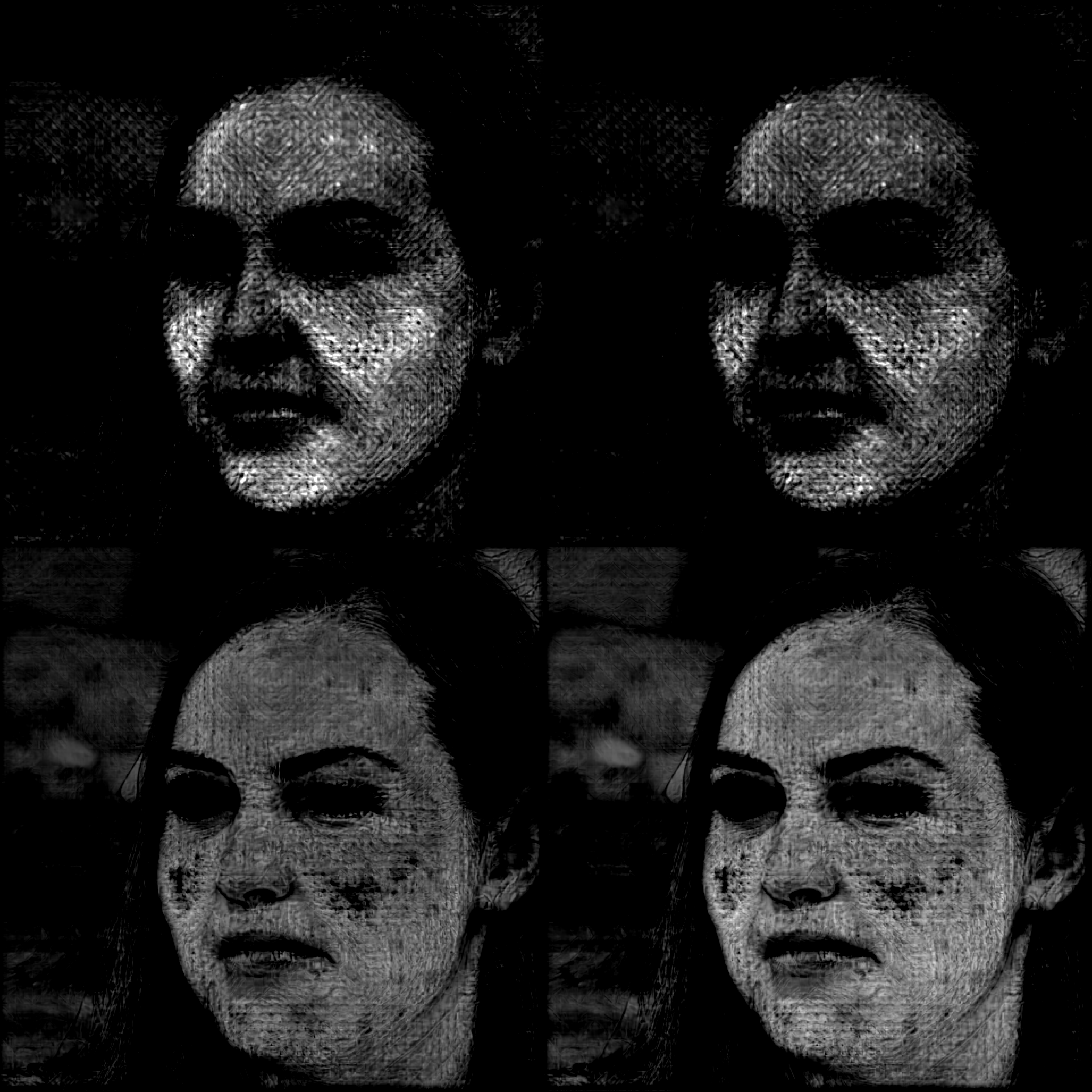}
        % \caption{First Attention: 4 heads}
        \caption{1024 Level, Layer 1}
        \label{fig:styleswin_l8a0}    
    \end{subfigure}
    % \\
    \begin{subfigure}[b]{0.49\linewidth}         
        \includegraphics[width=\linewidth]{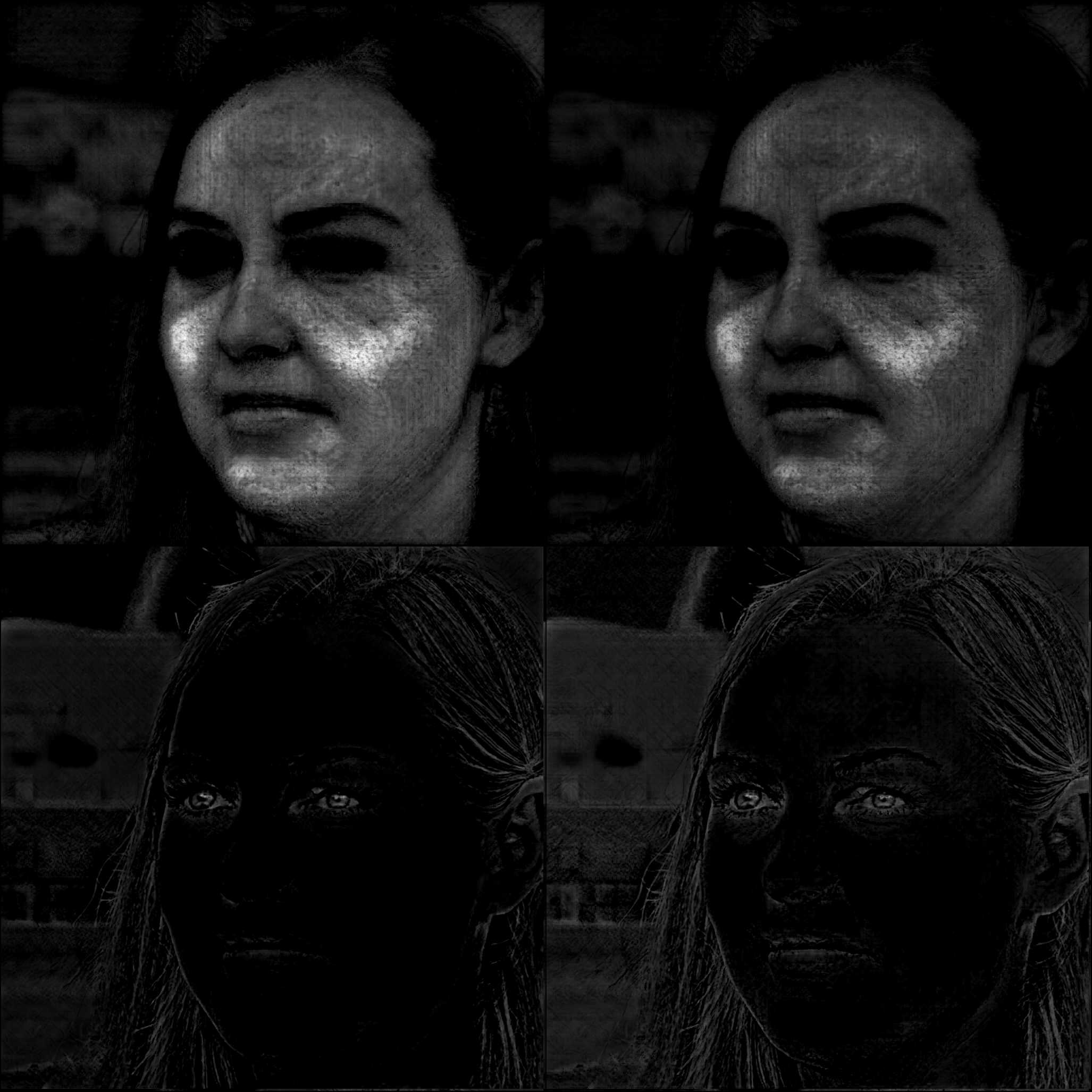}
        % \caption{Second Attention: 4 heads}
        \caption{1024 Level, Layer 2}
        \label{fig:styleswin_l8a1}    
    \end{subfigure}
    \caption{FFHQ StyleSwin attention maps at the level with 1024 resolution. Each layer has 4 heads with kernel size of 8 but half of them were trained with Shifted WSA. First level appears to concentrate on facial structure and texture. Second level appears to focus on symmetric features such as cheeks and eyes.}
    \label{fig:swin_ffhq_1024}
\end{figure*}
% Swin FFHQ 512
\begin{figure*}[ht]
    \centering
    \begin{subfigure}[b]{0.49\linewidth}         
        \includegraphics[width=\linewidth]{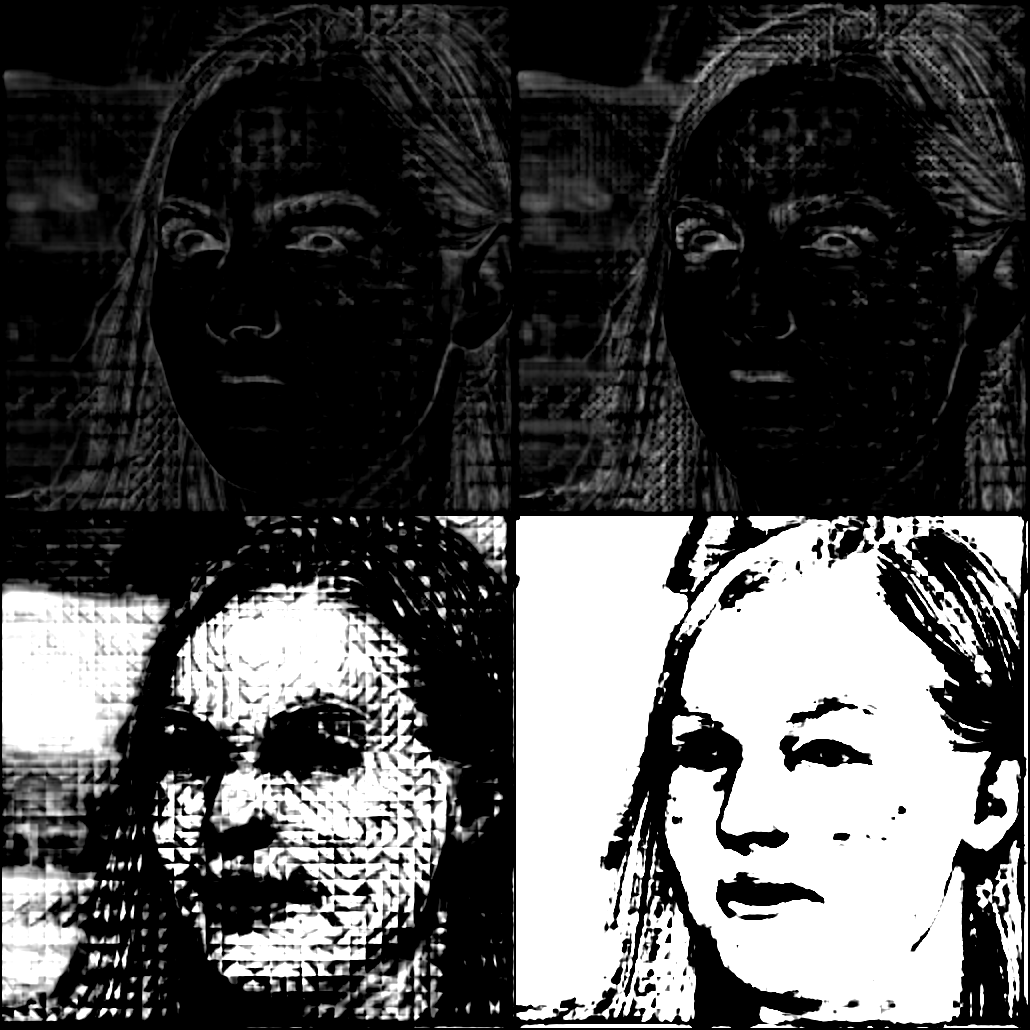}
        % \caption{First Attention: 4 heads}
        \caption{512 Level, Layer 1}
        \label{fig:styleswin_l7a0}    
    \end{subfigure}
    % \\
    \begin{subfigure}[b]{0.49\linewidth}         
        \includegraphics[width=\linewidth]{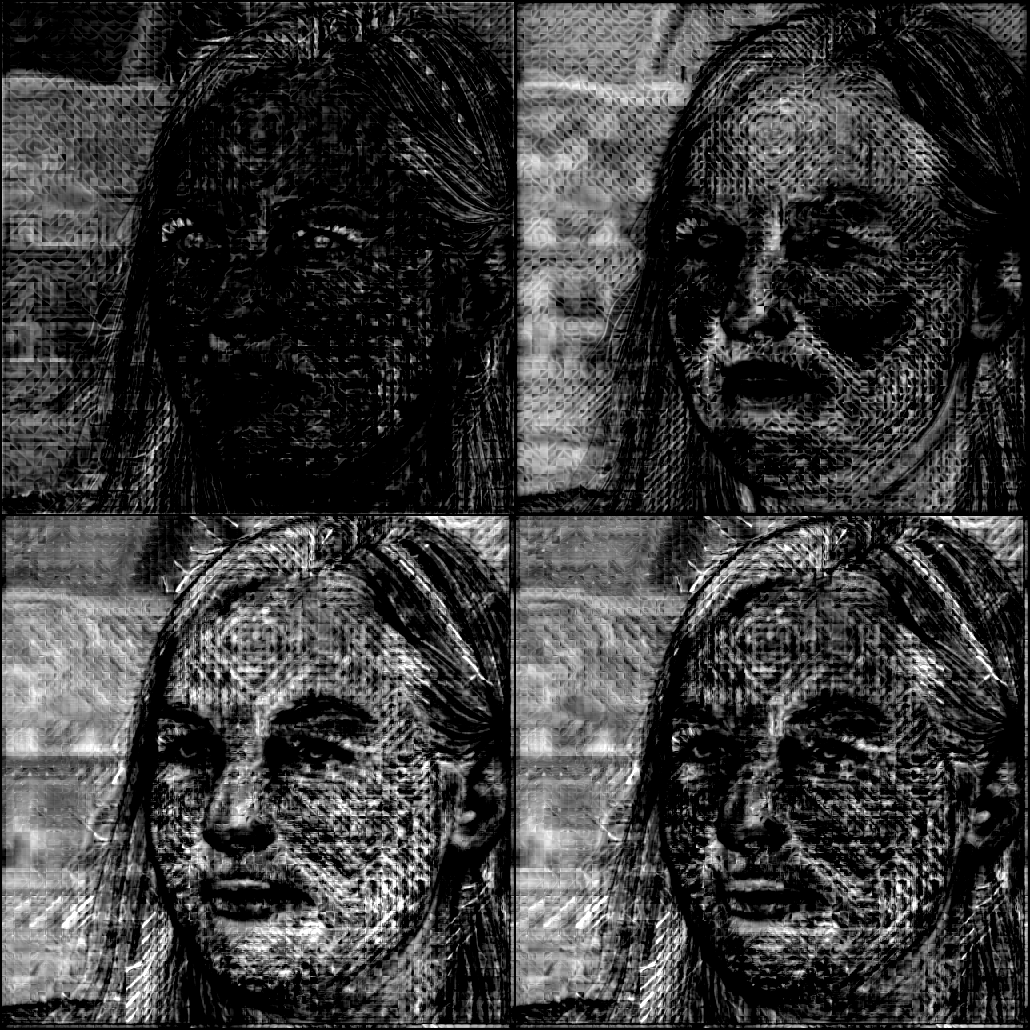}
        % \caption{Second Attention: 4 heads}
        \caption{512 Level, Layer 2}
        \label{fig:styleswin_l7a1}    
    \end{subfigure}
    % \caption{Attention maps at 512 Level for StyleSwin. Each has 4 heads with kernel size of 8 but the last 2 heads have a dilation of 64}
    \caption{FFHQ StyleSwin attention maps at the level with 512 resolution. Each layer has 4 heads with kernel size of 8 but half of them were trained with Shifted WSA. Generative artifacts are clearly visible on forehead in most maps. Heads have vastly different concentration levels.}
    \label{fig:swin_ffhq_512}
\end{figure*}
\clearpage
% Swin FFHQ 256 x 128
\begin{figure*}[ht]
    \centering
    \begin{subfigure}[b]{0.49\linewidth}         
        \includegraphics[width=0.98\linewidth]{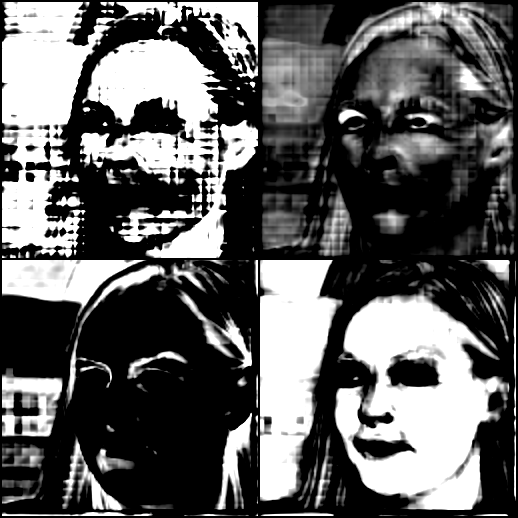}
        % \caption{256 First Attention: 4 heads}
        \caption{256 Level, Layer 1}
        \label{fig:styleswin_l6a0}    
    \end{subfigure}
    \begin{subfigure}[b]{0.49\linewidth}         
        \includegraphics[width=0.98\linewidth]{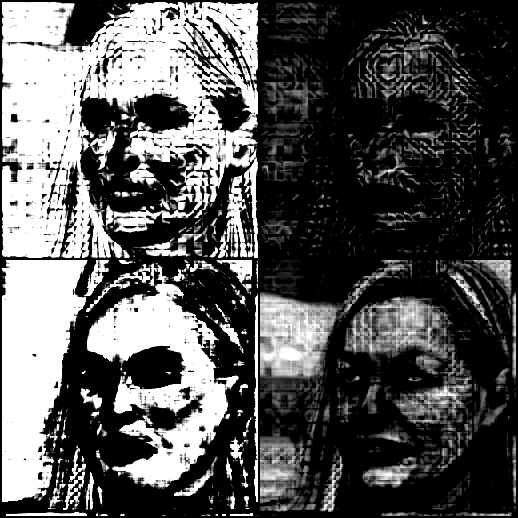}
        % \caption{256 Second Attention: 4 heads}
        \caption{256 Level, Layer 2}
        \label{fig:styleswin_l6a1}    
    \end{subfigure}
    \\
    \begin{subfigure}[b]{0.49\linewidth}         
        \includegraphics[width=.98\linewidth]{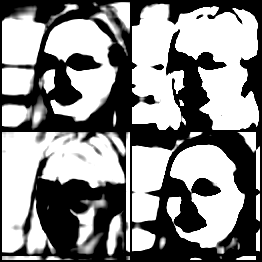}
        \label{fig:styleswin_l5a0}
        \caption{128 Level, Layer 1}
        % \caption{128 First Attention: 4 heads}
    \end{subfigure}
    \begin{subfigure}[b]{0.49\linewidth}         
        \includegraphics[width=.98\linewidth]{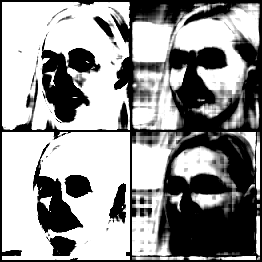}
        \caption{128 Level, Layer 2}
        \label{fig:styleswin_l5a1}
        % \caption{128 Second Attention: 4 heads}
    \end{subfigure}
    \caption{FFHQ StyleSwin attention maps at levels with 128 and 256 resolution. Every layer in each level has 4 heads with kernel size of 8 but half of them were trained with Shifted WSA. 128 resolution shows beginning indications of generative artifact.}
    \label{fig:swin_ffhq_256128}
\end{figure*}
\clearpage
% Swin FFHQ 64 x 32
\begin{figure*}[ht]
    \centering
    \begin{subfigure}[b]{0.49\linewidth}         
        \includegraphics[width=\linewidth]{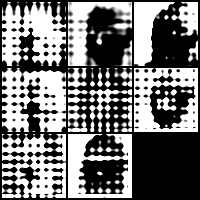}
        % \caption{64 First Attention: 4 heads}
        \caption{64 Level, Layer 1}
        \label{fig:styleswin_l4a0}    
    \end{subfigure}
    \begin{subfigure}[b]{0.49\linewidth}         
        \includegraphics[width=\linewidth]{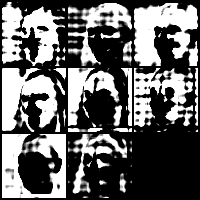}
        % \caption{64 Second Attention: 4 heads}
        \caption{64 Level, Layer 2}
        \label{fig:styleswin_l4a1}    
    \end{subfigure}
    \\
    \begin{subfigure}[b]{0.49\linewidth}         
        \includegraphics[width=\linewidth]{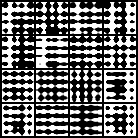}
        % \caption{32 First Attention: 4 heads}
        \caption{32 Level, Layer 1}
        \label{fig:styleswin_lXa0}    
    \end{subfigure}
    \begin{subfigure}[b]{0.49\linewidth}         
        \includegraphics[width=\linewidth]{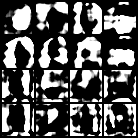}
        % \caption{32 Second Attention: 4 heads}
        \caption{32 Level, Layer 2}
        \label{fig:styleswin_lXa1}    
    \end{subfigure}
    \caption{FFHQ StyleSwin attention maps at levels with 32 and 64 resolution. Every layer in each 32\texttimes{}32 level has 16 heads, and every layer in each 64\texttimes{}64 level has 8 heads, all with kernel size of 8, but half of them were trained with Shifted WSA.}
    \label{fig:swin_ffhq_6432}
\end{figure*}
\clearpage

% Swin FFHQ 16 x 8
\begin{figure*}[ht]
    \centering
    \begin{subfigure}[b]{0.49\linewidth}         
        \includegraphics[width=\linewidth]{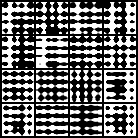}
        % \caption{16 First Attention: 4 heads}
        \caption{16 Level, Layer 1}
        \label{fig:styleswin_l3a0}    
    \end{subfigure}
    \begin{subfigure}[b]{0.49\linewidth}         
        \includegraphics[width=\linewidth]{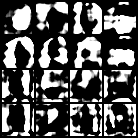}
        % \caption{16 Second Attention: 4 heads}
        \caption{16 Level, Layer 2}
        \label{fig:styleswin_l3a1}    
    \end{subfigure}
    \\
    \begin{subfigure}[b]{0.49\linewidth}         
        \includegraphics[width=\linewidth]{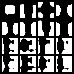}
        % \caption{8 First Attention: 4 heads}
        \caption{8 Level, Layer 1}
        \label{fig:styleswin_l2a0}    
    \end{subfigure}
    \begin{subfigure}[b]{0.49\linewidth}         
        \includegraphics[width=\linewidth]{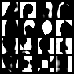}
        % \caption{8 Second Attention: 4 heads}
        \caption{8 Level, Layer 2}
        \label{fig:styleswin_l2a1}    
    \end{subfigure}
    \caption{FFHQ StyleSwin attention maps at levels with 8 and 16 resolution. Every layer in each level has 16 heads with kernel size of 8 but half of them were trained with Shifted WSA.}
    \label{fig:swin_ffhq_168}
\end{figure*}
\clearpage

%% file: Includes/Figures/na_church_attn_maps.tex
% 256 
\begin{figure*}[ht]
    \centering
    \begin{subfigure}[b]{0.49\linewidth}         
        \includegraphics[width=\linewidth]{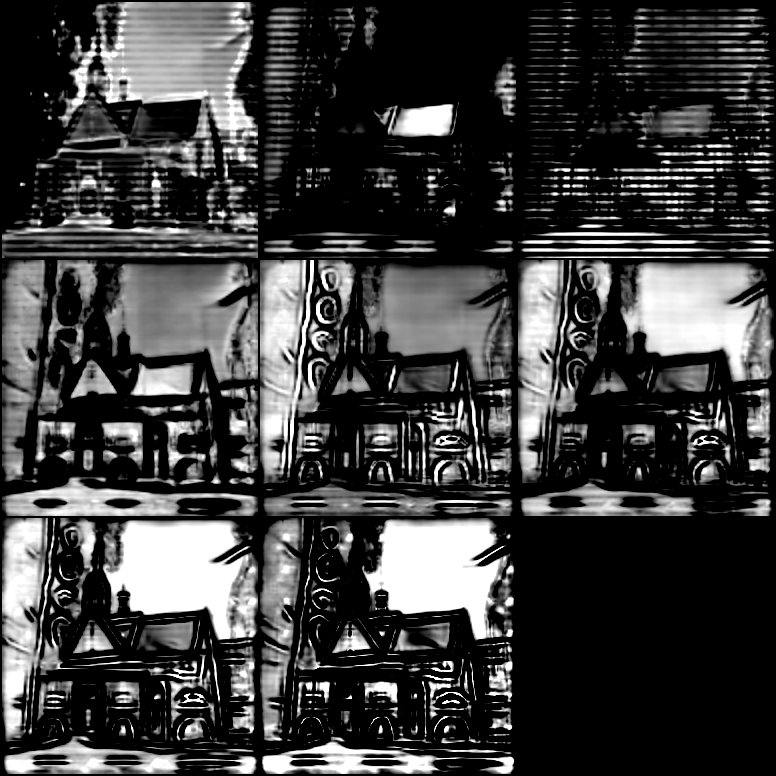}
        \caption{Good: 256 Level, Layer 1}
        % \label{fig:l4a0}    
    \end{subfigure}
    \hfill
    \begin{subfigure}[b]{0.49\linewidth}         
        \includegraphics[width=\linewidth]{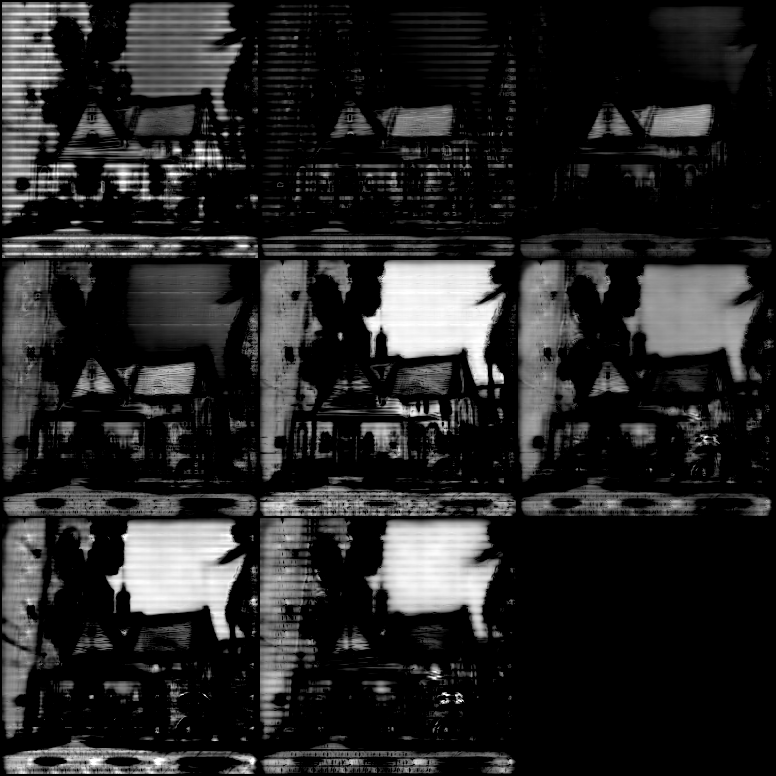}
        \caption{Good: 256 Level, Layer 2}
        % \label{fig:l4a1}    
    \end{subfigure}
    \\
    \begin{subfigure}[b]{0.49\linewidth}         
        \includegraphics[width=\linewidth]{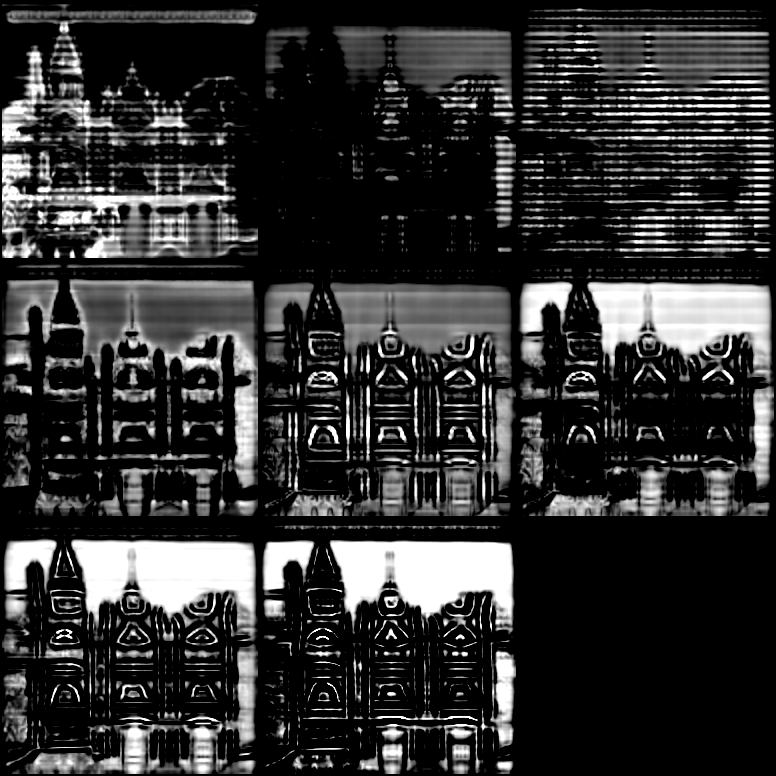}
        \caption{Bad: 256 Level, Layer 1}
        %\label{fig:badswin}    
    \end{subfigure}
    \hfill
    \begin{subfigure}[b]{0.49\linewidth}         
        \includegraphics[width=\linewidth]{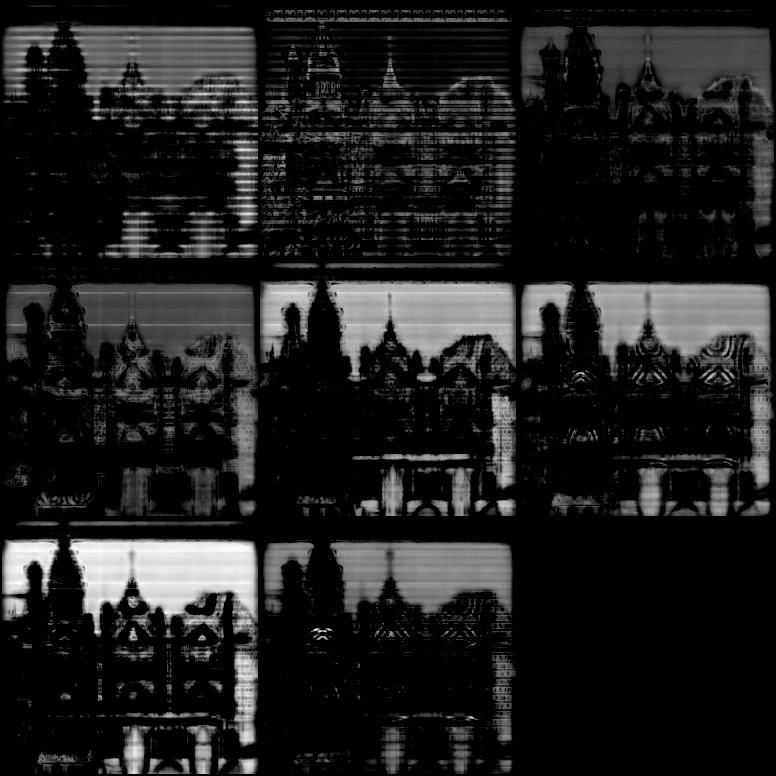}
        \caption{Bad: 256 Level, Layer 2}
        % \label{fig:lXa1}    
    \end{subfigure}
    \caption{Church StyleNAT 256 sized samples with bad and good samples. Generated images are highly predictable within both good and bad samples. Scanlines artifacts and hard lines are visible in both images, showing hard lines can be learned. ``Tree trunk'' like feature visible in good sample, with branches and swirls where foliage is located. Likely indicates a guide for the location of the tree in the scene rather than learning tree skeletal structures. Both maps can distinguish foreground and background. Bad sample looks more church like than the actual image. Notably the second layer, which we believe focuses on detail, has far lower activations in the bad sample. Despite progressive dilation, it is difficult to tell if heads are associated with local or global features, as was apparent in FFHQ.}
    \label{fig:snat_church_256}
\end{figure*}
\clearpage

% 128 
\begin{figure*}[ht]
    \centering
    \begin{subfigure}[b]{0.49\linewidth}         
        \includegraphics[width=\linewidth]{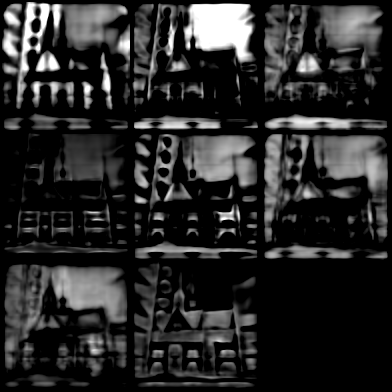}
        \caption{Good: 128 Level, Layer 1}
        % \label{fig:l4a0}    
    \end{subfigure}
    \hfill
    \begin{subfigure}[b]{0.49\linewidth}         
        \includegraphics[width=\linewidth]{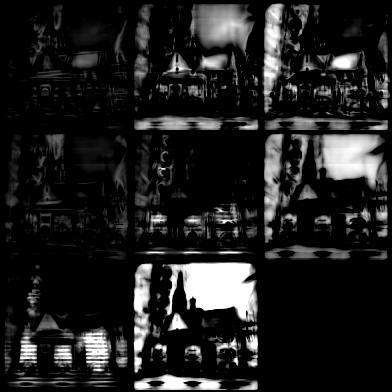}
        \caption{Good: 128 Level, Layer 2}
        % \label{fig:l4a1}    
    \end{subfigure}
    \\
    \begin{subfigure}[b]{0.49\linewidth}         
        \includegraphics[width=\linewidth]{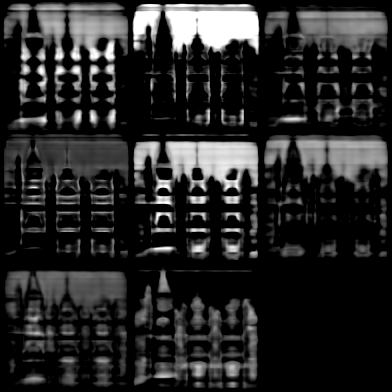}
        \caption{Bad: 128 Level, Layer 1}
        %\label{fig:badswin}    
    \end{subfigure}
    \hfill
    \begin{subfigure}[b]{0.49\linewidth}         
        \includegraphics[width=\linewidth]{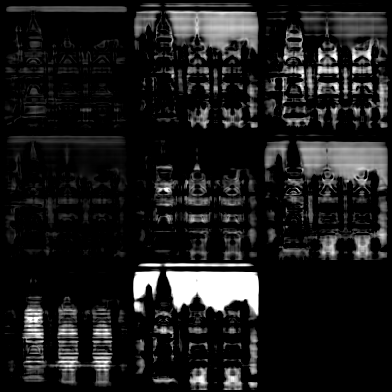}
        \caption{Bad: 128 Level, Layer 2}
        % \label{fig:lXa1}    
    \end{subfigure}
    \caption{Church StyleNAT 128 sized samples with bad and good samples. Final image fairly predictable in the good sample but the bad sample looks more akin to stacked housing apartments. Scanlines weaker in the good sample and we can see loss of coherence in the bad sample. In both samples the detail layer has lower activations with one head appearing to dominate. We believe this decoherence propagates, preventing network from learning enough detail before scaling. Similar difficulties within StyleSwin indicate that this dataset may be more challenging and that detail is more important in lower resolutions. The early heads, which have no dilations, also clearly struggle to capture fine details. This is exceptionally apparent in the second layer which is more oriented towards this task.}
    \label{fig:snat_church_128}
\end{figure*}
\clearpage
% 64 
\begin{figure*}[ht]
    \centering
    \begin{subfigure}[b]{0.49\linewidth}         
        \includegraphics[width=\linewidth]{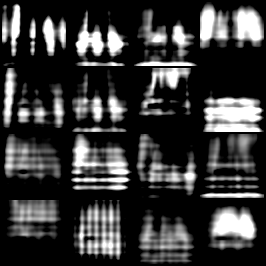}
        \caption{Good: 64 Level, Layer 1}
        % \label{fig:l4a0}    
    \end{subfigure}
    \hfill
    \begin{subfigure}[b]{0.49\linewidth}         
        \includegraphics[width=\linewidth]{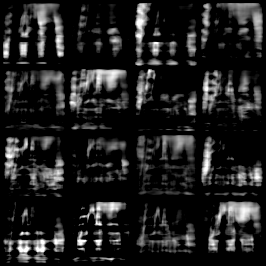}
        \caption{Good: 64 Level, Layer 2}
        % \label{fig:l4a1}    
    \end{subfigure}
    \\
    \begin{subfigure}[b]{0.49\linewidth}         
        \includegraphics[width=\linewidth]{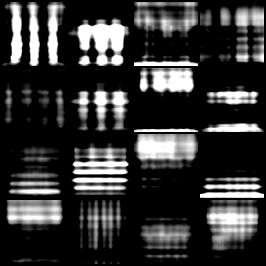}
        \caption{Bad: 64 Level, Layer 1}
        %\label{fig:badswin}    
    \end{subfigure}
    \hfill
    \begin{subfigure}[b]{0.49\linewidth}         
        \includegraphics[width=\linewidth]{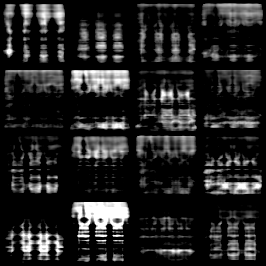}
        \caption{Bad: 64 Level, Layer 2}
        % \label{fig:lXa1}    
    \end{subfigure}
    \caption{Church StyleNAT 64 sized samples with bad and good samples. The final images are not easily predictable at this resolution and we see little coherence. General shapes can be distinguished but this is not as strong as in FFHQ. First layer clearly focuses on general structure while the second on more detail. We continue to have difficulties associating head dilation with the corresponding receptive fields of the scene. The many bands suggest that there are difficulties in locating the object's placement within the scene. Both samples have tall tower like structures within the attention maps despite not being in final image or maps of the subsequent resolution. There are a lot of similarities between both samples, especially within the first layer. This could indicate overfitting and a strong preference to a strategy.}
    \label{fig:snat_church_64}
\end{figure*}
\clearpage

% 32 
\begin{figure*}[ht]
    \centering
    \begin{subfigure}[b]{0.49\linewidth}         
        \includegraphics[width=\linewidth]{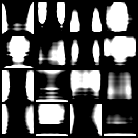}
        \caption{Good: 32 Level, Layer 1}
        % \label{fig:l4a0}    
    \end{subfigure}
    \hfill
    \begin{subfigure}[b]{0.49\linewidth}         
        \includegraphics[width=\linewidth]{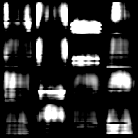}
        \caption{Good: 32 Level, Layer 2}
        % \label{fig:l4a1}    
    \end{subfigure}
    \\
    \begin{subfigure}[b]{0.49\linewidth}         
        \includegraphics[width=\linewidth]{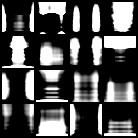}
        \caption{Bad: 32 Level, Layer 1}
        %\label{fig:badswin}    
    \end{subfigure}
    \hfill
    \begin{subfigure}[b]{0.49\linewidth}         
        \includegraphics[width=\linewidth]{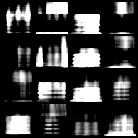}
        \caption{Bad: 32 Level, Layer 2}
        % \label{fig:lXa1}    
    \end{subfigure}
    \caption{Church StyleNAT 32 sized samples with bad and good samples. General structure is fairly coherent with blocky and tower like structures. Strong band at the bottom likely indicates attempt to generate shutterstock citation. In FFHQ we had clear placement of the subject within the scene at this level and even features like eyes and mouth. We see difficulties for this at this level, but do see towering structures. Unlike FFHQ this dataset has many differences in the general structure and location of main objects. This resolution has decent coherence for both samples but the lack of detail in the second layer may indicate how the lack of quality propagates within the network. Similar to StyleSwin these maps tend to put focus on the center of the image.}
    \label{fig:snat_church_32}
\end{figure*}
\clearpage

%% file: Includes/Figures/swin_church_attn_maps.tex
% 256 
\begin{figure*}[ht]
    \centering
    \begin{subfigure}[b]{0.49\linewidth}         
        \includegraphics[width=\linewidth]{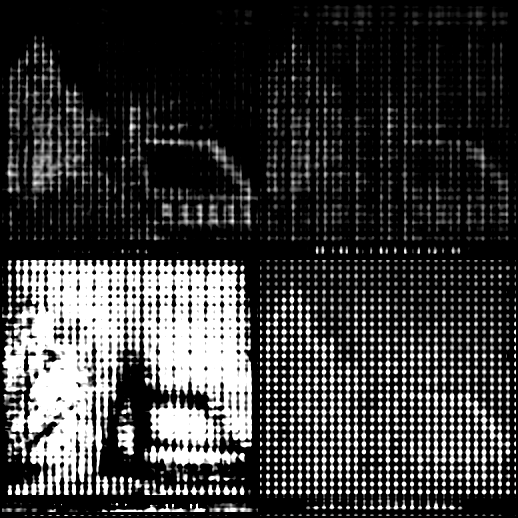}
        \caption{Good: 256 Level, Layer 1}
        % \label{fig:l4a0}    
    \end{subfigure}
    \hfill
    \begin{subfigure}[b]{0.49\linewidth}         
        \includegraphics[width=\linewidth]{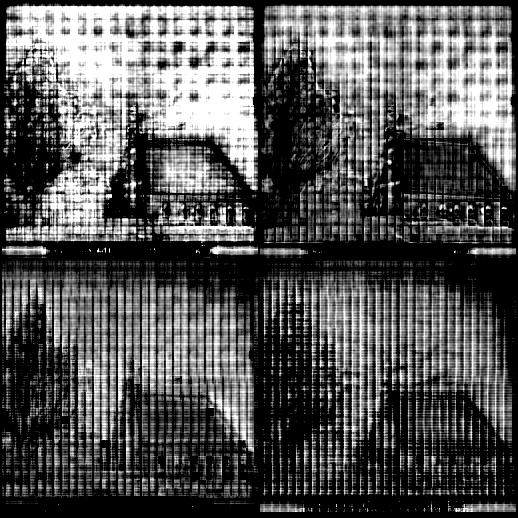}
        \caption{Good: 256 Level, Layer 2}
        % \label{fig:l4a1}    
    \end{subfigure}
    \\
    \begin{subfigure}[b]{0.49\linewidth}         
        \includegraphics[width=\linewidth]{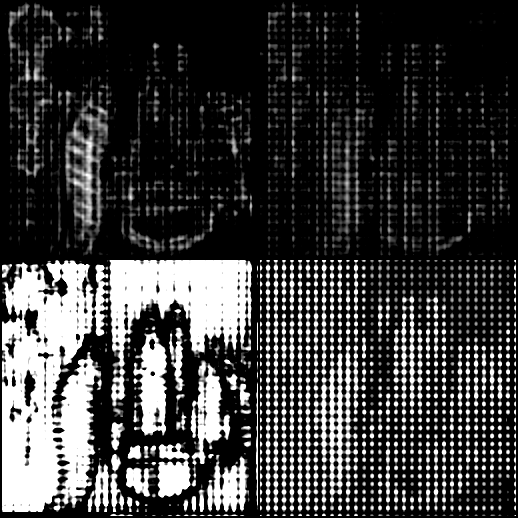}
        \caption{Bad: 256 Level, Layer 1}
        %\label{fig:badswin}    
    \end{subfigure}
    \hfill
    \begin{subfigure}[b]{0.49\linewidth}         
        \includegraphics[width=\linewidth]{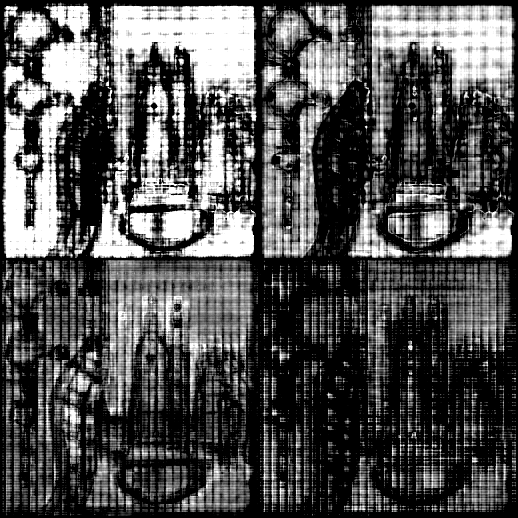}
        \caption{Bad: 256 Level, Layer 2}
        % \label{fig:lXa1}    
    \end{subfigure}
    \caption{Church StyleSwin 256 sized samples with bad and good samples. Blocky structure still exists akin to pointillism. Maps has filters reminiscent of edge filters, where the good sample can distinguish foreground and background. The tree and church are clearly visible and the good sample has a predictable final image. The floating telephone or watermark is not clearly identifiable here but we can see activations in the shutterstock citation at the bottom. Bad sample does not have as clear of an identification, and is more likely to have curved features. Similar to FFHQ the first 2 heads of the first layer have low activations while the other heads have disproportionately high.}
    \label{fig:swin_church_256}
\end{figure*}
\clearpage

% 128 
\begin{figure*}[ht]
    \centering
    \begin{subfigure}[b]{0.49\linewidth}         
        \includegraphics[width=\linewidth]{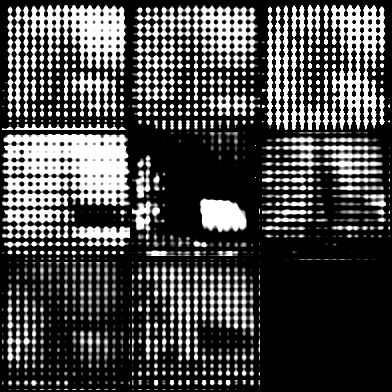}
        \caption{Good: 128 Level, Layer 1}
        % \label{fig:l4a0}    
    \end{subfigure}
    \hfill
    \begin{subfigure}[b]{0.49\linewidth}         
        \includegraphics[width=\linewidth]{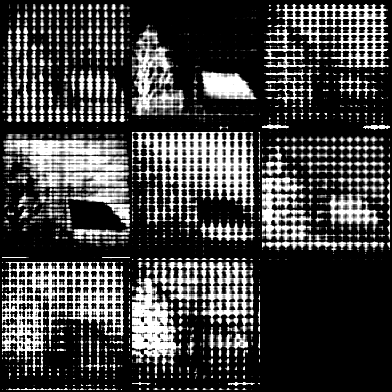}
        \caption{Good: 128 Level, Layer 2}
        % \label{fig:l4a1}    
    \end{subfigure}
    \\
    \begin{subfigure}[b]{0.49\linewidth}         
        \includegraphics[width=\linewidth]{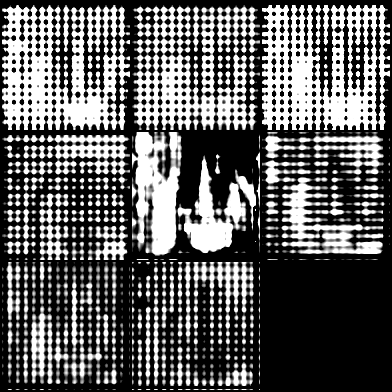}
        \caption{Bad: 128 Level, Layer 1}
        %\label{fig:badswin}    
    \end{subfigure}
    \hfill
    \begin{subfigure}[b]{0.49\linewidth}         
        \includegraphics[width=\linewidth]{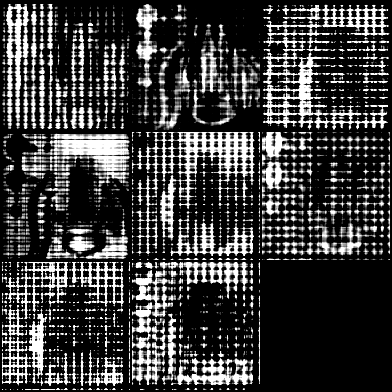}
        \caption{Bad: 128 Level, Layer 2}
        % \label{fig:lXa1}    
    \end{subfigure}
    \caption{Church StyleSwin 128 sized samples with bad and good samples. Good sample has clear good filters and some heads have strong focus on the main objects in the scene. One head in the bad sample has this same clear filter. Maps have less detailed focus, activating on many different points within the scene. Maps have less structure and features at this resolution than we saw within the FFHQ examples. In FFHQ we had less pointillism, especially in the first layer, but this is extremely prominent here indicating a difficulty in attending to the scene. Attention activation is highly disproportionate at this resolution.}
    \label{fig:swin_church_128}
\end{figure*}
\clearpage
% 64 
\begin{figure*}[ht]
    \centering
    \begin{subfigure}[b]{0.49\linewidth}         
        \includegraphics[width=\linewidth]{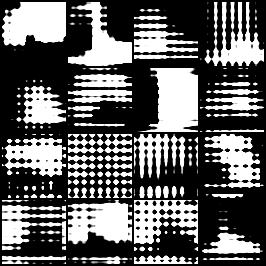}
        \caption{Good: 64 Level, Layer 1}
        % \label{fig:l4a0}    
    \end{subfigure}
    \hfill
    \begin{subfigure}[b]{0.49\linewidth}         
        \includegraphics[width=\linewidth]{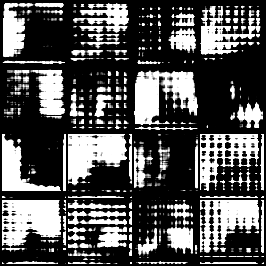}
        \caption{Good: 64 Level, Layer 2}
        % \label{fig:l4a1}    
    \end{subfigure}
    \\
    \begin{subfigure}[b]{0.49\linewidth}         
        \includegraphics[width=\linewidth]{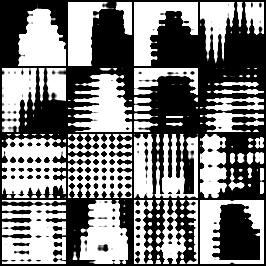}
        \caption{Bad: 64 Level, Layer 1}
        %\label{fig:badswin}    
    \end{subfigure}
    \hfill
    \begin{subfigure}[b]{0.49\linewidth}         
        \includegraphics[width=\linewidth]{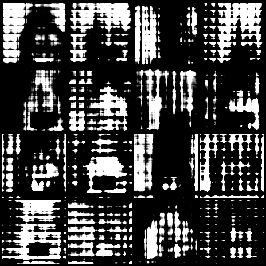}
        \caption{Bad: 64 Level, Layer 2}
        % \label{fig:lXa1}    
    \end{subfigure}
    \caption{Church StyleSwin 64 sized samples with bad and good samples. Samples are difficult to differentiate at this level and we have lower interpretability. In FFHQ the face was locatable at this resolution and the second layer started to reduce the blocking. This resolution still appears to be concentrating on the main structure of the objects, but has large range contexts in both layers. In the good sample we have difficulties identifying the tree or church, but they are somewhat visible. Attentions are highly checkerboard, likely due to the shifting of windows. The bottom of the images indicates concentration on the shutterstock citation in both samples.}
    \label{fig:swin_church_64}
\end{figure*}
\clearpage

% 32 
\begin{figure*}[ht]
    \centering
    \begin{subfigure}[b]{0.49\linewidth}         
        \includegraphics[width=\linewidth]{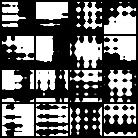}
        \caption{Good: 32 Level, Layer 1}
        % \label{fig:l4a0}    
    \end{subfigure}
    \hfill
    \begin{subfigure}[b]{0.49\linewidth}         
        \includegraphics[width=\linewidth]{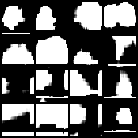}
        \caption{Good: 32 Level, Layer 2}
        % \label{fig:l4a1}    
    \end{subfigure}
    \\
    \begin{subfigure}[b]{0.49\linewidth}         
        \includegraphics[width=\linewidth]{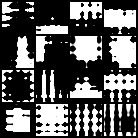}
        \caption{Bad: 32 Level, Layer 1}
        %\label{fig:badswin}    
    \end{subfigure}
    \hfill
    \begin{subfigure}[b]{0.49\linewidth}         
        \includegraphics[width=\linewidth]{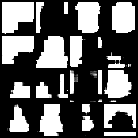}
        \caption{Bad: 32 Level, Layer 2}
        % \label{fig:lXa1}    
    \end{subfigure}
    \caption{Church StyleSwin 32 sized samples with bad and good samples. Global structure is generally lost and would be difficult to predict produced sample from these maps. The church is identifiable in the second layer of the good sample, but attention is a bit scattered. First level is more sporadic compared to the second level, which is more connected. Similar to FFHQ the first layer has large checkerboard patterns and second layer is smoother, though less general structure is identifiable. This appears to indicate that the first layer is matching structure to the upscaling and the second layer concentrates on details. Coherence is likely lost after this resolution.}
    \label{fig:swin_church_32}
\end{figure*}
\clearpage